%% file: main.tex
\documentclass[runningheads]{llncs}

\usepackage{graphicx}
\usepackage{comment}
\usepackage{amsmath,amssymb}
\usepackage{color}
\usepackage{url}
\usepackage{hyperref}
\usepackage{wrapfig}
\usepackage{float}

\usepackage{booktabs}
\usepackage{array}
\usepackage{subcaption}

\usepackage{colortbl}
\usepackage[table,xcdraw]{xcolor}
\usepackage{multirow}
\usepackage{arydshln}
\usepackage{tabularx}
\usepackage{booktabs}
\usepackage{cleveref}
\usepackage{tikz}
\usepackage{pgfplots}

\newcommand{\myparagraph}[1]{\noindent\textbf{#1.}}
\newcommand{\methodname}{VisualChef\xspace}

\usepackage{xspace}
\newcommand{\eg}{e.g.\@\xspace}
\newcommand{\ie}{i.e.\@\xspace}

\begin{document}

\title{VisualChef: Generating Visual Aids in Cooking via Mask Inpainting}

\author{Oleh Kuzyk\inst{1} \and
Li Zuoyue\inst{1} \and
Marc Pollefeys\inst{1,2} \and
Xi Wang\inst{1,3}
}

\authorrunning{O. Kuzyk et al.}

\institute{ETH Zürich \and Microsoft \and TU Munich \& MCML}

\maketitle

\begin{center}
   \vspace{-5mm}
   \centerline{ 
    \includegraphics[trim=0 0mm 0mm 0mm, clip=true, width=1.0\linewidth]{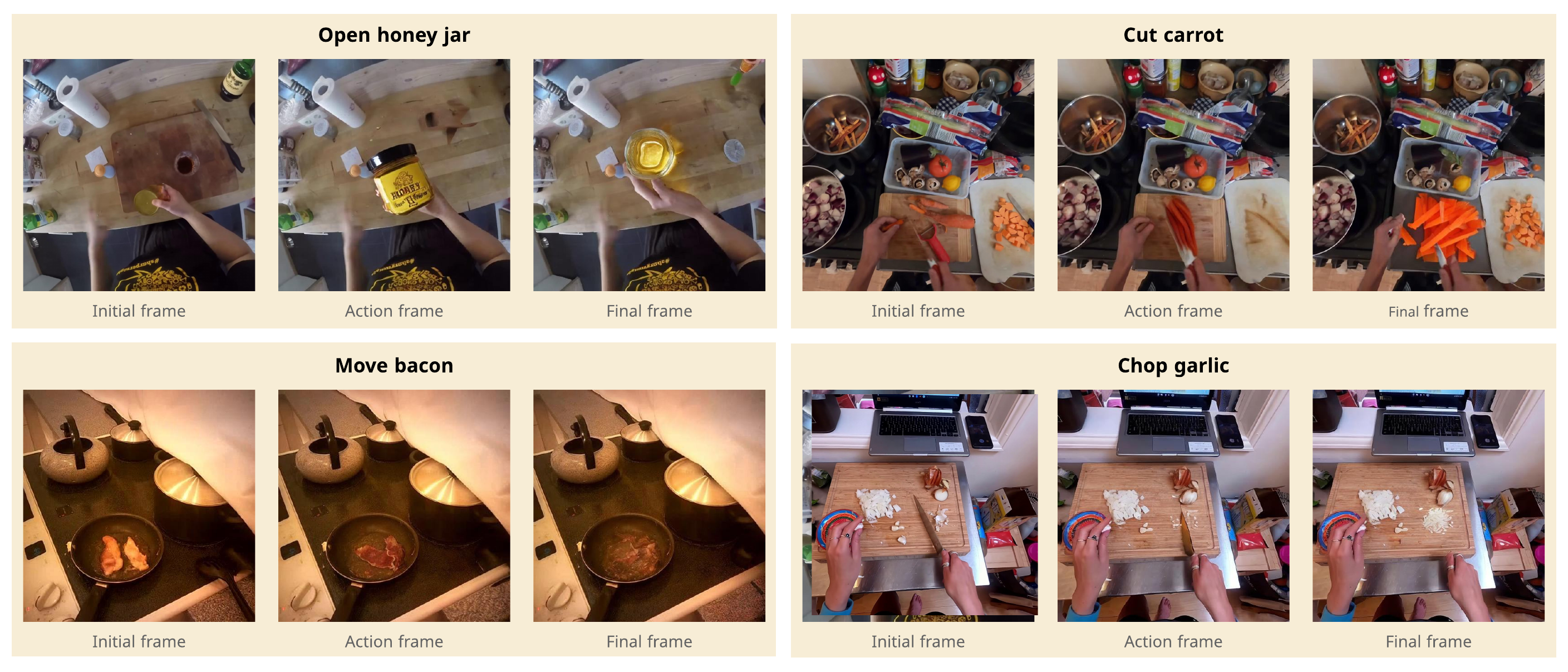}}
  \vspace{-0.05in}
\captionof{figure}{\textbf{Generating contextual action and final state frames via mask inpainting.} Given an initial frame and an action, \methodname generates two frames visualizing both the action's execution and the resulting appearance of the object while preserving the environment depicted in the input frame. 
}
\label{fig:teaser}
\vspace{-6mm}
\end{center}%

\input{sec/0_absract}
\input{sec/1_intro}

\input{figures/method}

\input{sec/2_related}

\input{sec/3_method}
\input{sec/4_exp}
\input{sec/5_conclusions}

\bibliographystyle{splncs04}
\bibliography{main}

\input{sec/X_suppl}

\end{document}

%% file: sec/0_absract.tex
\begin{abstract}
Cooking requires not only following instructions but also understanding, executing, and monitoring each step--a process that can be challenging without visual guidance. 
Although recipe images and videos offer helpful cues, they often lack consistency in focus, tools, and setup. 
To better support the cooking process, we introduce \methodname, a method for generating contextual visual aids tailored to cooking scenarios. 
Given an initial frame and a specified action, \methodname generates images depicting both the action's execution and the resulting appearance of the object, while preserving the initial frame's environment. 
Previous work aims to integrate knowledge extracted from large language models by generating detailed textual descriptions to guide image generation, which requires fine-grained visual-textual alignment and involves additional annotations.  
In contrast, \methodname simplifies alignment through mask-based visual grounding. 
Our key insight is identifying action-relevant objects and classifying them to enable targeted modifications that reflect the intended action and outcome while maintaining a consistent environment.
In addition, we propose an automated pipeline to extract high-quality initial, action, and final state frames. 
We evaluate \methodname quantitatively and qualitatively on three egocentric video datasets and show its improvements over state-of-the-art methods. 

\keywords{Image generation \and Visual grounding}
\end{abstract}

%% file: sec/1_intro.tex
\section{Introduction}
\label{sec:intro}

Cooking is a complex task that requires not only following instructions, but also understanding, executing, and monitoring each step. Often, the only resource available is a step-by-step recipe, which can leave important details ambiguous.

Many cooks turn to images or videos for guidance, as visual cues like sauce consistency or vegetable thickness greatly aid cooking~\cite{kuoppamaki2021designing,madera2013breaking,tran2005cook}.
However, these generic resources often differ in ingredients, tools, or setups, making them less compatible with one's environment and progress. 
They lack the contextual alignment needed to support a unique situation.  
Motivated by this, we aim to generate visual aids adapted to the cook's environment and progress in the recipe, offering more personalized, contextually relevant guidance. 
These aids benefit human users and robotic agents designed to perform cooking tasks in the kitchen. 

In this paper, we propose \methodname, a simple yet effective approach to generate visual aids in cooking scenarios. 
Given an initial frame and a specified action, \methodname generates two context-preserving images: one showing how the action is executed and the other depicting the state of the resulting object.
Previous methods integrate knowledge extracted from large language models by generating detailed textual descriptions to guide image generation~\cite{brooks2023instructpix2pix,hertz2022prompttoprompt,lai2024lego,souček2024genhowto,wang2023instructedit}. 
However, they require fine-grained visual-textual alignment and extra data annotations.  
In contrast, \methodname simplifies alignment through mask-based visual grounding and focuses on object-centric inpainting. 

Our central idea is to identify action-relevant objects and develop tailored strategies based on their roles to guide the inpainting process. Specifically, we classify objects into three categories: \textit{core objects}, central to the action (e.g., ingredients); \textit{location objects}, which define spatial context (e.g., cutting boards); and \textit{functional objects}, which assist the action without being essential to the final state (e.g., knives). This categorization enables selective editing. We use masked inpainting~\cite{kohler2014mask,lugmayr2022repaint,rombach2022highresolutionimagesynthesislatent,suvorov2022resolution,zhu2021image} to generate realistic, context-aware images aligned with the cooking action and final state. To train effectively, we introduce a data curation pipeline that extracts triplets of initial, action, and final frames from egocentric videos using hand detection, object presence, and relevance filtering.

We evaluate \methodname against state-of-the-art approaches on three egocentric video datasets: Ego4D~\cite{lai2024lego}, EGTEA Gaze+~\cite{li2020eye} and EPIC-KITCHENS-100 (EK-100)~\cite{Damen2022RESCALING}. 
\methodname achieves higher fidelity in depicting the action and final state frames compared to previous generation models~\cite{brooks2023instructpix2pix,lai2024lego,rombach2022highresolutionimagesynthesislatent,souček2024genhowto}, especially outperforming the state-of-the-art methods~\cite{lai2024lego,souček2024genhowto}, highlighting the effectiveness of selective inpainting. 
Our code, data, and models will be available for research purposes. 
In summary, our contributions are as follows.

\begin{itemize}
    \item We propose \methodname, a mask-based image inpainting model that generates visual aids in cooking scenarios, given an initial frame and a specified action. 
    \item We introduce an automatic data curation pipeline that extracts initial-action-final frame triplets from egocentric videos with action annotations. 
    \item We evaluate \methodname on 3 egocentric video datasets with qualitative and quantitative results that show improvements over state-of-the-art methods. 
\end{itemize}

%% file: figures/method.tex
\begin{figure*}[ht]
    \centering
    \includegraphics[width=\linewidth]{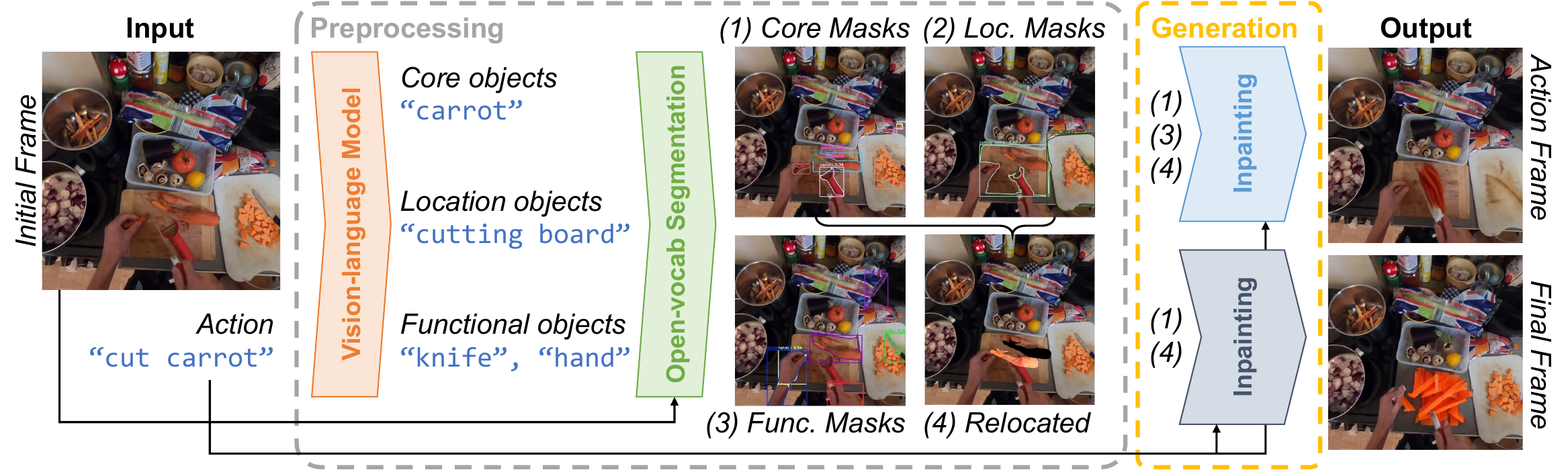}
    \caption{\textbf{The \methodname pipeline for context-aware inpainting within a cooking scenario.} It starts with an \textit{Initial Frame} ($f_\text{in}$) as input, paired with an \textit{Action} description (\eg, \texttt{"cut carrot"}). The vision-language model LLaVA~\cite{liu2023visualinstructiontuning} is employed to identify relevant objects and classify them into three categories: Core objects (\eg, \texttt{"carrot"}), Location objects (\eg, \texttt{"cutting board"}), and Functional objects (\eg, \texttt{"knife"} and \texttt{"hand"}). Using the open-vocabulary segmentation model Grounding DINO~\cite{liu2024groundingdinomarryingdino}, the masks for these objects are generated: (1) \textit{Core Masks}, (2) \textit{Location Masks}, and (3) \textit{Functional Masks}. Core objects are \textit{Relocated} in an additional step (4) as needed. The generation phase involves two different inpainting modules based on Stable Diffusion~\cite{rombach2022highresolutionimagesynthesislatent}, conditioned on different combinations of the masks for creating the \textit{Action Frame} ($f_\text{action}$) that reflects the step being performed and the \textit{Final Frame} ($f_\text{final}$) showing the status upon action completion. The output thus visualizes the progression of the cooking scenario in a realistic manner.
    }
    \label{fig:method}
    \vspace{-5mm}
\end{figure*}

%% file: sec/2_related.tex
\section{Related work}
\label{sec:related}

Our work is at the cross section of image editing, conditional image generation, and egocentric vision.
We discuss related works according to the underlying categories in the following.

\noindent\textbf{Conditional image generation} has evolved from traditional GANs~\cite{mirza2014conditionalgenerativeadversarialnets} to modern diffusion models~\cite{ho2020denoisingdiffusionprobabilisticmodels}, enabling high-resolution generation based on given condition signals.
While GANs were used for text- and image-conditioned synthesis~\cite{huang2021multimodalconditionalimagesynthesis,odena2017conditionalimagesynthesisauxiliary}, diffusion models~\cite{dhariwal2021diffusionmodelsbeatgans,ho2020denoisingdiffusionprobabilisticmodels,zhang2023addingconditionalcontroltexttoimage,zhu2023conditionaltextimagegeneration} offer stronger performance by better modeling complex data distributions. 
Unlike general-purpose methods targeting broad use cases, our approach is tailored to cooking, enabling targeted, context-aware image updates that preserve visual consistency.

\noindent\textbf{Image editing} has advanced significantly with generative models, particularly diffusion models that excel in text-conditioned image generation.
Notable methods include Imagic~\cite{kawar2023imagic}, which enables precise edits without disrupting the global structure, and Prompt-to-Prompt~\cite{hertz2022prompttoprompt}, which allows fine-grained visual control through text.
InstructPix2Pix~\cite{brooks2023instructpix2pix} supports detailed user-driven edits, while SDEdit~\cite{meng2022sdedit} uses a stochastic process for realistic, context-aware alterations.
ProxEdit~\cite{han2023improving} and InstructEdit~\cite{wang2023instructedit} further enhance controlled editing via text-driven modifications.
Recent advancements have integrated multimodal and action-guided generation, such as GenHowTo~\cite{souček2024genhowto}, LEGO~\cite{lai2024lego}, and Instruct-Imagen~\cite{hu2024instructimagen}, which synthesize visual content from combined image-text inputs.
GenHowTo and LEGO rely on vision-language models to caption input frames and generate output prompts. LEGO also uses in-context learning to extract fine-grained action descriptions with hand and object bounding boxes.
Although many of these models are built on Stable Diffusion~\cite{rombach2022highresolutionimagesynthesislatent}, alternatives such as Text2LIVE~\cite{bartal2022text2live} offer text-guided, real-time edits. 
These developments reflect the field's shift toward interactive, user-friendly visual content creation.
For mask-based editing, models like DiffEdit~\cite{couairon2022diffeditdiffusionbasedsemanticimage} and Locate and Forget~\cite{li2024textguidedimageediting} allow for high-precision mask-guided modifications.
LAR-Gen \cite{pan2024locateassignrefinetaming} improves inpainting control via textual guidance, and BrushNet \cite{ju2024brushnetplugandplayimageinpainting} advances plug-and-play editing with a dual-branch diffusion setup.
Further contributions such as Point\&Instruct \cite{helbling2024pointinstructenablingprecise}, MAG-Edit \cite{mao2023mageditlocalizedimageediting}, and CoVLM \cite{li2023covlmcomposingvisualentities} focus on localized and contextual editing, while FoI \cite{guo2023focusinstructionfinegrainedmultiinstruction} employs attention mechanisms for multi-instruction editing.
Collectively, these works drive precise mask-based image editing, broadening the scope and flexibility of visual content manipulation.
Our work relies on the principles of mask-based inpainting to selectively alter relevant parts of the frame.
However, our key novelty lies in categorizing objects (core, location, functional) to guide which regions to inpaint, aligning visual modifications with specific actions in cooking sequences.
Our method goes beyond the capabilities of the state-of-the-art by emphasizing object relevance and context retention during editing.

\noindent\textbf{Egocentric vision} research has prospered with the emergence of rich multimodal datasets capturing real-world activities.
EGTEA Gaze+\cite{li2020eye} focuses on cooking, providing fine-grained annotations such as hand masks and gaze tracking.
EPIC-KITCHENS\cite{Damen2018EPICKITCHENS,Damen2022RESCALING} expands on this with unscripted, culturally diverse kitchen videos and detailed action labels, making it one of the largest egocentric cooking datasets for kitchen tasks.
Ego4D~\cite{grauman2022ego4d} and EgoExo4D~\cite{grauman2023egoexo4d} also offer diverse daily activities, including cooking, captured with multimodal data such as 3D scans and gaze.
Our work leverages such egocentric kitchen data to extract high-quality triplets (initial, action, final) for training.
This enables our model to learn task-specific edits aligned with real-world practice, setting our approach apart from general vision-language models that lack this cooking-specific focus.

%% file: sec/3_method.tex
\section{Method}
\label{sec:method}
We first describe the task of generating action-conditioned images and the empirical observations that motivated our approach. These are followed by vision-language model prompting and diffusion-based inpainting.   

\subsection{Overview}

Given an input frame $f_{\text{in}}$ representing the current cooking state and the next action $a$ from a recipe, we aim to generate images illustrating how this action is performed in $f_{\text{action}}$ and how relevant object(s) appear afterward in $f_{\text{final}}$.
Here, $f_{\text{in}}$ is the \textit{initial} frame, $f_{\text{action}}$ depicts the action execution, and $f_{\text{final}}$ represents the post-action state. Both $f_{\text{action}}$ and $f_{\text{final}}$ build on $f_{\text{in}}$ as the starting environment.
See Figure \ref{fig:method} for an example.

Unlike previous approaches~\cite{lai2024lego,souček2024genhowto} that rely on vision-language models to generate detailed captions, our method \methodname distinguishes between different types of objects
and applies tailored strategies for each (see Figure \ref{fig:method}).

\subsection{Selecting Image Triplets}

While most egocentric video datasets provide action labels with start/end timestamps, few annotate critical frames.
To obtain image triplets from these videos, we develop a selection strategy.
We define $t_s$ and $t_f$ as the start and end times of the action, respectively.
We select the frame at the beginning of each action as the \textit{initial frame}, \ie $f_{\text{in}} = f_{t_s}$.
Similar to previous work~\cite{lai2024lego}, the action frame is chosen from the midpoint of the action, \ie $f_{\text{action}}=f_{\frac{1}{2}(t_s + t_f)}$, under the assumption that this frame captures the core of the activity.
Although multiple frames could serve as the action frame, our empirical results show that selecting the middle frame yields more qualitative results (see \cref{ssec:curation} for details).
For the \textit{final frame}, we opt for the frame at 90\% of duration, \ie, $f_{\text{final}}=f_{\frac{1}{10}t_s + \frac{9}{10}t_f}$ to ensure that relevant objects remain visible before cleanup or transition begins.

\myparagraph{Filtering} We then filter out image triplets that do not contain action-relevant objects through the following steps.
\vspace{-2mm}
\begin{enumerate}
    \item \textbf{Object Identification:} We use LLaVA \cite{liu2023visualinstructiontuning} to identify visible objects relevant to the action (see \cref{ssec:llava}). The input to the model consists of the action $a$ and the initial frame $f_\text{in}$. The output is a list of objects $[o_i]$ present in $f_\text{in}$ and associated with $a$.
    \item \textbf{Object Detection:} An open-vocabulary object detector \cite{liu2024groundingdinomarryingdino} processes $f_\text{in}$ and $[o_i]$. The output includes detection scores $[s_i]$.
    Detection for $f_\text{final}$ is omitted to avoid excluding informative triplets, as objects may change shape or disappear at the end of the action.
    Any detection with $s_i<0.3$ is discarded.  

    \item \textbf{Hand Detection:}
    We observe that all quality frames tend to include hands in $f_\text{in}$, as the person is likely preparing to perform an action, often with their hands already in contact with objects. Additionally, the hand may partly cover an object, making it challenging for the model to accurately identify the object. 
    We use the model to detect hands in both
    $f_\text{in}$ and $f_\text{action}$.
    Detection is omitted in $f_\text{final}$ because it usually shows the completed action without hands. 

    \item \textbf{Frame Filtering:} Based on the detection results, we filter the image triplets ($f_\text{in}$, $f_\text{action}$, $f_\text{final}$) where $f_\text{in}$ does not include at least hands or any of the identified relevant objects, or where $f_\text{action}$ lacks visible hands. 
\end{enumerate}
\vspace{-2mm}

\noindent These criteria are input requirements for using the model in real-world cases.

\subsection{Classifying Relevant Objects}
\label{ssec:llava}

\begin{figure*}
    \centering
    \includegraphics[width=\linewidth]{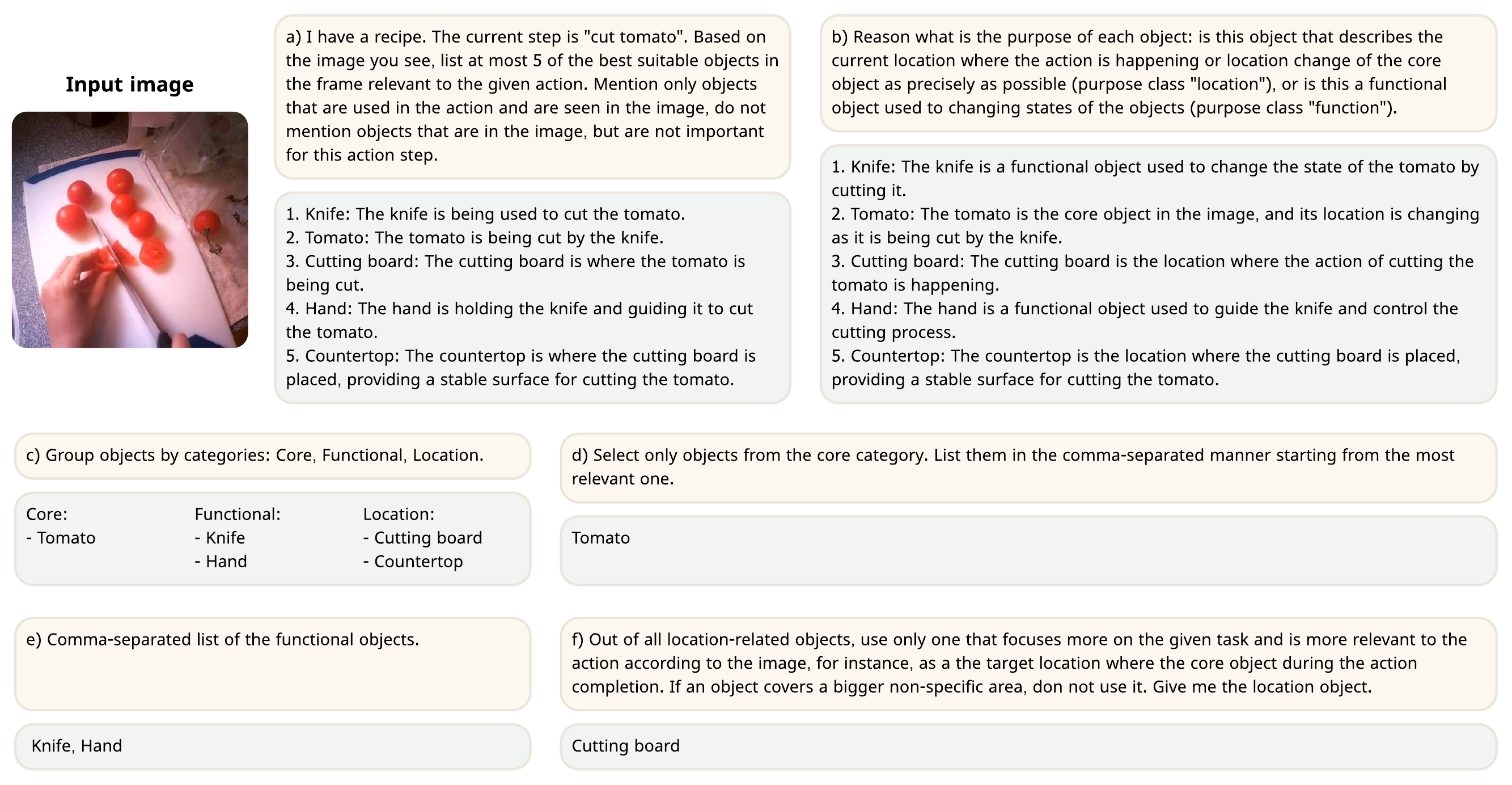}
    \caption{\textbf{Chain-of-thoughts reasoning for relevant object identification.} Given an input image, we use the chain-of-thoughts strategy to prompt LLaVA~\cite{liu2023visualinstructiontuning} to get a categorized list of objects relevant to the given action.}
    \label{fig:llava-prompt}
    \vspace{-6mm}
\end{figure*}

\methodname first identifies the key objects relevant to a given action;
for example, a knife is generally required for cutting. 
We ground the requirements in the scene, adapting based on the available objects in case an alternative way is feasible; \eg, if a cutting board is absent, the action may still proceed on a plate.

Real-world scenes are often cluttered with many objects,
yet not all are relevant to the action. 
Conversely, we observe that relying solely on the objects explicitly mentioned in the action is insufficient (see supp. mat. for details). 
\methodname, therefore, identifies action-relevant objects and categorizes them based on specific roles. 
Specifically, we consider the following three object categories:

\begin{enumerate}
    \item \textit{Core objects} that are essential to the action itself (e.g., the tomato in a ``Cut tomato" action).
    \item \textit{Location objects} that describe the initial or final locations of the objects involved (e.g., a cutting board, pan, or stove).
    \item \textit{Functional objects}: These serve a purpose in the action but do not move other objects, nor is their location important (e.g., hands or a knife when cutting a tomato).
\end{enumerate}

We follow the chain-of-thoughts strategy~\cite{wei2023chainofthoughtpromptingelicitsreasoning} to prompt the open-source vision language model, LLaVA~\cite{liu2023visualinstructiontuning}, for object relevance and categorization.
The model is first asked to identify action-relevant objects and then categorize them by type (see \cref{fig:llava-prompt} for prompt examples).

Often, multiple \textit{location objects} refer to the same spatial area.
For example, in the action of placing cheese (see supp. mat.), LLaVA may return the burger, pan, and stove. Although all of them are valid, the burger defines the target location the most precisely. To address this, we add a reasoning step prompting LLaVA to filter and retain only the most specific location object for the action.

\subsection{Masking Relevant Objects}
If we apply a diffusion pipeline to the entire image, we risk generating an image that looks completely different.
While the target action may be depicted, the environment, tools, and core objects can change drastically.
Our goal, instead, is to preserve the original setting, modifying only elements that require change.

To achieve this, we use existing visual grounding methods~\cite{liu2024groundingdinomarryingdino,ren2024grounded} 
to segment the relevant objects. 
Their bounding boxes form a mask used for inpainting to generate the target frames.
Empirical results show that bounding boxes are an effective, compact representation that provides object-related information while being coarse enough to allow shape and pose changes.

We always mask all \textit{core objects}, as they are central to the action, and use their locations for generating both $f_{\text{action}}$ and $f_{\text{final}}$. 
The bounding box of the \textit{location object} is used to move the \textit{core object} to its final location in both frames. 
\textit{Functional objects} are used only for generating $f_{\text{action}}$, where they support the depiction of the action, but are not involved in $f_{\text{final}}$ generation.

\subsection{Diffusion-based Inpainting}

The final stage of our pipeline uses Stable Diffusion~\cite{rombach2022highresolutionimagesynthesislatent} for inpainting.
In this stage, we provide the initial image $f_{\text{in}}$, a natural language description of the action $a$, and the masks generated earlier.
The stage output includes the generated target frames $f_{\text{action}}$ and $f_{\text{final}}$.
To generate $f_{\text{action}}$, we use a two-step inpainting procedure. First, we inpaint \textit{functional objects} to put them into position for performing $a$. Then we inpaint this intermediate result using \textit{core object} masks. When generating $f_{\text{final}}$, we skip the first step and directly inpaint the $f_{\text{in}}$.
For optimization, we use a loss function based on the negative log CLIP score: 
\begin{equation}
    \mathcal{L}(f_\text{in}, \hat{f}_\text{out}) = -\log \textbf{CLIP}(f_\text{in}, \hat{f}_\text{out}),
\end{equation}
where \textbf{CLIP} measures cosine similarity between the CLIP embeddings~\cite{radford2021learning} of $f_\text{in}$ and the target frame $\hat{f}_\text{out}$ ($f_\text{action}$ or $f_\text{final})$.
Similar to negative log-likelihood, this approach %
ensures that the generated image remains semantically consistent with the given action description while achieving high-quality inpainting results.

\myparagraph{Implementation Details}
We use the version of Stable Diffusion specifically designed for masked inpainting tasks~\cite{rombach2022highresolutionimagesynthesislatent}. 
The model is fine-tuned separately on our datasets for two target frames  over 5 epochs.%

%% file: sec/4_exp.tex
\section{Experiments}
\label{sec:exp}

We introduce the experimental setup in \cref{sec:setup},
compare to previous SOTA methods in Section~\ref{sec:main_results}, and discuss the ablation studies in Section~\ref{sec:ablations}. 

\subsection{Experimental Setup}
\label{sec:setup}

\myparagraph{Datasets} To ensure a fair and dataset-specific evaluation, the model was trained and evaluated separately on each of the three datasets. Specifically, for each experiment, the model was trained using only the training split of a single dataset and evaluated on its corresponding test split.

\begin{itemize}
\item  \textbf{Ego4D} \cite{grauman2022ego4d}. 
We extract cooking activities from the Ego4D dataset for the \textit{Forecasting + Hands \& Objects} task, yielding 80 videos with 11K actions and annotations for three key frames per action. 
These are the pre-condition (\textit{PRE}), point-of-no-return (\textit{PNR}), and post-condition (\textit{POST}) frames, which correspond to $f_{\text{in}}$, $f_{\text{action}}$ and $f_{\text{final}}$. 
Following a setup similar to LEGO and GenHowTo, we split 80\% of the data for training and 20\% for testing.

\item \textbf{EGTEA Gaze+} \cite{li2020eye} is a multimodal egocentric cooking video dataset.
It contains 28 hours of video content, with frame-level action annotations, pixel-level hand masks, and gaze-tracking data.
Using our strategy, we select 3K image triplets, allocating 80\% for training and 20\% for testing. 

\item \textbf{EK-100}~\cite{Damen2022RESCALING} is the largest egocentric cooking dataset with 100 hours of unscripted daily routine cooking videos with automatically annotated hand-object bounding boxes and object masks. We select 3K triplets, splitting them into 80\% for training and 20\% for testing.  

\end{itemize}
\vspace{-2mm}

\myparagraph{Baselines}
We compare our method with state-of-the-art methods for action-conditioned image generation, including Stable Diffusion~\cite{rombach2022highresolutionimagesynthesislatent}, InstructPix2Pix~\cite{brooks2023instructpix2pix}, GenHowTo~\cite{souček2024genhowto}, and LEGO~\cite{lai2024lego}. 

\begin{itemize}

\item \textbf{Stable Diffusion} \cite{rombach2022highresolutionimagesynthesislatent} is included in its original form to demonstrate the improvement achieved for this task compared to the standard model.

\item\textbf{InstructPix2Pix} \cite{brooks2023instructpix2pix} is a model based on Stable Diffusion, trained for editing images according to instructions.

\item\textbf{GenHowTo} \cite{souček2024genhowto} shares the same task setting and focuses on action and final state frames. It builds on Stable Diffusion, using U-Net and ControlNet.

\item\textbf{LEGO} \cite{lai2024lego} is a model for action image generation in egocentric videos that utilizes VLM to enrich narrations. 
\end{itemize}

GenHowTo and LEGO are two closest baselines; however,\methodname offers a simpler, more efficient alternative with improved alignment and lower overhead (full details in the supp. material).

For the SOTA baselines, we used the official implementations and pretrained weights from the original authors. When dataset-specific weights were available (e.g., LEGO), we used those. For models that were designed to be general-purpose, we relied on publicly available weights without additional finetuning.

\myparagraph{Evaluation Metrics}
\label{sec:metrics}
Following standard evaluation protocols for generative models~\cite{ho2020denoisingdiffusionprobabilisticmodels,kasem2019,ramesh2021zeroshottexttoimagegeneration,saharia2022photorealistictexttoimagediffusionmodels,SONG202083}, we report FID~\cite{heusel2017gans}, PSNR~\cite{psnr5596999} and SSIM~\cite{ssim} scores. 
Recognizing the limitations of these metrics~\cite{stein2024exposing}, we additionally introduce a set of novel evaluation metrics based on CLIP similarity between pairs of images. 

\begin{itemize}
\item \textbf{CLIP}~\cite{radford2021learning} cosine similarity. 
\item \textbf{M-CLIP} compares only the masked areas where changes are applied. 
\item  \textbf{D-CLIP} considers the similarity between $f_\text{in}$ and $f^\text{GT}_\text{out}$. When similarity is high, the metric penalizes major changes to preserve the initial frame. When low, it allows and even encourages greater deviations in the generated output.
\begin{equation*}
    \textbf{D-CLIP}(f_\text{in}, \hat{f}_\text{out}) = \frac{\textbf{CLIP}(f_\text{in}, f^\text{GT}_\text{out}) - \textbf{CLIP}(f_\text{in}, \hat{f}_\text{out})}{\textbf{CLIP}(f_\text{in}, f^\text{GT}_\text{out})}
\end{equation*}
\end{itemize}

\subsection{Evaluation of Dataset Curation Strategy}
\label{ssec:curation}

To evaluate our data curation strategy, we manually select 50 triplets from EGTEA Gaze+ that best represent the initial, action, and final states of each action, creating a benchmark evaluation set. 
We then measured the similarity between these and the automatically selected frames using CLIP scores.

\begin{wraptable}{r}{0.55\textwidth}
    \vspace{-7mm}
    \scriptsize
    \centering
    \caption[]{\textbf{Data curation evaluation.} We assess our strategy on a manually selected set and show the impact of filtering.
    \label{tab:eval_egtea_dataset}}
    
    \begin{tabular}{ccccc}
    \hline
     \multirow{2}{*}{Frame} & \multicolumn{2}{c}{Before filtering} & \multicolumn{2}{c}{After filtering} \\
     \cmidrule(rl){2-3} \cmidrule(rl){4-5}
     & CLIP & Quantile $\geq$ 80 & CLIP & Quantile $\geq$ 80 \\
    \hline
		$f_{\text{in}}$ & 83.53 & 70.8$\%$ & 85.56 & 81.5\%\\
		$f_{\text{action}}$ & 90.87 & 91.7$\%$ & 91.14 & 92.6\%\\
		$f_{\text{final}}$ &82.11  & 62.5$\%$ & 85.71 & 74.1\%\\
    \hline
    \end{tabular}
    \vspace{-7mm}
\end{wraptable}

The quantitative results in \cref{tab:eval_egtea_dataset} demonstrate the effectiveness of the strategy, with 80 empirically set as the threshold for strong similarity.
On average, the automatically selected frames align well with the manual evaluation set.
The action frames $f_{\text{action}}$ show a high degree of similarity, as evidenced by the defined quantile. 
This can be attributed to the repetitive nature of many actions; \eg, during a tomato cutting, most frames are nearly identical, differing only in incremental progress. 
The initial frames $f_{\text{in}}$ also perform well, while the final frames $f_{\text{final}}$ score the lowest, reflecting the common tendency in cooking videos to move quickly to the next step after completing an action without stopping to show the result.
This challenge has been observed consistently throughout the project and explains why other models (e.g.,~\cite{lai2024lego}) prioritize $f_{\text{action}}$ over $f_{\text{final}}$.
After selection, we apply filtering to exclude low quality or insufficient frames which significantly improve CLIP scores (see \cref{tab:eval_egtea_dataset}; more on the supp. mat).

\begin{table}[htbp!]
    \centering
    \scriptsize
    \caption[]{\textbf{Comparison with the state-of-the-art.} Our method \methodname outperforms state-of-the-art methods by a large margin in terms of content relevance, indicated by CLIP and D-CLIP.
    \label{tab:main_res}}
    \vspace{2mm}
    \begin{tabular}{ccccccccc}
    \hline
     Dataset & Target & Method & CLIP $\uparrow$ & M-CLIP $\uparrow$ & D-CLIP $\downarrow$& FID $\downarrow$& PSNR $\uparrow$& SSIM $\uparrow$\\
    \hline
        \multirow{9}{*}{Ego4D} & \multirow{5}{*}{$f_\text{action}$} & StableDiffusion \cite{rombach2022highresolutionimagesynthesislatent} & 53.75 & 64.84 & 41.57  & 162.47 & 28.33 & 37.03\\
		&& InstructPix2Pix \cite{brooks2023instructpix2pix} & 56.95 & 64.72 & 38.11  & 137.57 & 28.20 & 38.12 \\
        && GenHowTo \cite{souček2024genhowto} & 69.36 & \textbf{70.85} & 24.70  & \underline{67.47} & \underline{28.43} & \underline{38.19} \\
        && LEGO \cite{lai2024lego} & \underline{73.39} & \underline{70.48} & \underline{20.32}  & 87.71 & 28.09 & 36.33 \\
        && \cellcolor{gray!20} \methodname & \cellcolor{gray!20} \textbf{82.87} & \cellcolor{gray!20} 68.69 & \cellcolor{gray!20} \textbf{10.11}  & \cellcolor{gray!20} \textbf{55.14} & \cellcolor{gray!20} \textbf{29.08 }&\cellcolor{gray!20} \textbf{43.07}\\\cdashline{2-9}

        & \multirow{4}{*}{$f_\text{final}$} & StableDiffusion \cite{rombach2022highresolutionimagesynthesislatent} & 51.17 & 62.96 & 41.33  & 162.84 & \underline{28.20} & 32.59\\
		&& InstructPix2Pix \cite{brooks2023instructpix2pix} & 52.71 & 61.70 & 39.55 & 151.10 & 28.03 & \underline{33.26} \\
		&& GenHowTo \cite{souček2024genhowto} & \underline{59.36} & \underline{64.94} & \underline{32.07}  & \underline{85.44} & 28.07 & 31.01\\
        && \cellcolor{gray!20} \methodname & \cellcolor{gray!20} \textbf{79.99} & \cellcolor{gray!20} \textbf{68.15} & \cellcolor{gray!20} \textbf{8.69}  & \cellcolor{gray!20} \textbf{57.98} & \cellcolor{gray!20} \textbf{28.50} & \cellcolor{gray!20} \textbf{36.71} \\\hline

		\multirow{8}{*}{EGTEA Gaze+} &\multirow{4}{*}{$f_\text{action}$} &  StableDiffusion \cite{rombach2022highresolutionimagesynthesislatent} & 46.32 & 63.94 & 40.89  & 139.50 & 28.03 & 36.53\\
        && InstructPix2Pix \cite{brooks2023instructpix2pix} & 50.07 & 64.35 & 35.21  & 144.01 &  28.00 & \textbf{42.71}\\
		&& GenHowTo \cite{souček2024genhowto} & 61.89 & \underline{71.86} & 22.13  & \underline{82.53} & \underline{28.14} & 39.22 \\
        && \cellcolor{gray!20}\methodname & \cellcolor{gray!20}\textbf{69.62} & \cellcolor{gray!20} 66.34 & \cellcolor{gray!20}
        \textbf{12.44}  & \cellcolor{gray!20} \textbf{78.78} & \cellcolor{gray!20} \textbf{28.26} & \cellcolor{gray!20} 40.48\\\cdashline{2-9}
        
        & \multirow{4}{*}{$f_\text{final}$} & StableDiffusion \cite{rombach2022highresolutionimagesynthesislatent} & 42.61 & 63.10 & 44.05  & 151.00 & \underline{28.01} & 35.80 \\
		&& InstructPix2Pix \cite{brooks2023instructpix2pix} & 47.20 & 63.20 & 38.09  & 153.49 & 27.96 & \textbf{43.14}\\
		&& GenHowTo \cite{souček2024genhowto} & \underline{48.95} & \underline{64.49} & \underline{36.09}  & \underline{109.08} & 27.98 & 37.57 \\
        && \cellcolor{gray!20} \methodname & \cellcolor{gray!20} \textbf{68.44} & \cellcolor{gray!20} \textbf{65.91} & \cellcolor{gray!20} \textbf{11.60}  & \cellcolor{gray!20} \textbf{79.26} & \cellcolor{gray!20} \textbf{28.19} & \cellcolor{gray!20} \underline{40.37} \\\hline

        \multirow{9}{*}{EPIC-KITCHENS} & \multirow{5}{*}{$f_\text{action}$} & StableDiffusion \cite{rombach2022highresolutionimagesynthesislatent} & 27.49 & 49.49 & 67.08  & 389.27
        & 27.91 & 7.32 \\
		&& InstructPix2Pix \cite{brooks2023instructpix2pix} & 49.14 & 63.81 & 42.17 &  125.94 & 28.03 &\textbf{28.73}  \\
		&& GenHowTo \cite{souček2024genhowto} & 62.51 & \underline{70.90}  & 26.61  & \textbf{72.66} & \underline{28.19}  & 27.46 \\
        && LEGO \cite{lai2024lego} & \underline{66.33} & \textbf{72.65} & \underline{21.81} & 98.59 & 28.02 & 25.83 \\
        && \cellcolor{gray!20} \methodname & \cellcolor{gray!20} \textbf{69.97} & \cellcolor{gray!20} 65.02 & \cellcolor{gray!20} \textbf{18.08}  & \cellcolor{gray!20} \underline{94.33} & \cellcolor{gray!20} \textbf{28.36} & \cellcolor{gray!20} \underline{28.52}
        \\\cdashline{2-9}

        & \multirow{4}{*}{$f_\text{final}$} & StableDiffusion \cite{rombach2022highresolutionimagesynthesislatent} & 27.96 & 49.69 & 65.34 & 387.22 & 27.90 & 6.71 \\
		&& InstructPix2Pix \cite{brooks2023instructpix2pix} & 47.39 & 62.75 & 41.87  & 132.13 & 27.97 & \textbf{27.48}\\
		&& GenHowTo \cite{souček2024genhowto} & \underline{51.37} & \underline{63.63} & \underline{36.65}  & \textbf{95.02} & \underline{27.99} & 23.87 \\
        && \cellcolor{gray!20} \methodname & \cellcolor{gray!20} \textbf{70.06} & \cellcolor{gray!20}
        \textbf{66.39} & \cellcolor{gray!20}\textbf{14.72} & \cellcolor{gray!20} \underline{99.12} & \cellcolor{gray!20} \textbf{28.25} & \cellcolor{gray!20} \underline{27.36}\\
    \hline
    \end{tabular}
    \vspace{-7mm}
\end{table}

\subsection{Comparison with state-of-the-art}
\label{sec:main_results}

Table \ref{tab:main_res} compares baseline models and our method on Ego4D, EGTEA Gaze+, and EK-100 datasets, evaluating their ability to generate accurate \textit{action} and \textit{final} \textit{frames}. LEGO~\cite{lai2024lego}'s goal is to generate action frames only, we report its results only for that. We do not report LEGO results on EGTEA Gaze+, because it only provides pre-trained weights for Ego4D and EK-100.

CLIP and M-CLIP scores reflect semantic similarity to reference images, with higher values indicating closer alignment. D-CLIP captures semantic discrepancy between images (lower is better). 
The evaluation is performed on the test sets: 1,000 triplets from Ego4D, 650 from EGTEA Gaze+, and 750 from EK-100.

First of all, the results exhibit consistent behavior across all datasets, especially EGTEA Gaze+ and EK-100, indicating that \methodname generalizes well to egocentric cooking data and consistently produces high-quality results.

\methodname consistently achieves the highest CLIP and lowest D-CLIP scores, indicating strong semantic and contextual alignment.  Occasionally, it achieves a higher CLIP score, but a lower M-CLIP score compared to LEGO or GenHowTo, suggesting they may better refine masked regions but introduce unwanted changes elsewhere.
\methodname’s SSIM values are lower than other models like InstructPix2Pix, indicating that while it is semantically accurate, its structural quality could be marginally improved. 
StableDiffusion and InstructPix2Pix show competitive SSIM values but are generally outperformed by \methodname and GenHowTo in content relevance. At the same time,  \methodname achieves the highest PSNR scores and among the best FID results. Overall, \methodname emerges as the top-performing model across semantic metrics. 

\begin{figure}[htb!]
    \centering
    \footnotesize
    \begin{tabular}{ccc@{\hskip 2pt}c@{\hskip 2pt}c@{\hskip 2pt}cc@{\hskip 2pt}c@{\hskip 2pt}c}
         & Input & \multicolumn{4}{c}{Action frame} & \multicolumn{3}{c}{Final frame}\\
        &  & IP2P & GHT & LEGO & VC & IP2P & GHT & VC\\

        \multirow{4}{*}{\rotatebox{90}{Ego4D}}
        & \includegraphics[width=0.1\textwidth]{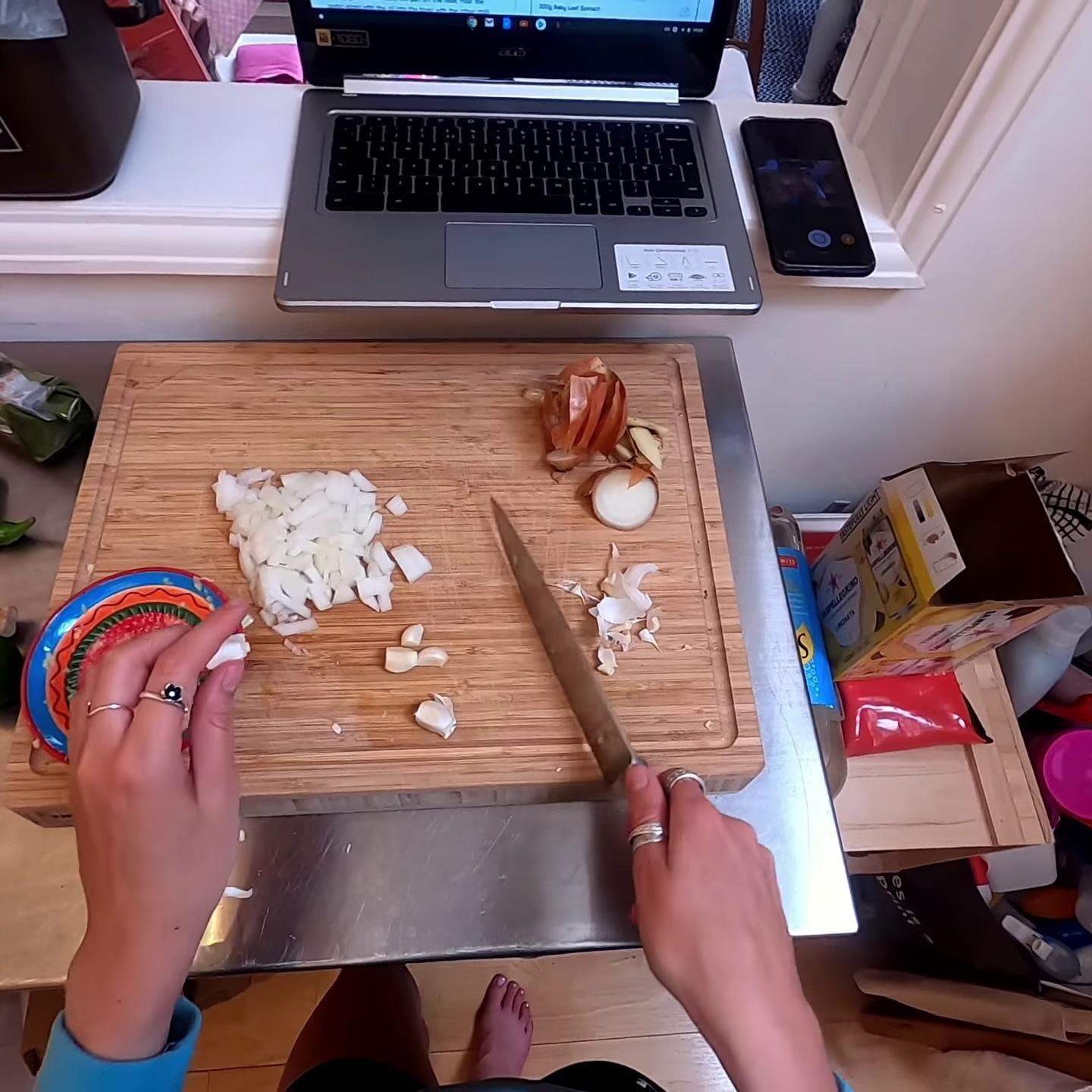}  & \includegraphics[width=0.1\textwidth]{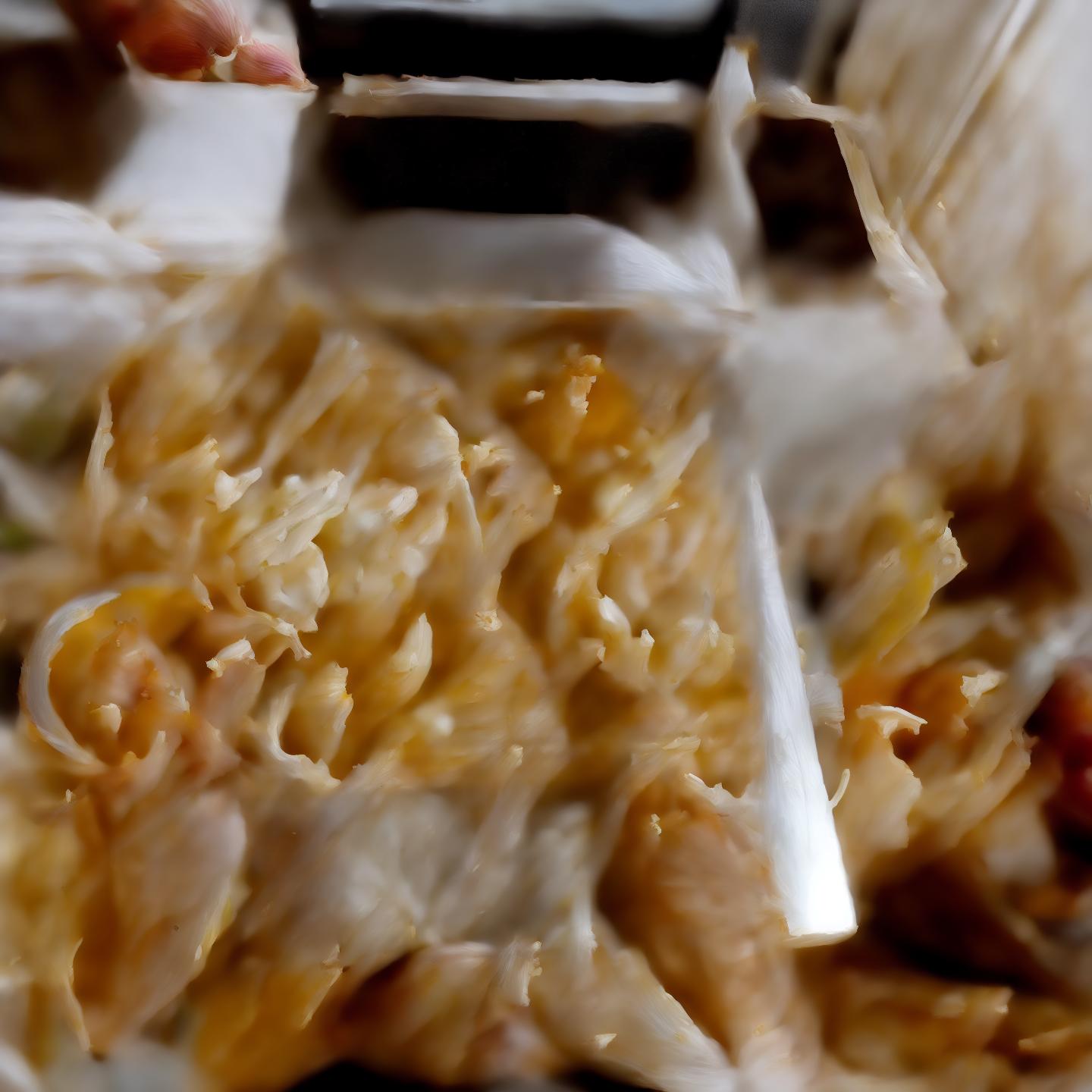} & \includegraphics[width=0.1\textwidth]{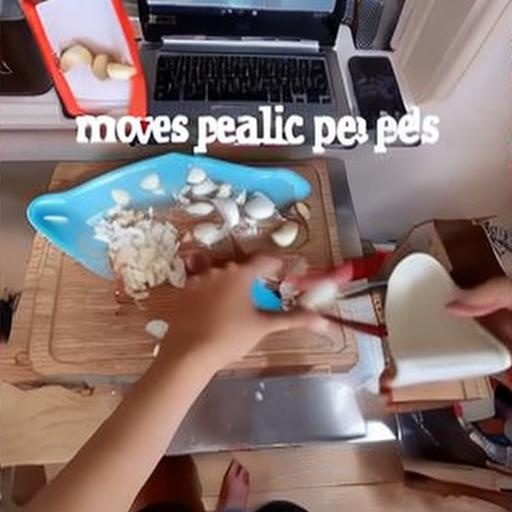}& \includegraphics[width=0.1\textwidth]{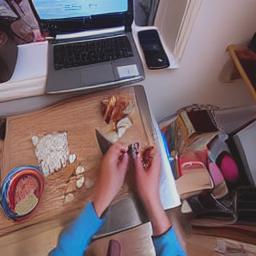} & \includegraphics[width=0.1\textwidth]{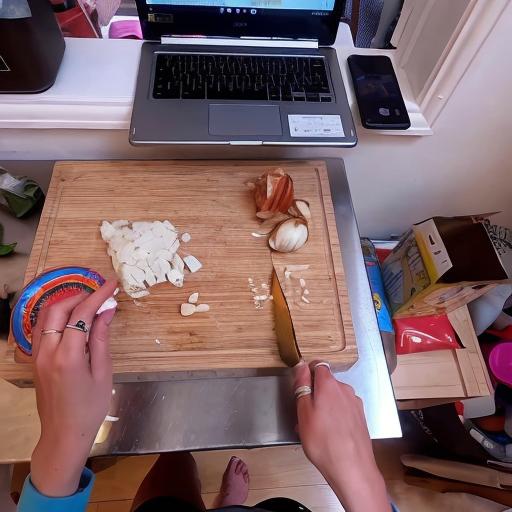} & \includegraphics[width=0.1\textwidth]{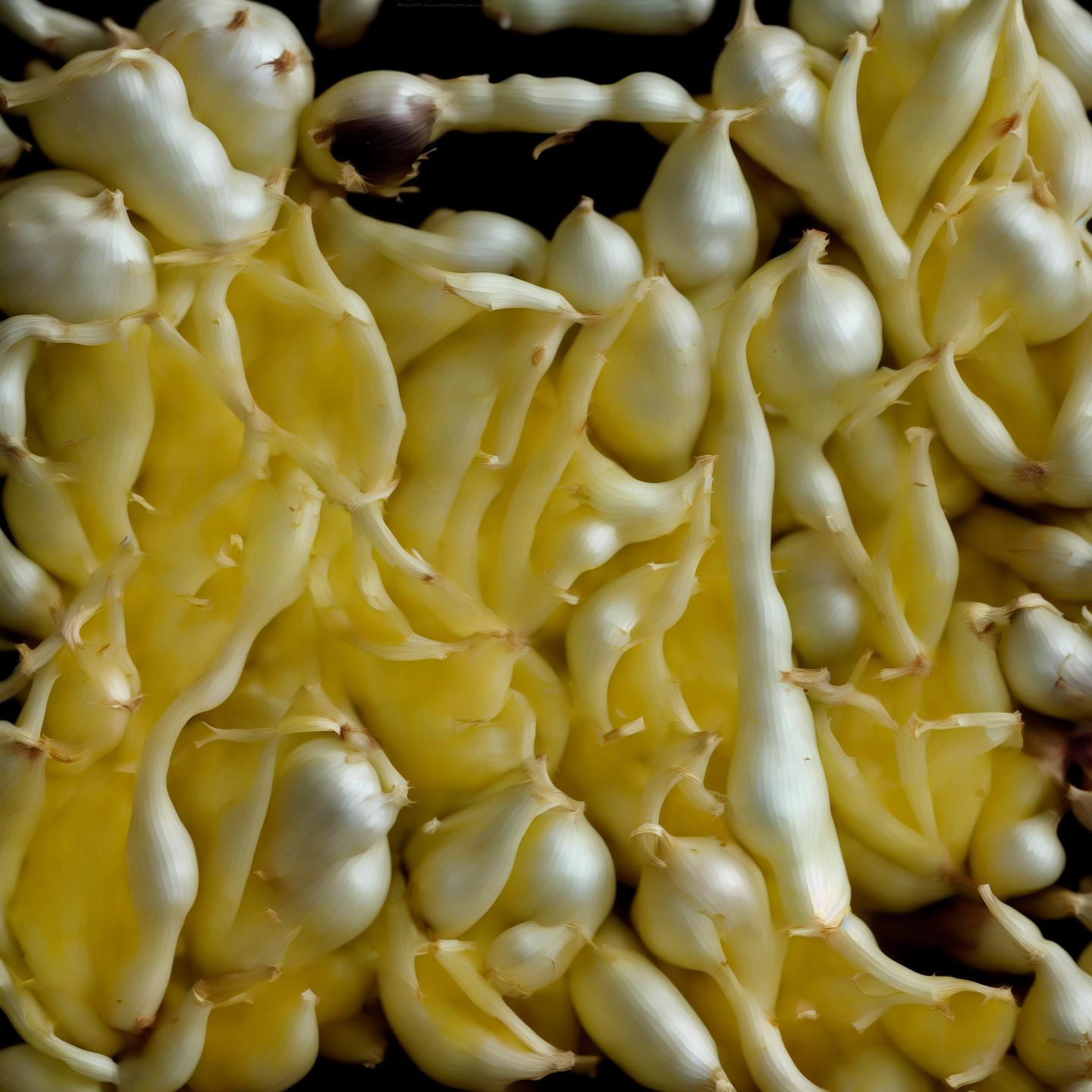} & \includegraphics[width=0.1\textwidth]{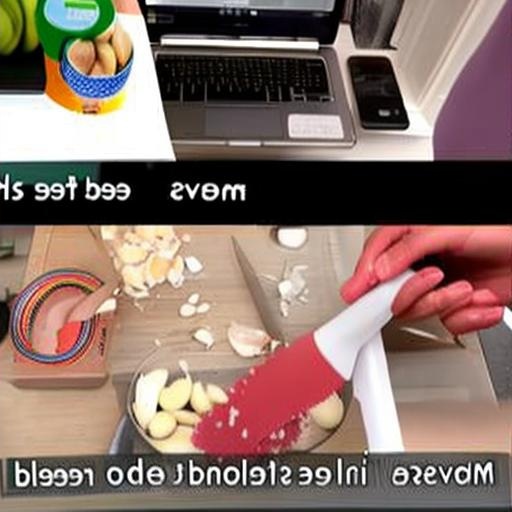} & \includegraphics[width=0.1\textwidth]{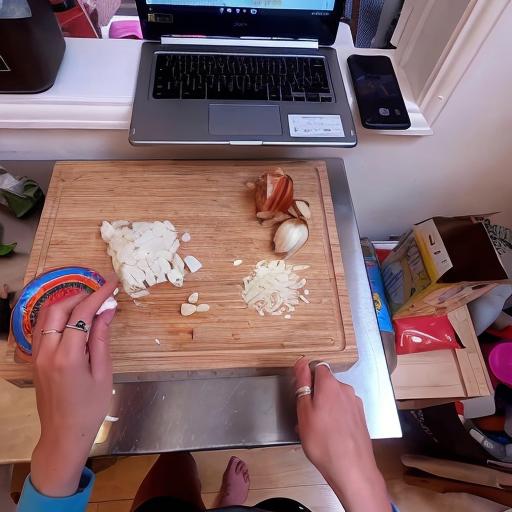} \\
        & Peel garlic \\

        & \includegraphics[width=0.1\textwidth]{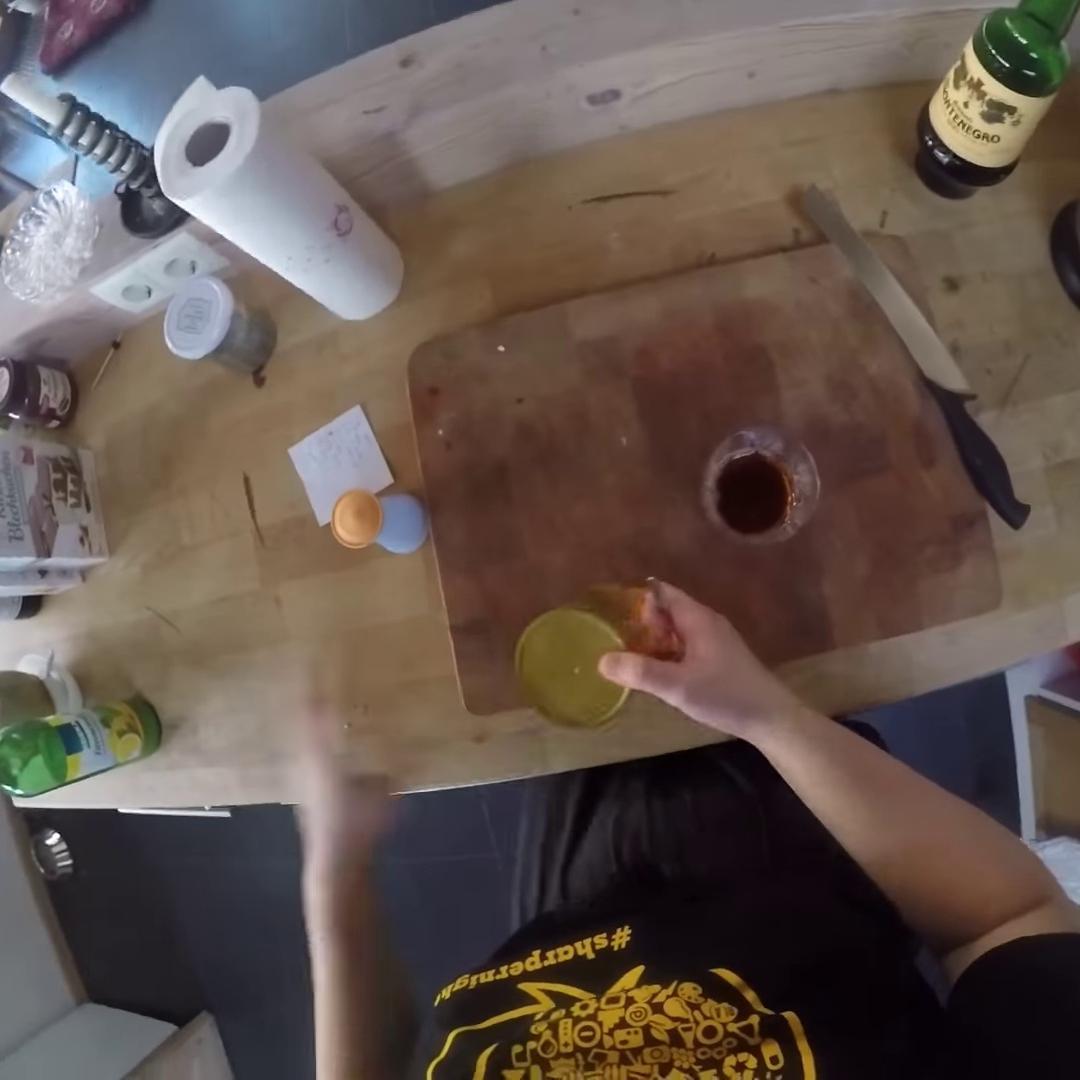} & \includegraphics[width=0.1\textwidth]{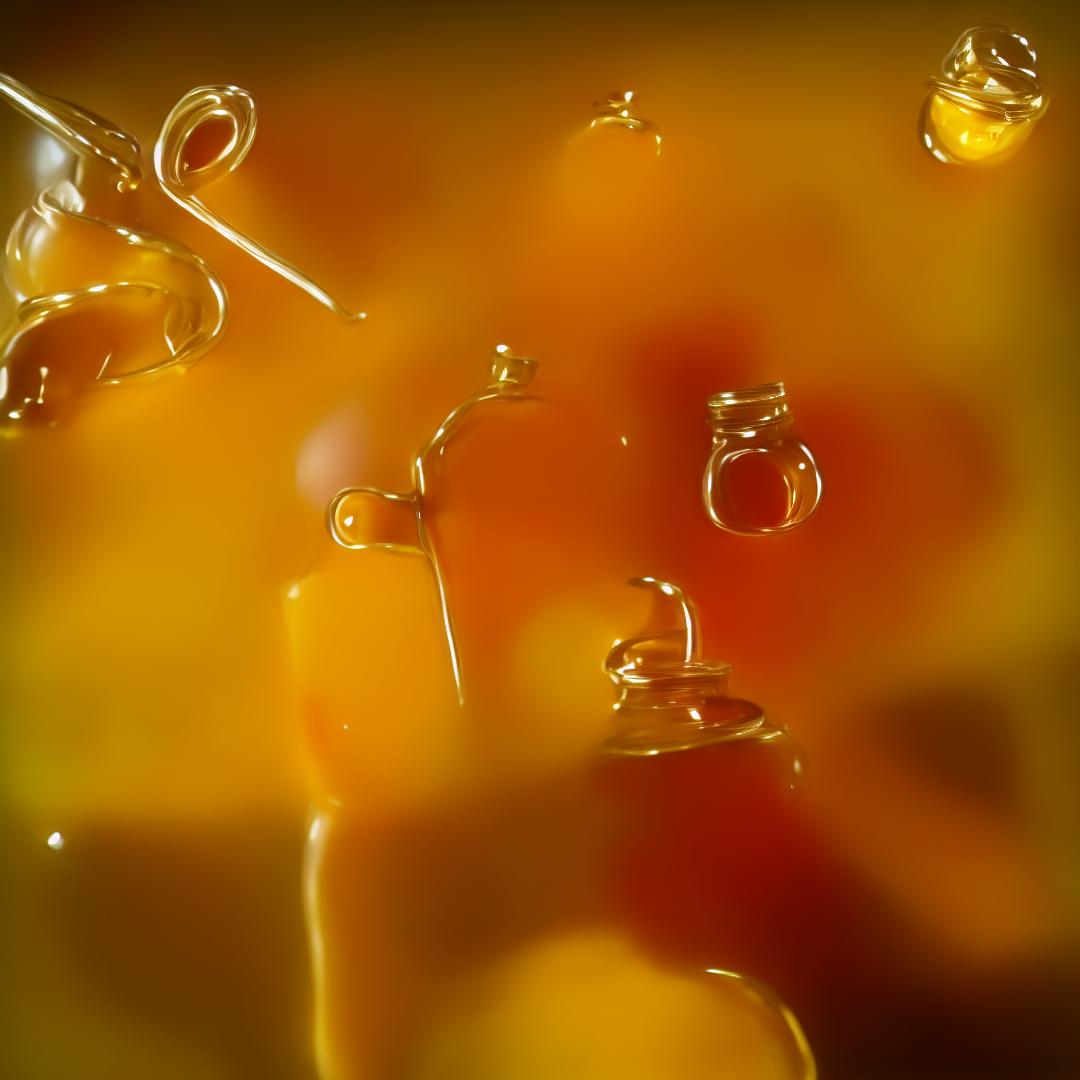} & \includegraphics[width=0.1\textwidth]{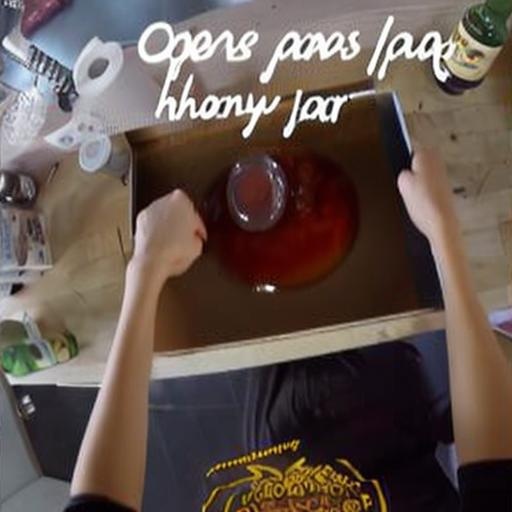}& \includegraphics[width=0.1\textwidth]{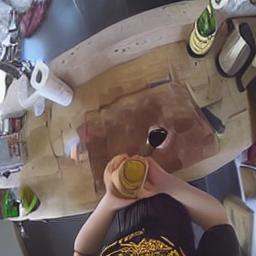} & \includegraphics[width=0.1\textwidth]{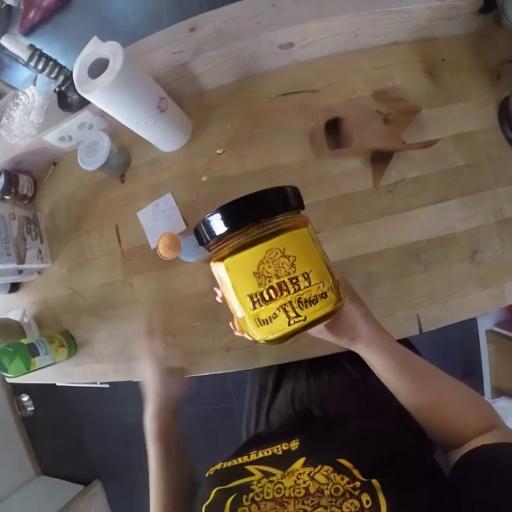} & \includegraphics[width=0.1\textwidth]{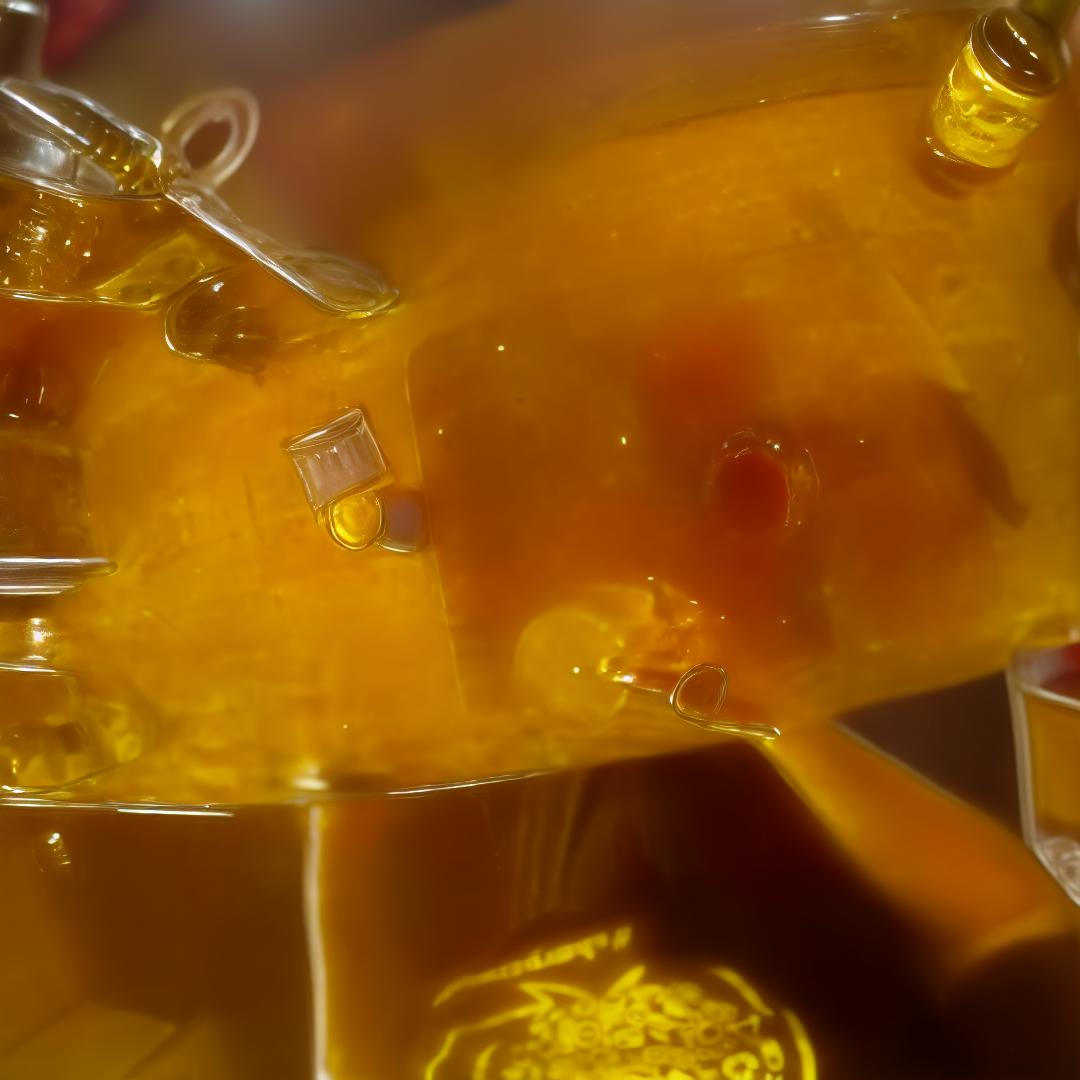} & \includegraphics[width=0.1\textwidth]{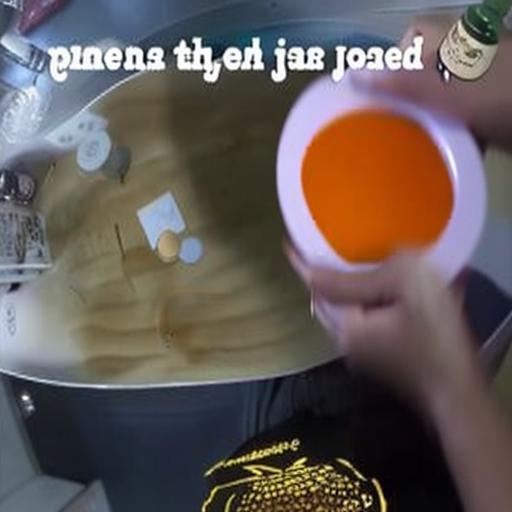} & \includegraphics[width=0.1\textwidth]{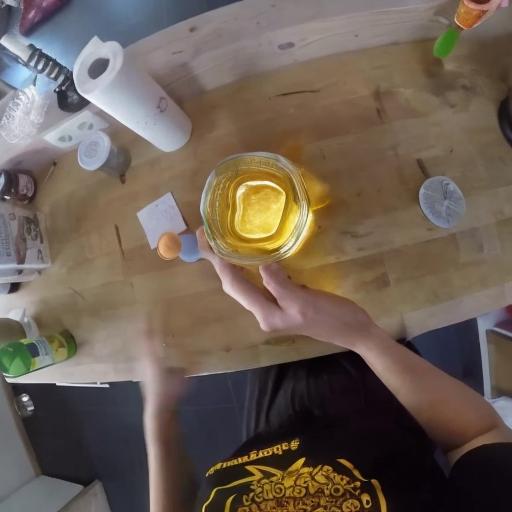} \\

        & Open honey jar \\
        
        \multirow{4}{*}{\rotatebox{90}{EGTEA Gaze+}}
        & \includegraphics[width=0.1\textwidth]{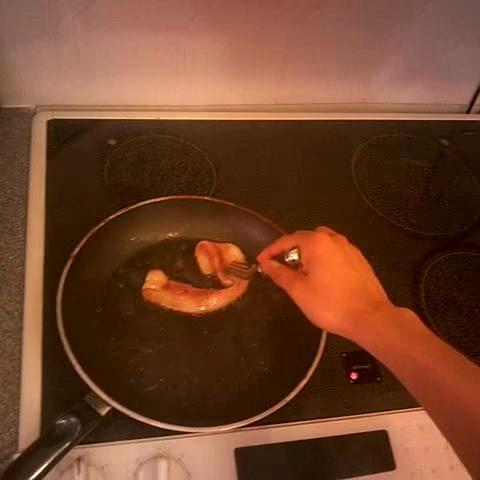} & \includegraphics[width=0.1\textwidth]{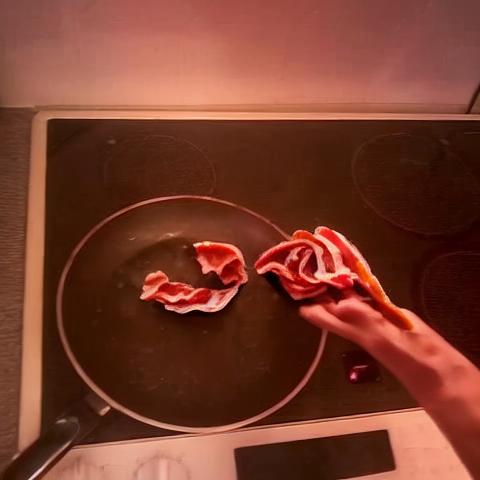} & \includegraphics[width=0.1\textwidth]{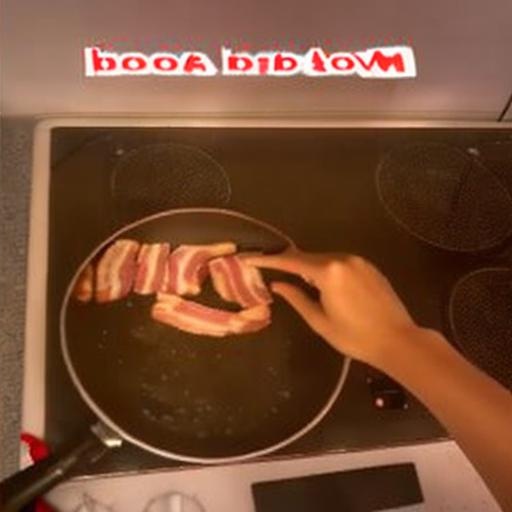}& \includegraphics[width=0.1\textwidth]{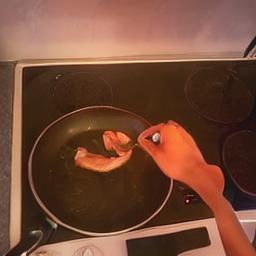} & \includegraphics[width=0.1\textwidth]{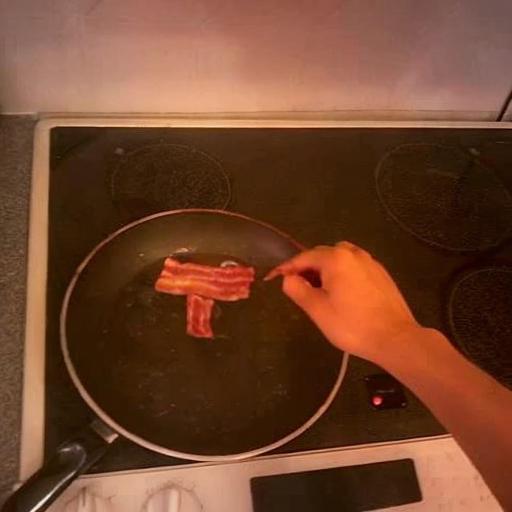} & \includegraphics[width=0.1\textwidth]{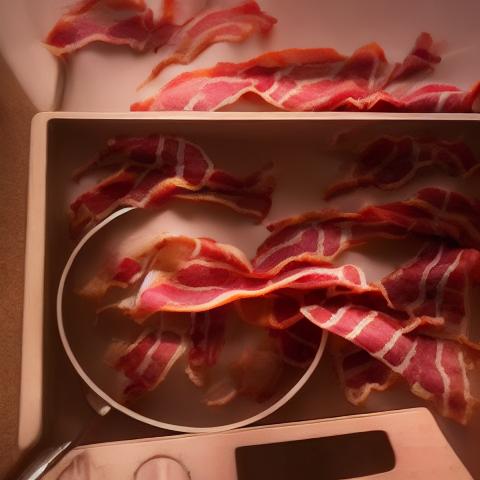} & \includegraphics[width=0.1\textwidth]{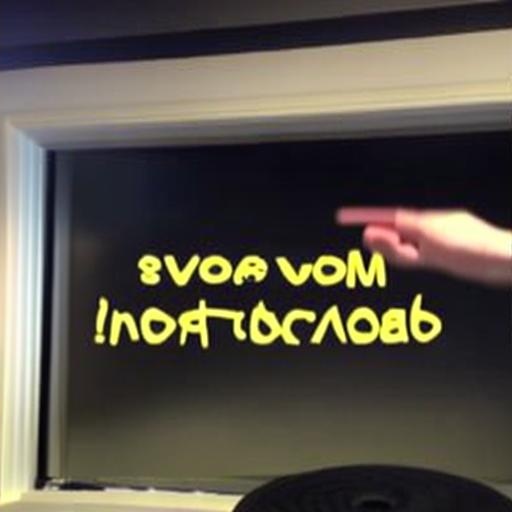} & \includegraphics[width=0.1\textwidth]{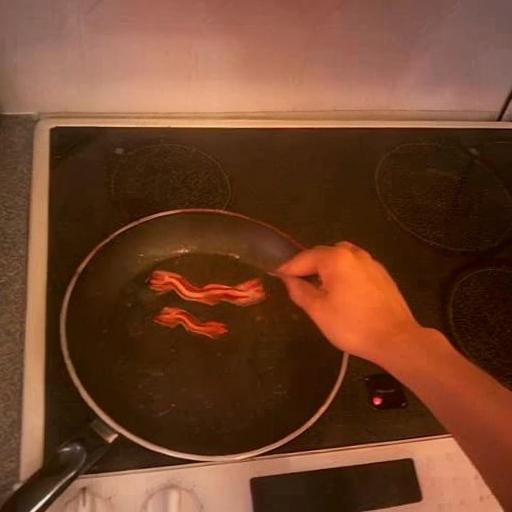} \\

        & Move bacon \\

        & \includegraphics[width=0.1\textwidth]{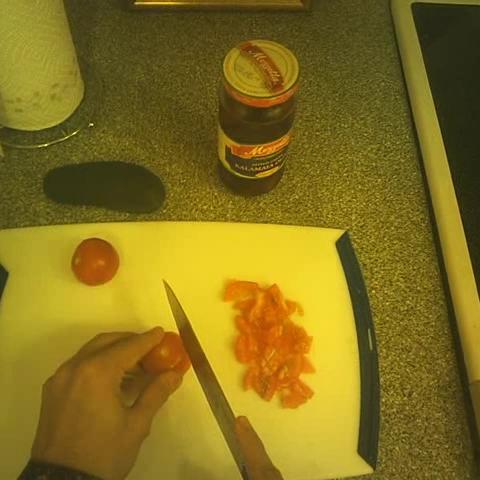} & \includegraphics[width=0.1\textwidth]{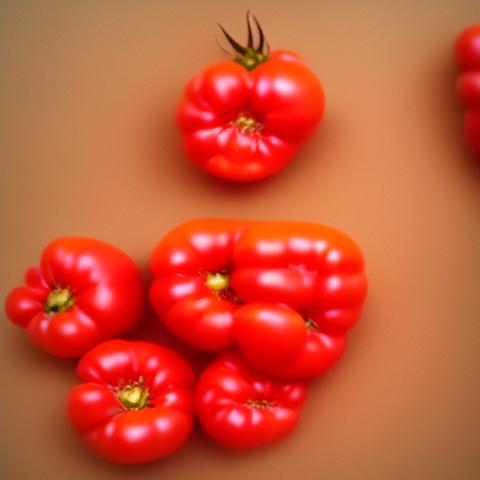} & \includegraphics[width=0.1\textwidth]{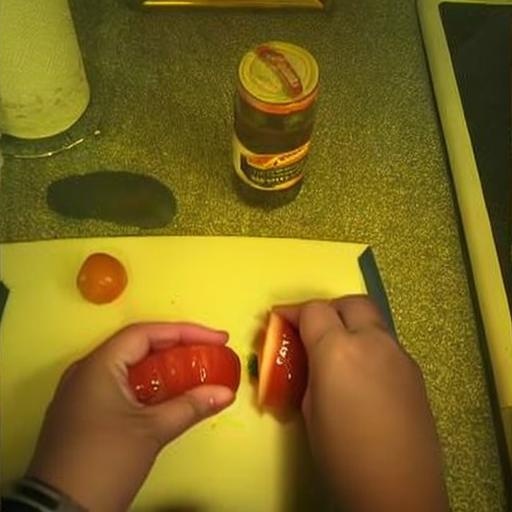}& \includegraphics[width=0.1\textwidth]{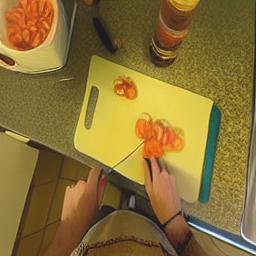} & \includegraphics[width=0.1\textwidth]{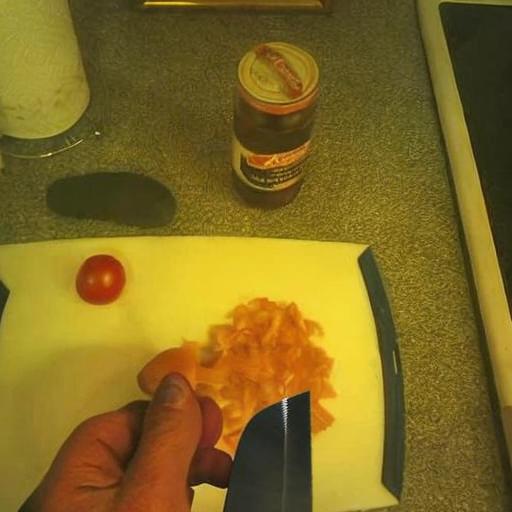} & \includegraphics[width=0.1\textwidth]{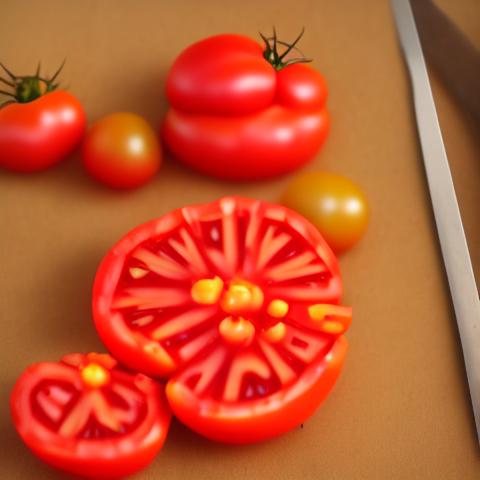} & \includegraphics[width=0.1\textwidth]{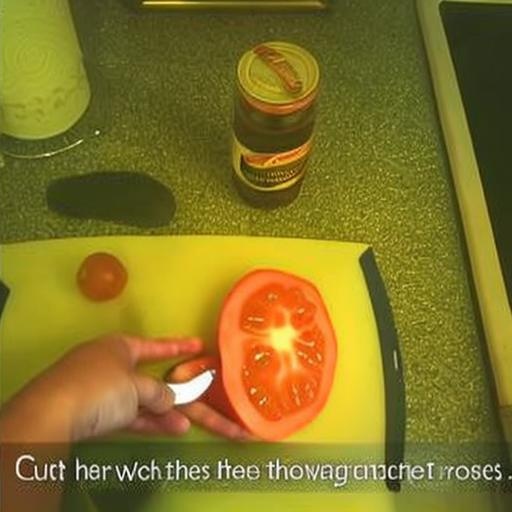} & \includegraphics[width=0.1\textwidth]{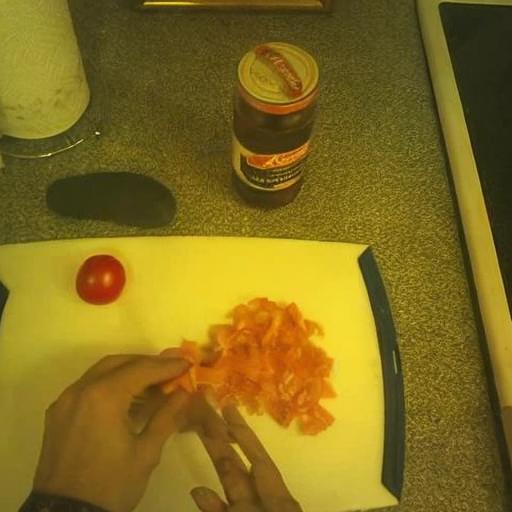} \\ 
        & Cut tomato \\

        \multirow{4}{*}{\rotatebox{90}{EK-100}}
        & \includegraphics[width=0.1\textwidth]{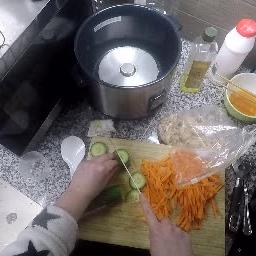} & \includegraphics[width=0.1\textwidth]{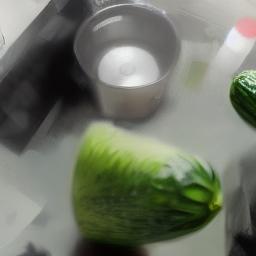} & \includegraphics[width=0.1\textwidth]{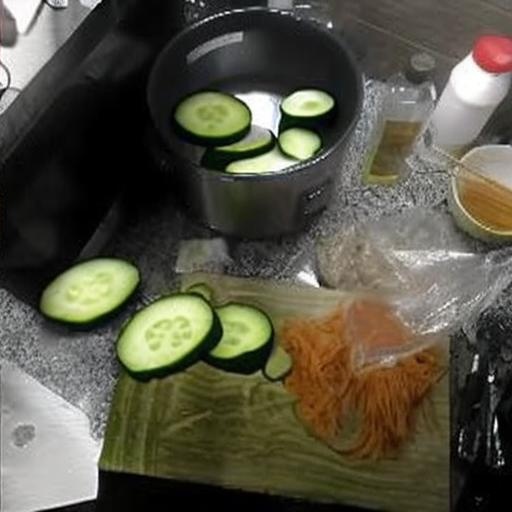}& \includegraphics[width=0.1\textwidth]{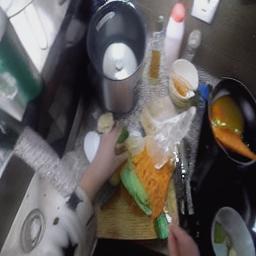} & \includegraphics[width=0.1\textwidth]{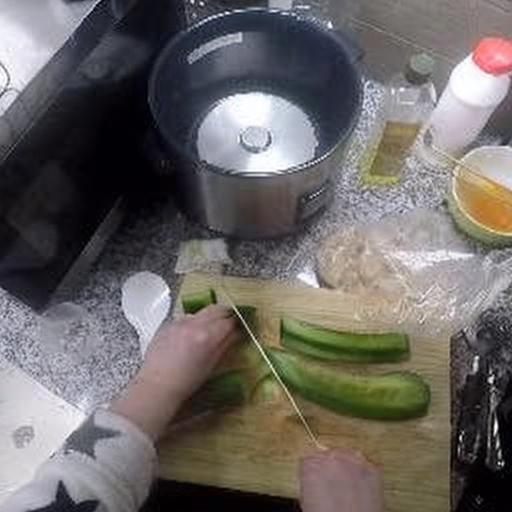} & \includegraphics[width=0.1\textwidth]{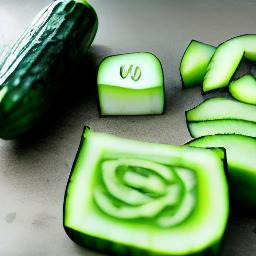} & \includegraphics[width=0.1\textwidth]{images/test/epic-1-final-genhowto.jpg} & \includegraphics[width=0.1\textwidth]{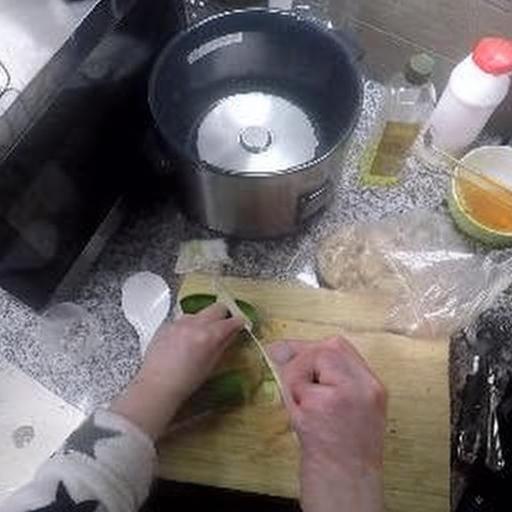} \\
        & Slice cucumber    \\
        & \includegraphics[width=0.1\textwidth]{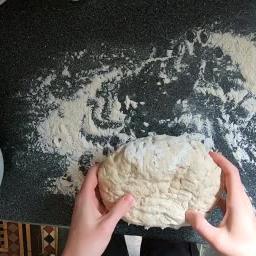} & \includegraphics[width=0.1\textwidth]{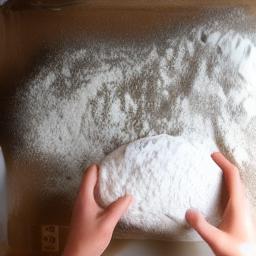} & \includegraphics[width=0.1\textwidth]{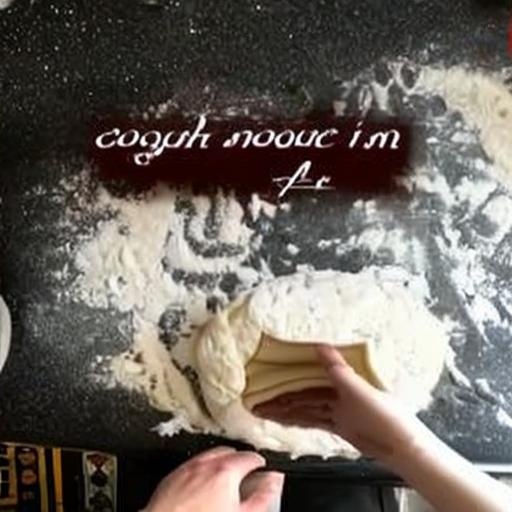}& \includegraphics[width=0.1\textwidth]{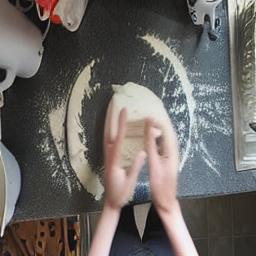} & \includegraphics[width=0.1\textwidth]{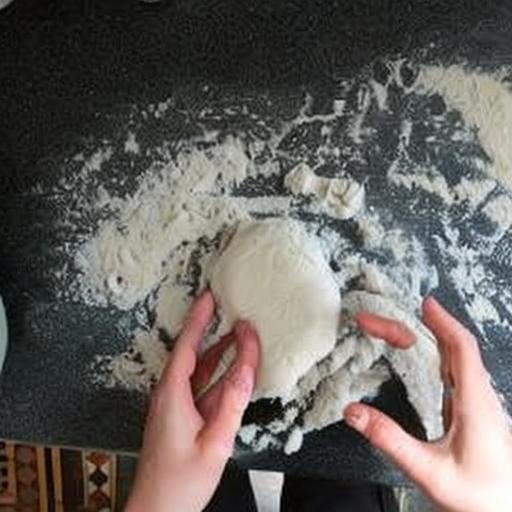} & \includegraphics[width=0.1\textwidth]{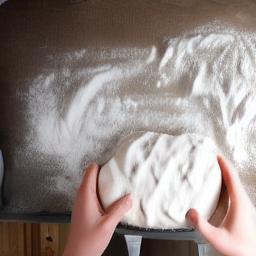} & \includegraphics[width=0.1\textwidth]{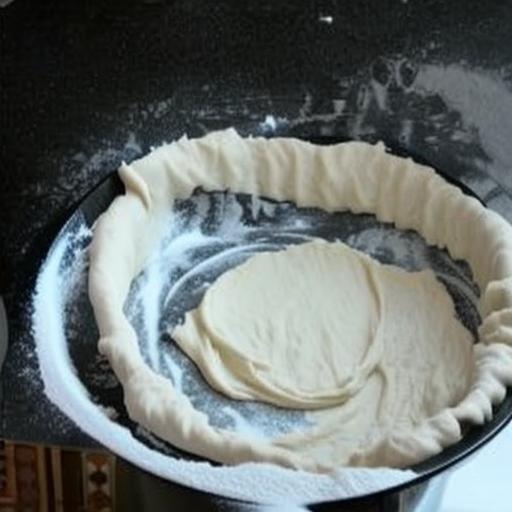} & \includegraphics[width=0.1\textwidth]{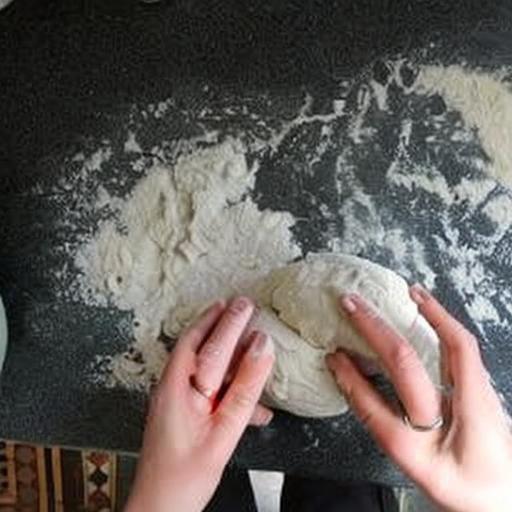} \\
        & Cover dough in flour \\

    \end{tabular}
    \vspace{-0.7em}
    \caption{\textbf{Qualitative comparison with related work.} \methodname has the best performance in aligning generated images to the input action and preserving the environment compared to state-of-the-art methods.}
    \label{fig:qualitative-comparison}
    \vspace{-5mm}
\end{figure}

Figure \ref{fig:qualitative-comparison} shows the generation results from \methodname and SOTA methods across three datasets. InstructPix2Pix and GenHowTo often fail to maintain scene consistency. LEGO exhibits a stronger capability for preserving scene consistency, but it occasionally introduces deformations. LEGO does not generate $f_{\text{final}}$, which positions our model as the most complete and effective approach.

Since automatic evaluation of egocentric video generation is not a mature field, we conducted a binary comparison study with 30 users. Each participant viewed 25 image pairs (one generated by \methodname and the other by either LEGO or GenHowTo) with the action and initial image. The users selected the best image that depicts the action or its result in terms of accuracy, quality, and clarity. Figure \ref{fig:human} shows that \methodname was preferred in 70\% of action frames and 80\% of final frames. While LEGO often depicts the action well, object deformations reduce visual appeal. GenHowTo shows more variable quality ranging from high to poor, which led to higher average preference over LEGO.

\subsection{Generation of Various Actions}

Figure~\ref{fig:generalization} shows an example where the same initial frame is used to generate different actions. To qualitatively assess generalization, we used a custom image as the input. The resulting final frames accurately reflect the intended actions while preserving the scene, highlighting \methodname’s practical applicability.

\subsection{Ablations}
\label{sec:ablations} 

\textbf{Effect of fine-tuning} Table \ref{tab:finetuning} shows that fine-tuning the inpainting model on a randomly selected subset of EGTEA Gaze+ improves the quality of generated frames compared to using the pretrained model.

\noindent\textbf{Cross-dataset generalization} Both \methodname and LEGO~\cite{lai2024lego} use dataset-specific weights. To test generalization, we generated images on the EGTEA Gaze+ evaluation set using weights trained on Ego4D. As shown in Table~\ref{tab:generalization}, \methodname generalizes better than LEGO.

\begin{table}[t]
    \centering
    \scriptsize
    \caption{\textbf{Ablation study.} (a) We analyze the effect of fine-tuning the diffusion pipeline. Higher scores for the fine-tuned model indicate a positive training impact. (b) We evaluate cross-dataset generalization by applying Ego4D-trained weights of \methodname and LEGO to the EGTEA Gaze+ test set. \methodname consistently outperforms LEGO.}
    \vspace{-2mm}
    \begin{subtable}[t]{0.48\linewidth}
        \centering
        \caption{Effect of fine-tuning}
        \label{tab:finetuning}
        \begin{tabular}{ccccc}
            \toprule
            \multirow{2}{*}{Method} & \multicolumn{2}{c}{Action} & \multicolumn{2}{c}{Final} \\
            & CLIP $\uparrow$ & D-CLIP $\downarrow$ & CLIP $\uparrow$ & D-CLIP $\downarrow$ \\
            \midrule
            Inpainting & 70.65 & 16.58 & 71.00 & 13.80 \\
            VisualChef & \textbf{71.33} & \textbf{15.72} & \textbf{71.07} & \textbf{13.75} \\
            \bottomrule
        \end{tabular}
    \end{subtable}
    \hfill
    \begin{subtable}[t]{0.48\linewidth}
        \centering
        \caption{Cross-dataset generalization}
        \label{tab:generalization}
        \begin{tabular}{cccc}
            \toprule
            Method & CLIP $\uparrow$ & M-CLIP $\uparrow$ & D-CLIP $\downarrow$ \\
            \midrule
            LEGO & 68.93 & \textbf{68.24} & 18.57 \\
            VisualChef & \textbf{71.22} & 63.66 & \textbf{15.96} \\
            \bottomrule
        \end{tabular}
    \end{subtable}
    \label{tab:ablations}
\end{table}

\noindent
\begin{minipage}[t]{0.48\linewidth}
  \centering
  \scriptsize
  \setlength{\abovecaptionskip}{2pt}
  \setlength{\belowcaptionskip}{2pt}
  \includegraphics{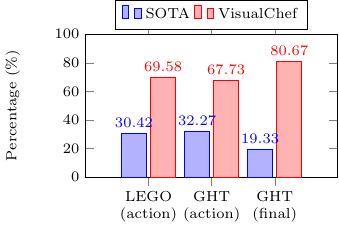}
  \captionof{figure}{\textbf{Human comparison of SOTA models and \methodname.} The users tend to select images generated by \methodname more often.}
  \label{fig:human}
\end{minipage}
\hfill
\begin{minipage}[t]{0.48\linewidth}
  \centering
  \scriptsize
  \vspace{-40mm}
  \begin{tabular}{cc}
     \includegraphics[width=0.3\linewidth]{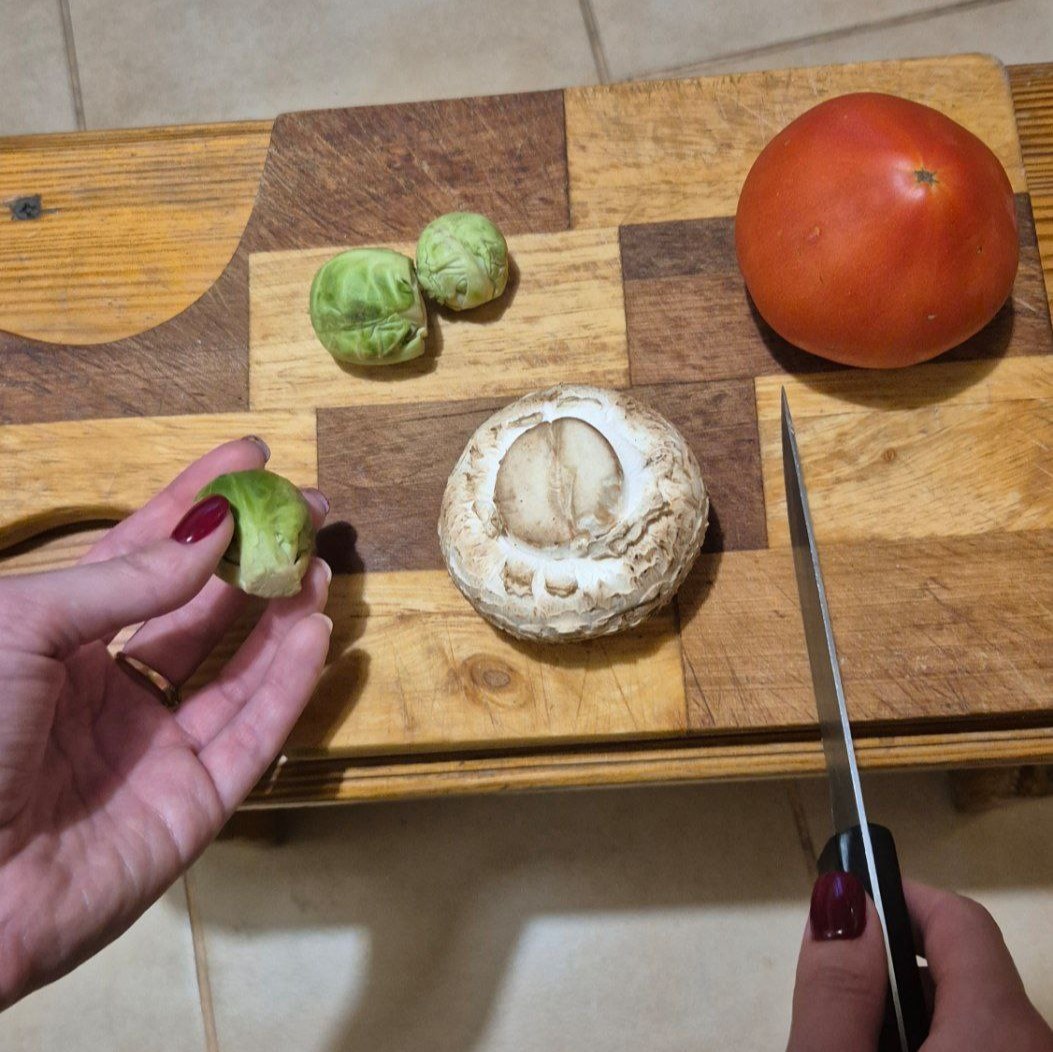}  & \includegraphics[width=0.3\linewidth]{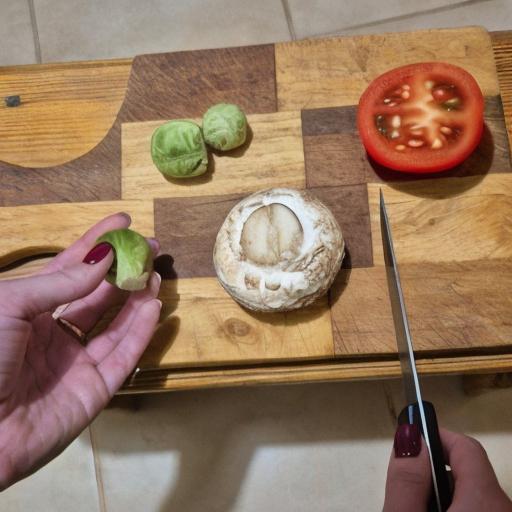} \\ 
     input & cut tomato \\ \includegraphics[width=0.3\linewidth]{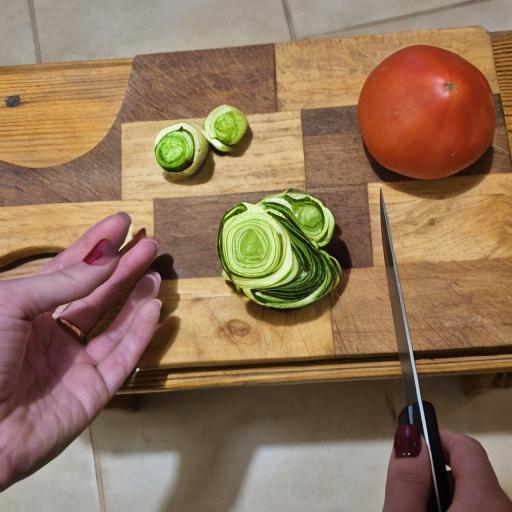} & \includegraphics[width=0.3\linewidth]{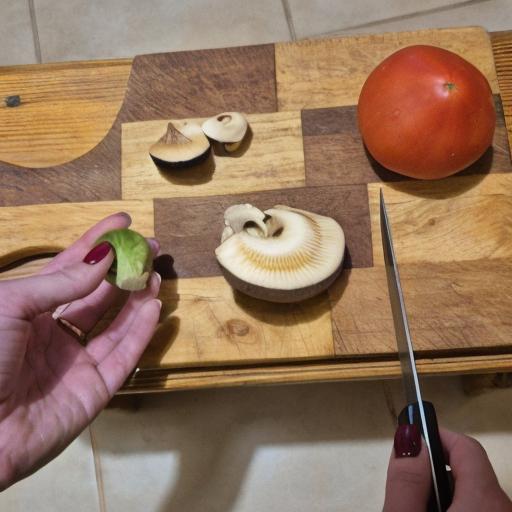} \\
      cut sprouts & cut mushroom
  \end{tabular}
  \vspace{-2mm}
  \captionof{figure}{\textbf{Different actions with the same input.} \methodname generalizes across actions while retaining context.}
  \label{fig:generalization}
\end{minipage}

%% file: sec/5_conclusions.tex
\section{Conclusion}

In this work, we present a novel approach for improving cooking instructions with automated visual aids that addresses the limitations of text-based recipes. Leveraging egocentric datasets like Ego4D, EGTEA Gaze+, and EK-100, our model effectively selects and generates frames that depict key stages of cooking actions. By using a diffusion-based pipeline with masked inpainting, we ensure precise edits that preserve scene consistency while focusing on core elements.

\noindent\textbf{Limitations.} In some cases, we observe a slight drop in the quality of generated action frames due to the difficulty of synthesizing plausible hand poses. To account for shape changes, we use bounding boxes for inpainting and to handle location changes, the relocation procedure; however, to handle complex shape changes, one might refer to a more complex model (\eg, \cite{sudhakar2024controllingworldsleighthand}).

Most actions in the datasets are atomic — for instance, instead of a single action like “put a tomato from the fridge onto the table,” we typically see a sequence of smaller steps: “open fridge,” “take tomato,” “put tomato on table,” and “close fridge.” As a result, cases where objects disappear due to viewpoint changes are rare. However, when they do occur, our model cannot handle them.

\label{sec:conclusion}

%% file: sec/X_suppl.tex
\clearpage
\setcounter{page}{1}

\newpage
\begin{center}
    {\LARGE \textbf{VisualChef: Generating Visual Aids in Cooking via Mask Inpainting}}\\[0.5em]
    Supplementary Material\\[1.0em]
\end{center}

\section{Examples of Intermediate Steps}

\subsection{Classifying Relevant Objects}

Figure~\ref{fig:llava-prompt-further} shows further examples of chain-of-thought reasoning with LLaVA~\cite{liu2023visualinstructiontuning} to get a categorized list of objects that are relevant to the action given the initial state.

\subsection{Multiple Location Objects}

\begin{figure}
    \centering
    \includegraphics[width=\linewidth]{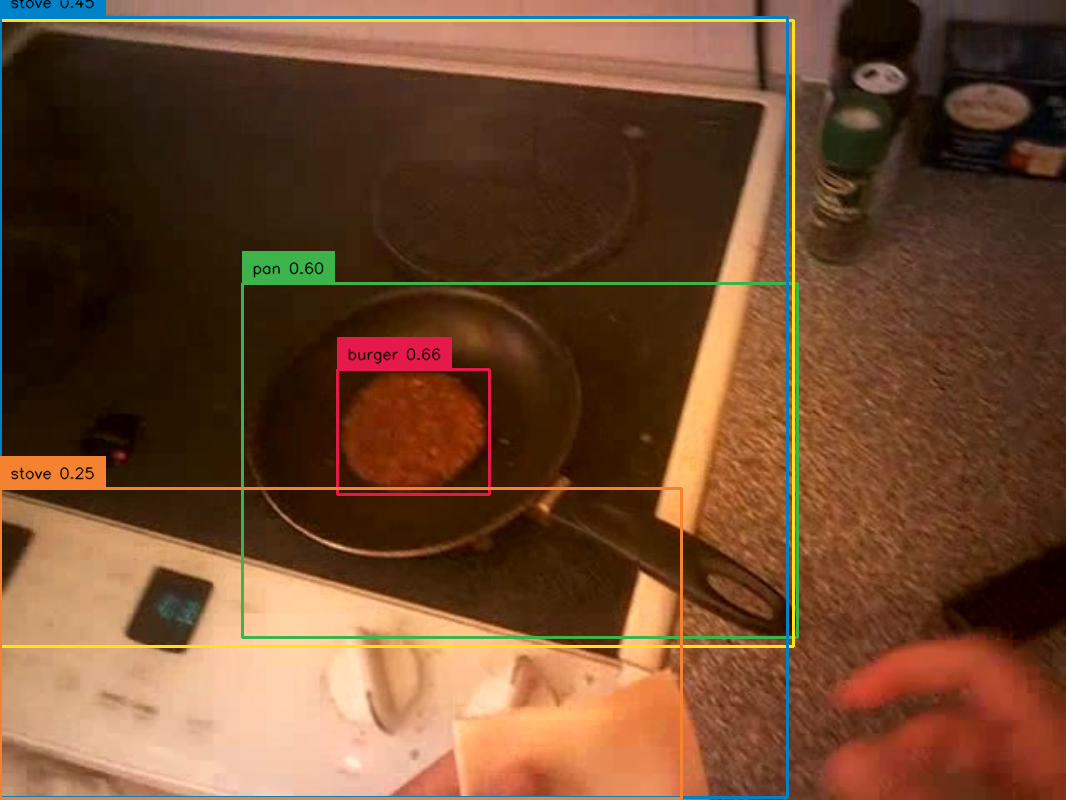}
    \caption{\textbf{Example of the dataset EGTEA Gaze+ where multiple location objects overlap.} The identified location objects ``stove", ``pan", and ``burger" cover the same subspace with ``burger" describing the destination as precisely as possible.}
    \label{fig:multiple-locations}
\end{figure}

\begin{figure*}
    \centering
    \includegraphics[width=\linewidth]{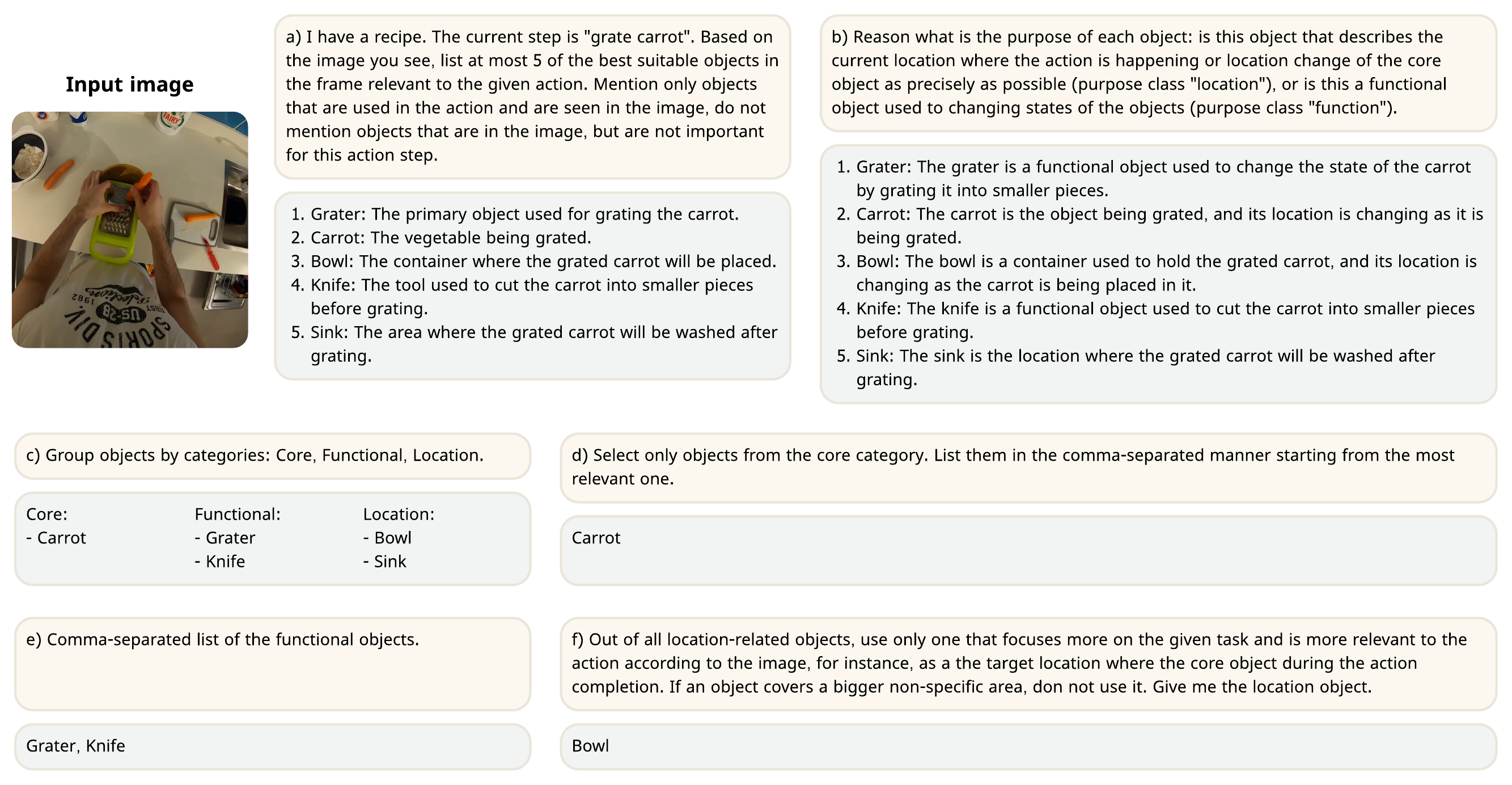}
    \includegraphics[width=\linewidth]{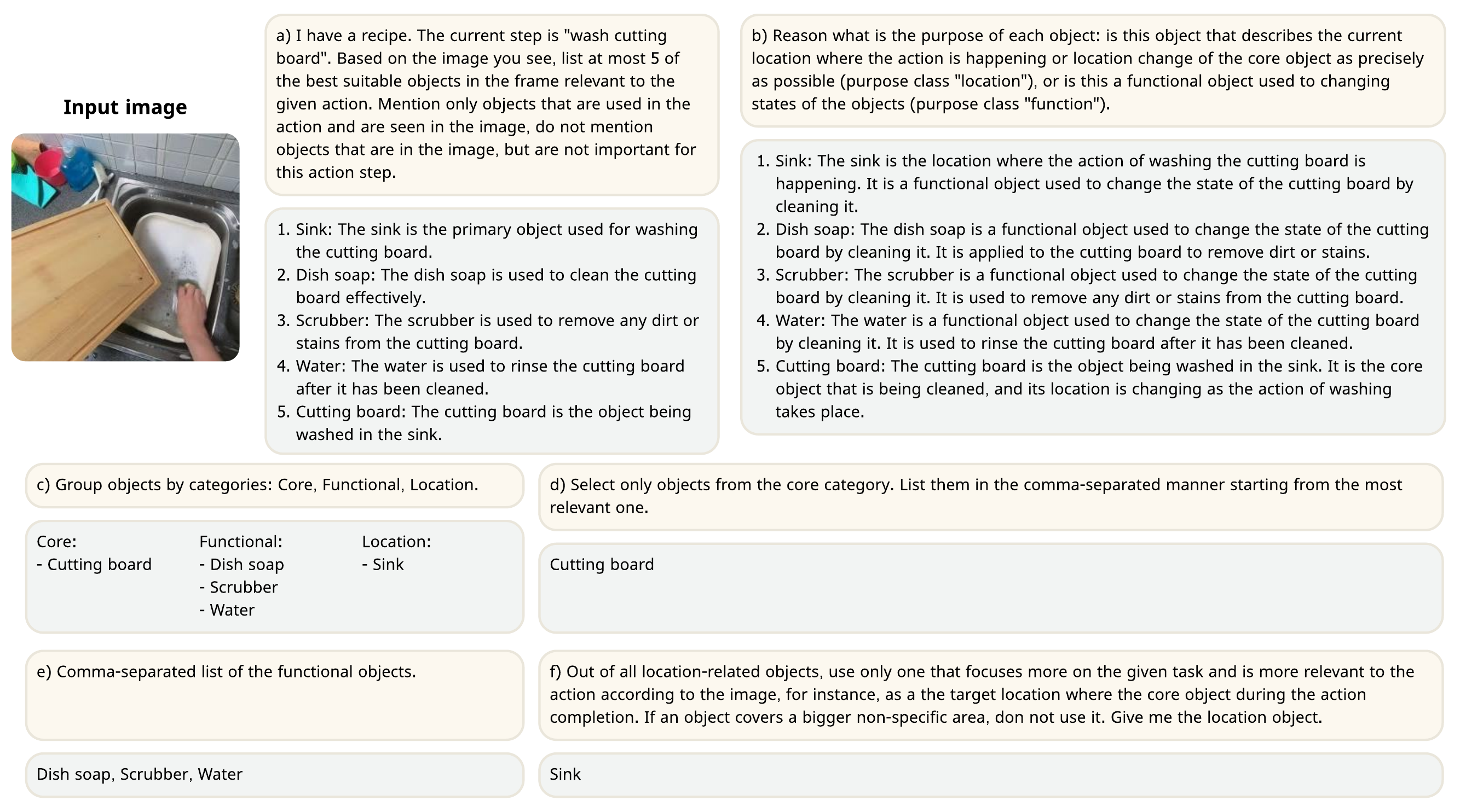}
    \caption{\textbf{Further examples of the chain-of-thoughts reasoning for relevant object identification.} Given an input image, we use the chain-of-thoughts strategy to prompt LLaVA~\cite{liu2023visualinstructiontuning} to get a categorized list of objects that are relevant to the given action.}
    \label{fig:llava-prompt-further}
    \vspace{-5mm}
\end{figure*}

As mentioned in \cref{ssec:llava}, there are cases where the VLM returns multiple \textit{location objects}, even though they often overlap and cover the same subspace of the image. As seen in \cref{fig:multiple-locations}, objects ``stove", ``pan", and ``burger" were produced by the VLM, and they cover the same subspace. However, we are interested in performing the action as precisely as possible and changing the environment as little as possible; therefore, using the same VLM (see \cref{fig:llava-prompt-further}f), we filter the \textit{location objects} to use only the one that describes the destination as precisely as possible.

\subsection{Visualization of the Filtering Process}

\begin{figure*}
    \centering
    \scriptsize
    \begin{tabular}{c@{\hskip 2pt}c@{\hskip 2pt}cc@{\hskip 2pt}c@{\hskip 2pt}cc@{\hskip 2pt}c}
        $f_\text{in}^\text{GT}$ & $f_\text{in}$ & $f_\text{in}$ det. &  $f_\text{action}^\text{GT}$ & $f_\text{action}$ & $f_\text{action}$ det. & $f_\text{final}^\text{GT}$ & 
        $f_\text{final}$  \\

        \includegraphics[width=0.11\textwidth]{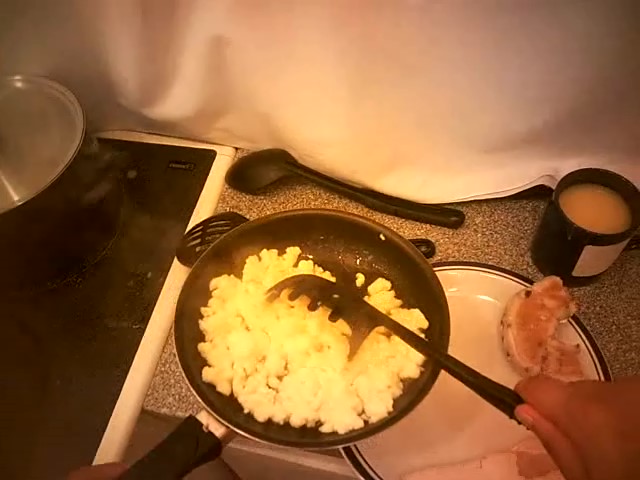} & \includegraphics[width=0.11\textwidth]{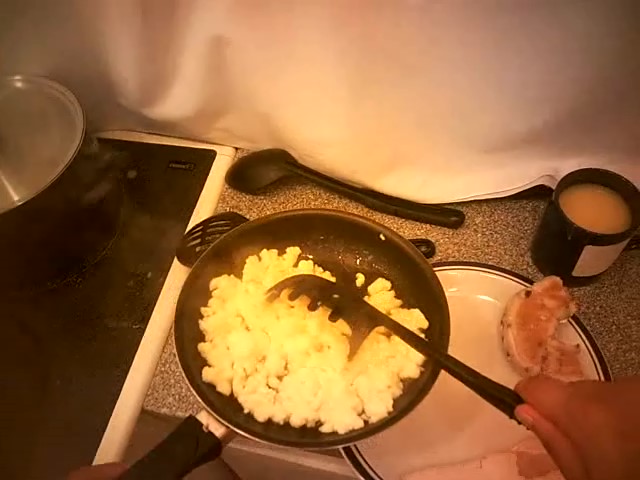} & \includegraphics[width=0.11\textwidth]{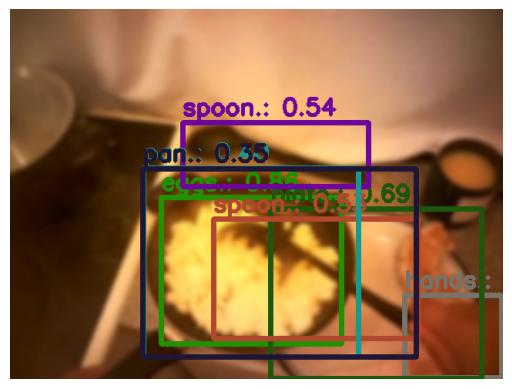} & \includegraphics[width=0.11\textwidth]{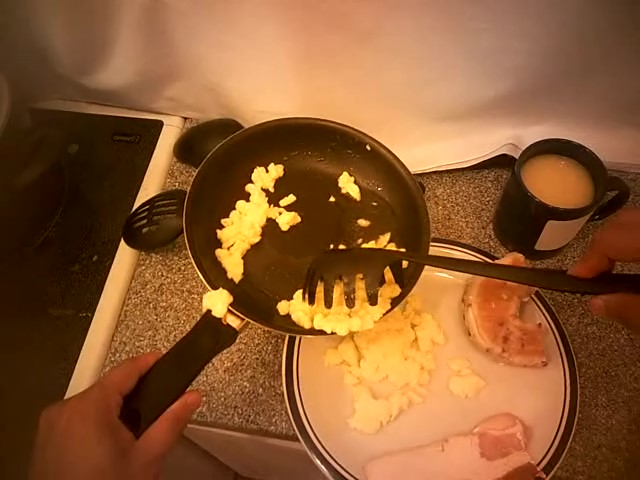} & \includegraphics[width=0.11\textwidth]{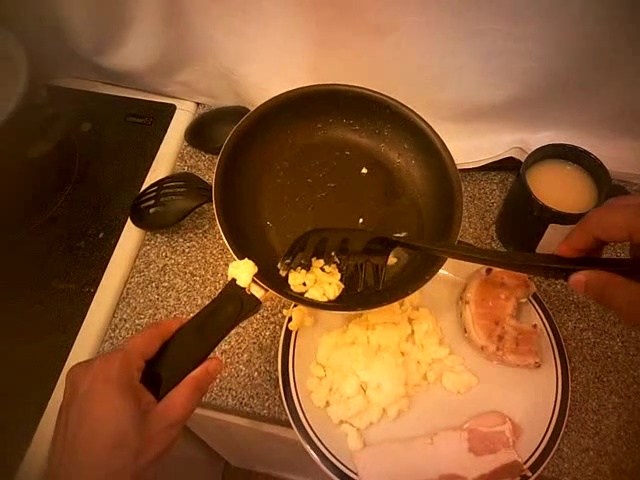} & \includegraphics[width=0.11\textwidth]{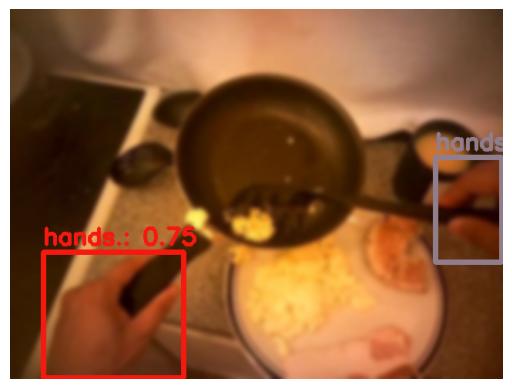} & \includegraphics[width=0.11\textwidth]{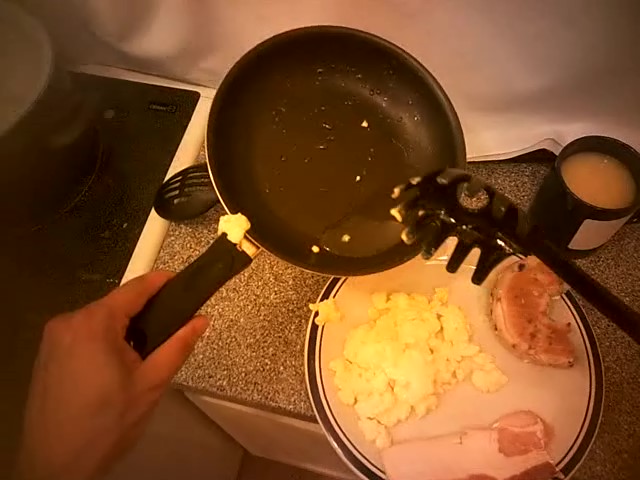} & \includegraphics[width=0.11\textwidth]{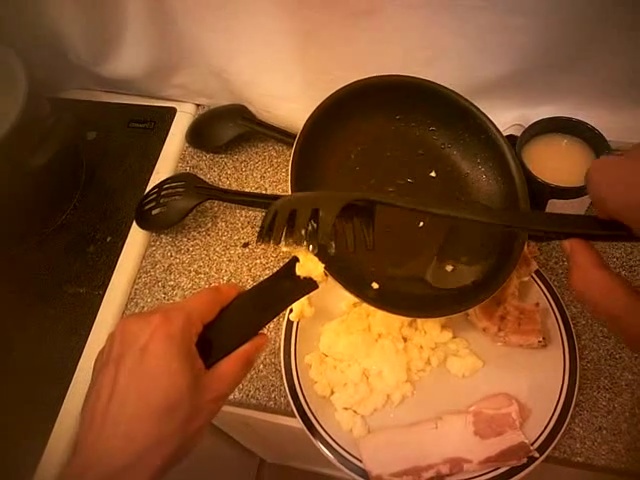} \\

        \includegraphics[width=0.11\textwidth]{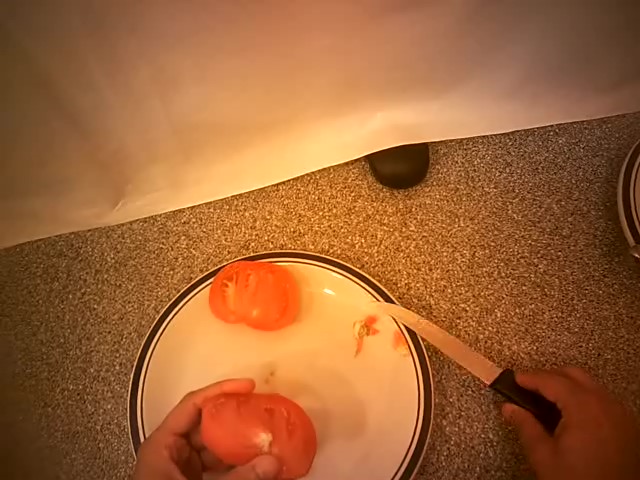} & \includegraphics[width=0.11\textwidth]{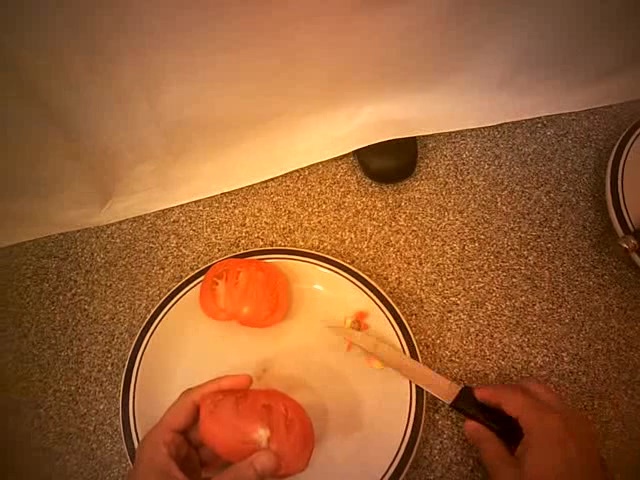} & \includegraphics[width=0.11\textwidth]{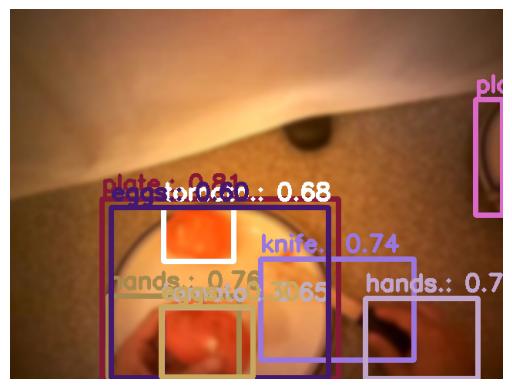} & \includegraphics[width=0.11\textwidth]{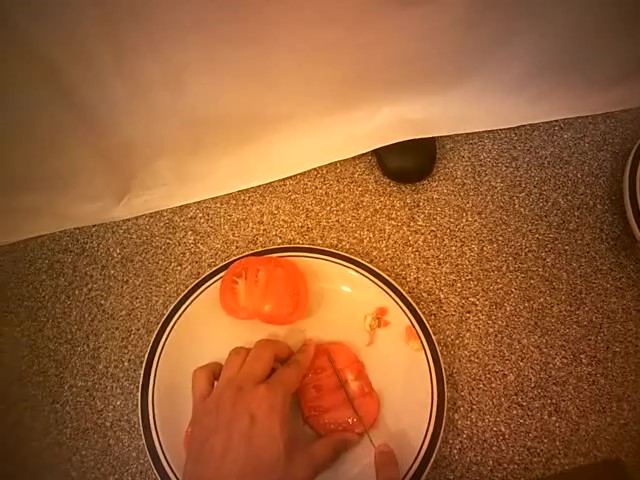} & \includegraphics[width=0.11\textwidth]{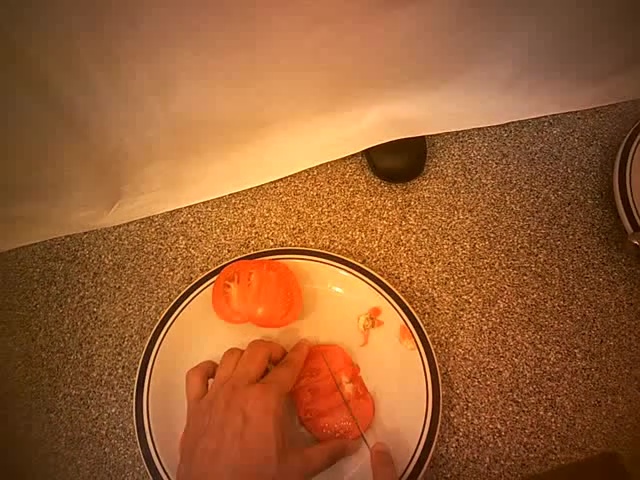} & \includegraphics[width=0.11\textwidth]{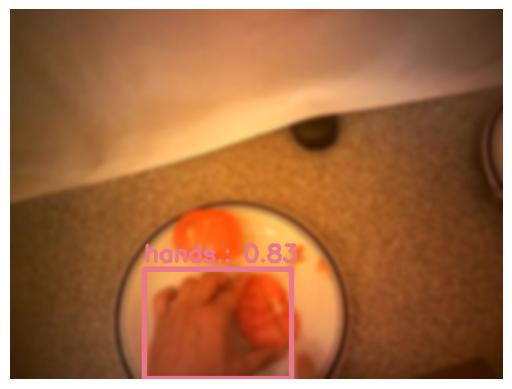} & \includegraphics[width=0.11\textwidth]{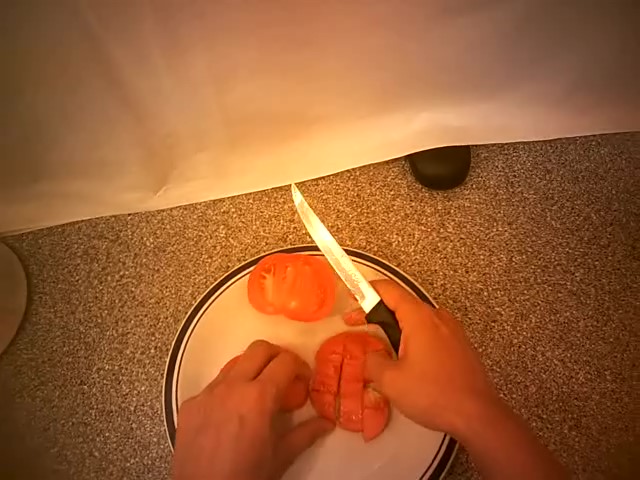} & \includegraphics[width=0.11\textwidth]{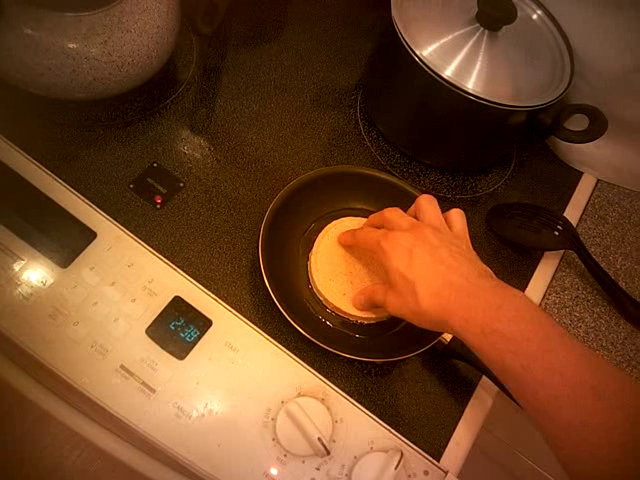} \\

        \includegraphics[width=0.11\textwidth]{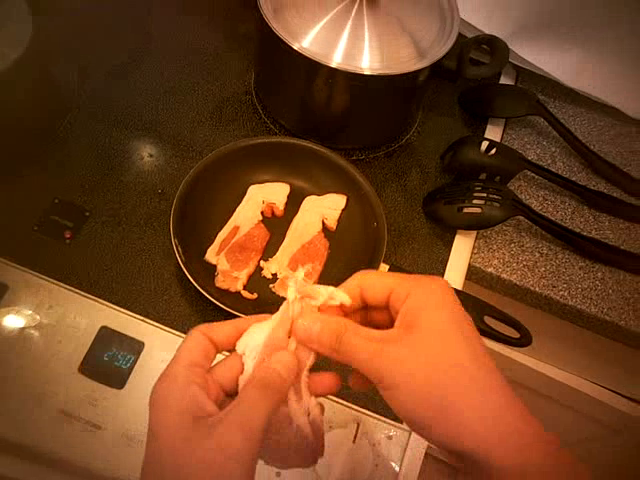} & \includegraphics[width=0.11\textwidth]{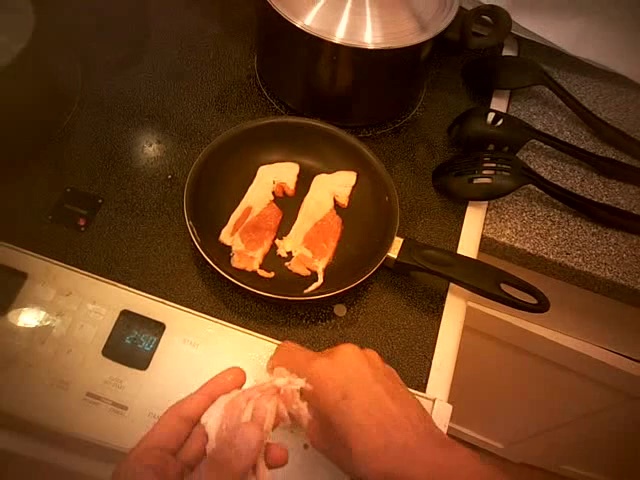} & \includegraphics[width=0.11\textwidth]{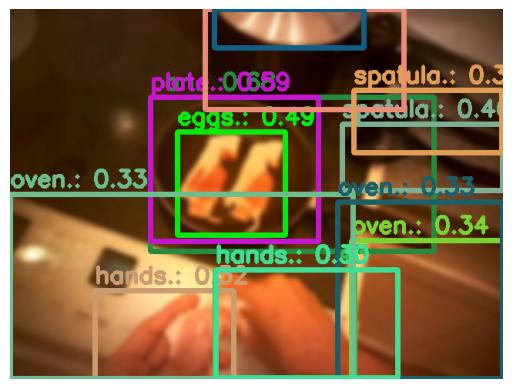} & \includegraphics[width=0.11\textwidth]{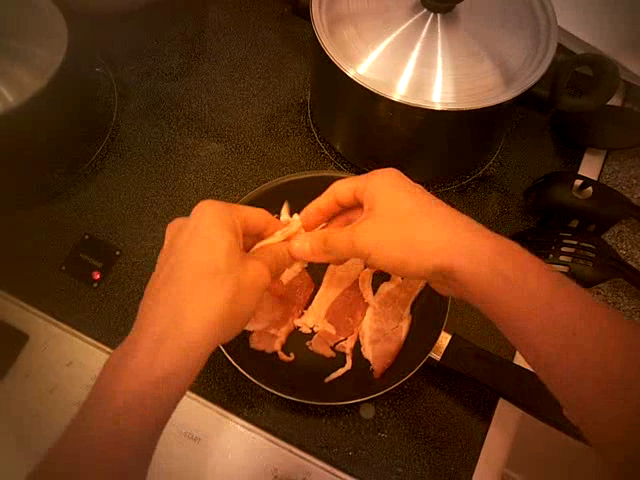} & \includegraphics[width=0.11\textwidth]{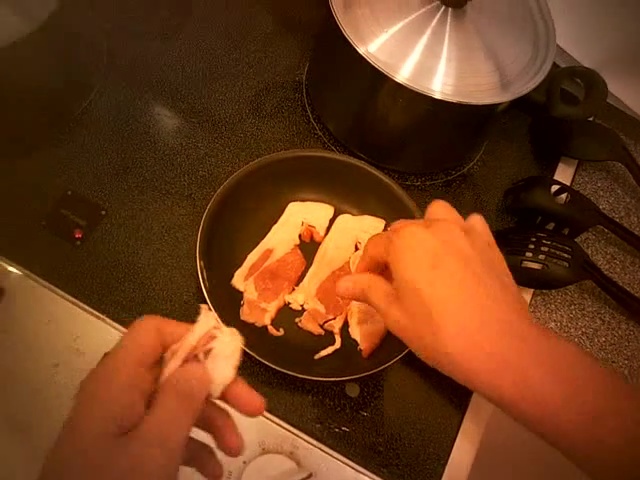} & \includegraphics[width=0.11\textwidth]{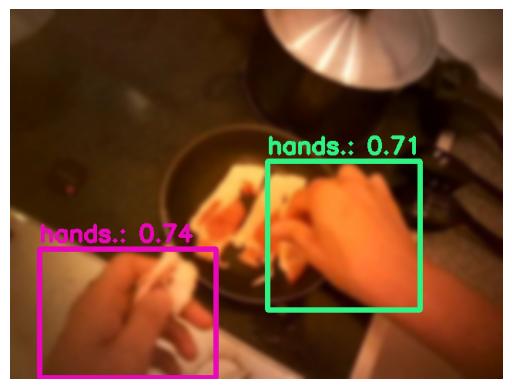} & \includegraphics[width=0.11\textwidth]{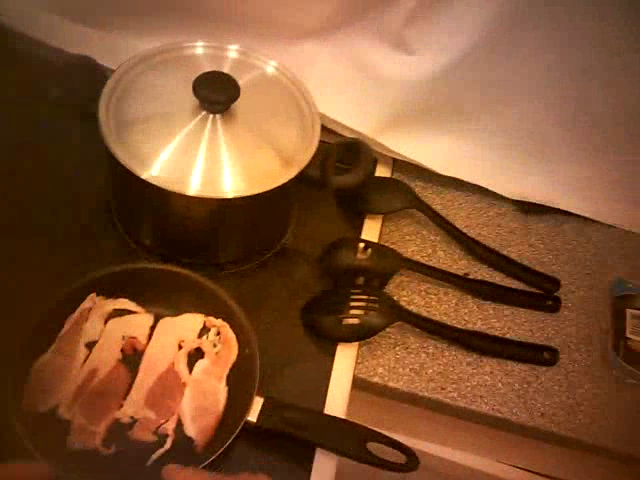} & \includegraphics[width=0.11\textwidth]{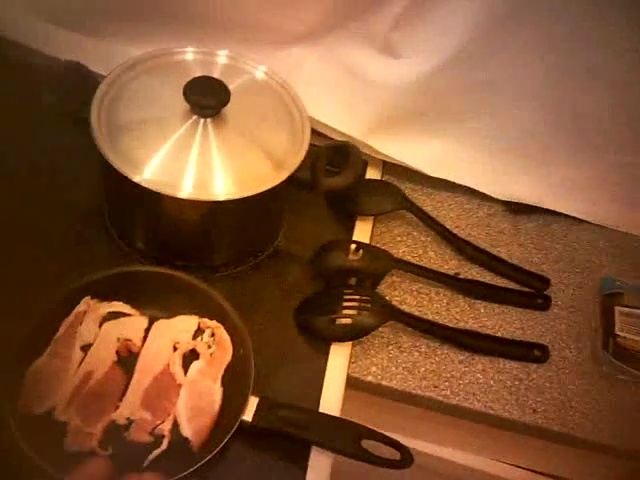} \\
    \end{tabular}
    \caption{\textbf{Visualization of the data curation procedure.} Human-selected ground truth frames, frames selected and filtered strategy, and visualization of detection used for filtering.}
    \label{fig:filtering-process}
\end{figure*}

Figure \ref{fig:filtering-process} demonstrates frames selected by our data curation strategy from the EGTEA~Gaze+ dataset. For qualitative comparison, we provide a ground truth image independently selected by a human for each frame. For each \textit{initial} and \textit{action frame}, we illustrate object/hand detection that was used in the filtration process. We do not perform any detection on final frames as previously described.

\subsection{Object Detecting}

\begin{figure}
    \centering
    \footnotesize
    \begin{tabular}{c@{\hskip 2pt}c@{\hskip 2pt}c}
        \includegraphics[width=0.32\linewidth]{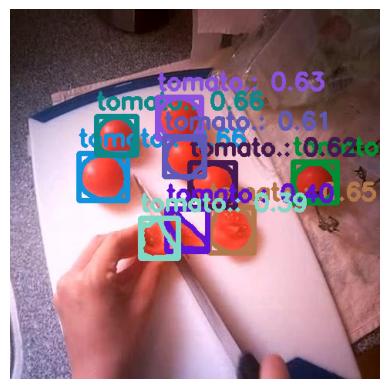} & \includegraphics[width=0.32\linewidth]{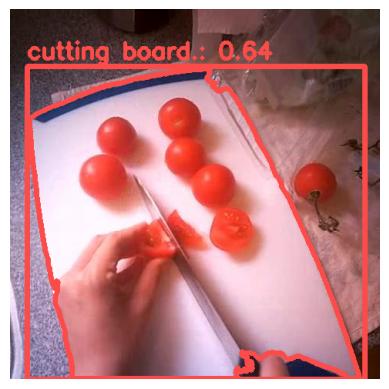} & \includegraphics[width=0.32\linewidth]{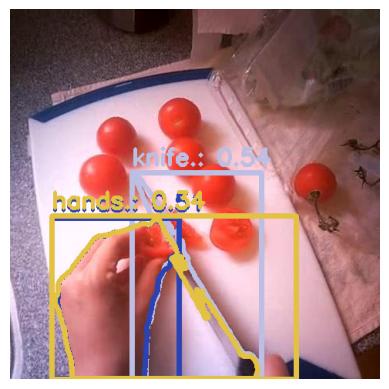} \\
        core & location & functional \\
         \includegraphics[width=0.32\linewidth]{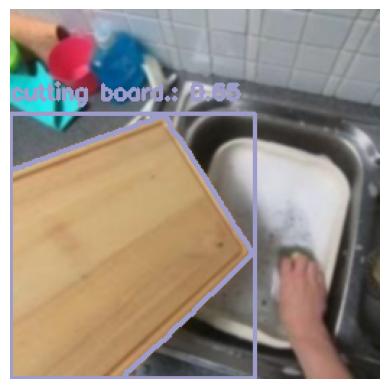} & \includegraphics[width=0.32\linewidth]{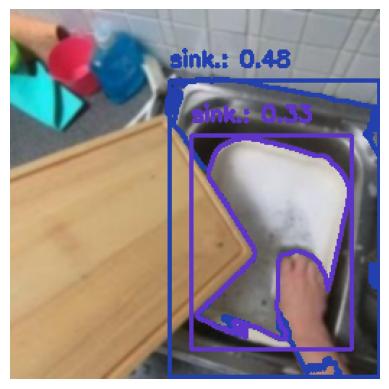} & \includegraphics[width=0.32\linewidth]{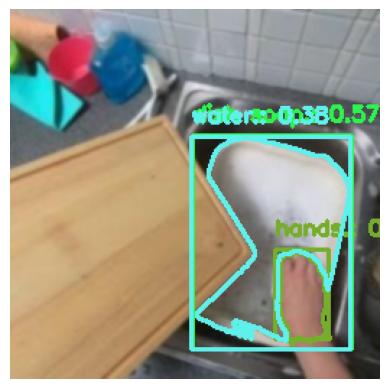}\\
    \end{tabular}
    \caption{\textbf{Examples of segmenting relevant objects in the initial frame.} We segment core, location, and functional objects for actions ``cut tomato" and ``wash cutting board" using the segmentation model.}
    \label{fig:segmenting-objects}
\end{figure}

Figure \ref{fig:segmenting-objects} illustrates relevant object detecting and segmentation in the initial frame. For actions ``cut tomato" and ``wash cutting board", we segment the core, location, and functional objects separately to apply specific strategies in our pipeline for each object category. For each object within each category, we obtain its bounding box, precise mask, and confidence score.

\begin{figure}
    \centering
    \includegraphics[width=\linewidth]{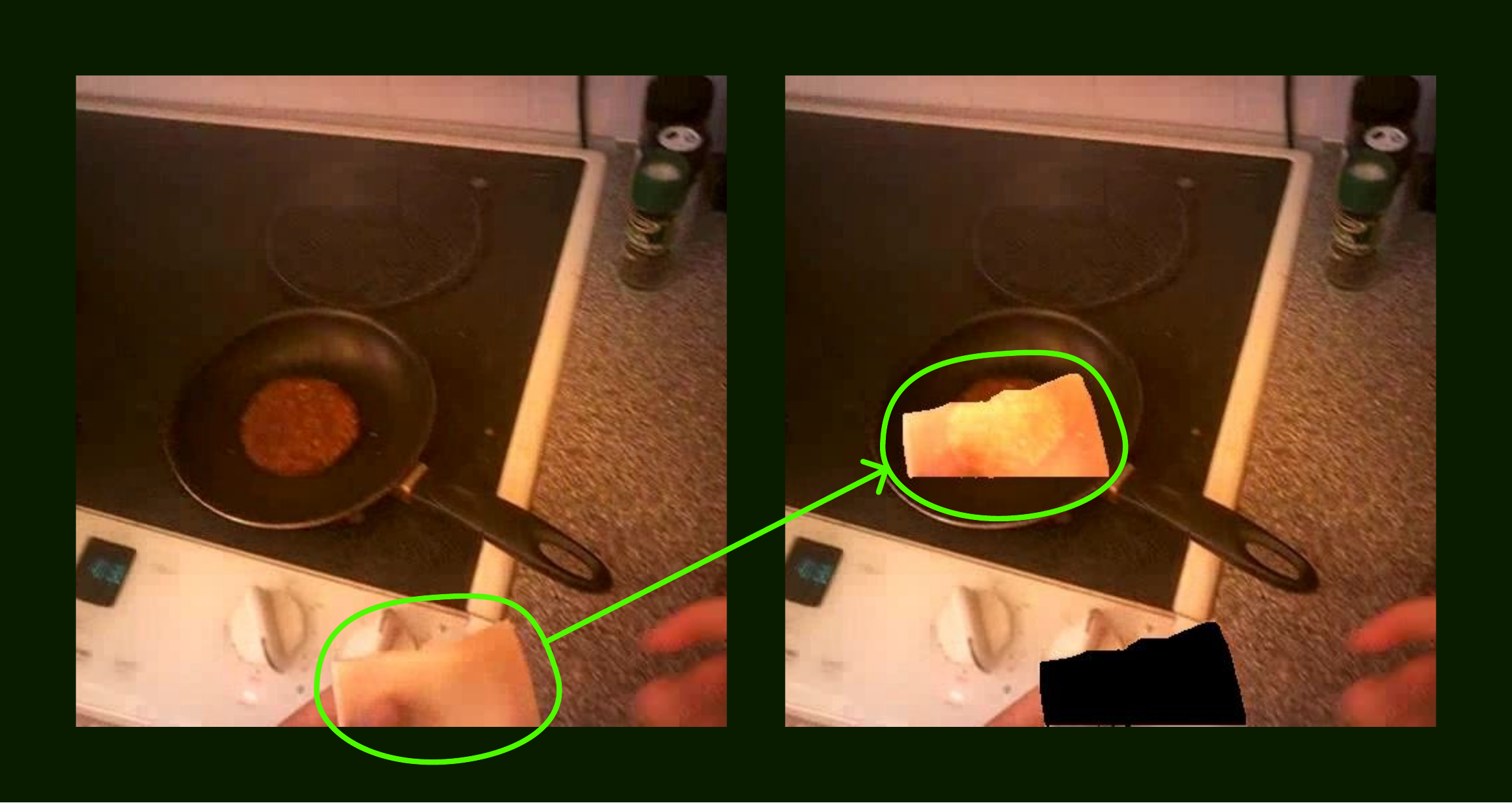}
    \caption{\textbf{Example of relocating the core object.} The core object ``cheese" is relocated to its target location ``burger" according to the action ``put cheese".}
    \label{fig:relocation}
\end{figure}

\subsection{Core Objects Relocation}

One of our goals is to perform as few changes as possible. However, using the entire location object for masked inpainting would introduce a larger area where the changes are permitted. For this reason, we use the location object not as a mask, but as the final destination of the core object after the action is complete. Using Grounded-SAM~\cite{ren2024grounded}, we segment the exact mask of the core object and move it to the final location as shown in \cref{fig:relocation}. The output of this stage is used as one of the inputs to the diffusion pipeline (see \cref{fig:method}).

\section{Metrics Intuition}

\subsection{Similarity between the Input and Target Images}

 \begin{table}
     \caption[]{\textbf{Baseline similarity between the input and target images in the EGTEA~Gaze+ and Ego4D evaluations sets.} High CLIP similarity score can be achieved without any processing; therefore, we need to account for this similarity when defining metrics. \label{tab:baseline_similarity}}
    \centering
    \footnotesize
    \vspace{2mm}
    \begin{tabular}{c c c c}
    \hline
     Frame & Score & EGTEA Gaze+ & Ego4D\\
    \hline
    \multirow{3}{*}{Action frame}
		& CLIP & 84.91 & 90.00 \\
        & STD & 7.16 & 6.85 \\
        & CLIP $\geq$ 80 & 79 & 92 \\
        \hline
        \multirow{3}{*}{Final frame}
		& CLP & 82.52 & 86.05\\
        & STD & 9.68 & 9.14 \\
		&CLIP $\geq$ 80 & 75 & 82\\
    \hline
    \end{tabular}
\end{table}

We conducted an experiment to evaluate the baseline similarity between input and target images. As previously mentioned, the majority of actions occur with a stable camera pose, so our goal was to assess the similarity within entries in the EGTEA Gaze+ and Ego4D datasets. Table \ref{tab:baseline_similarity} presents the statistics for CLIP scores, comparing the input images with both action and final frames.

The results indicate not only a high mean CLIP score but also a substantial number of individual entries scoring 80 or above, with low variance across the dataset. This suggests a high similarity score can be achieved without any processing at all. Therefore, we need to account for this baseline similarity, which is why we introduced different metrics designed to address this issue.

\subsection{Evaluation with Image-Text Similarity}
\begin{figure}
    \centering
    \footnotesize
{\includegraphics[width=\linewidth]{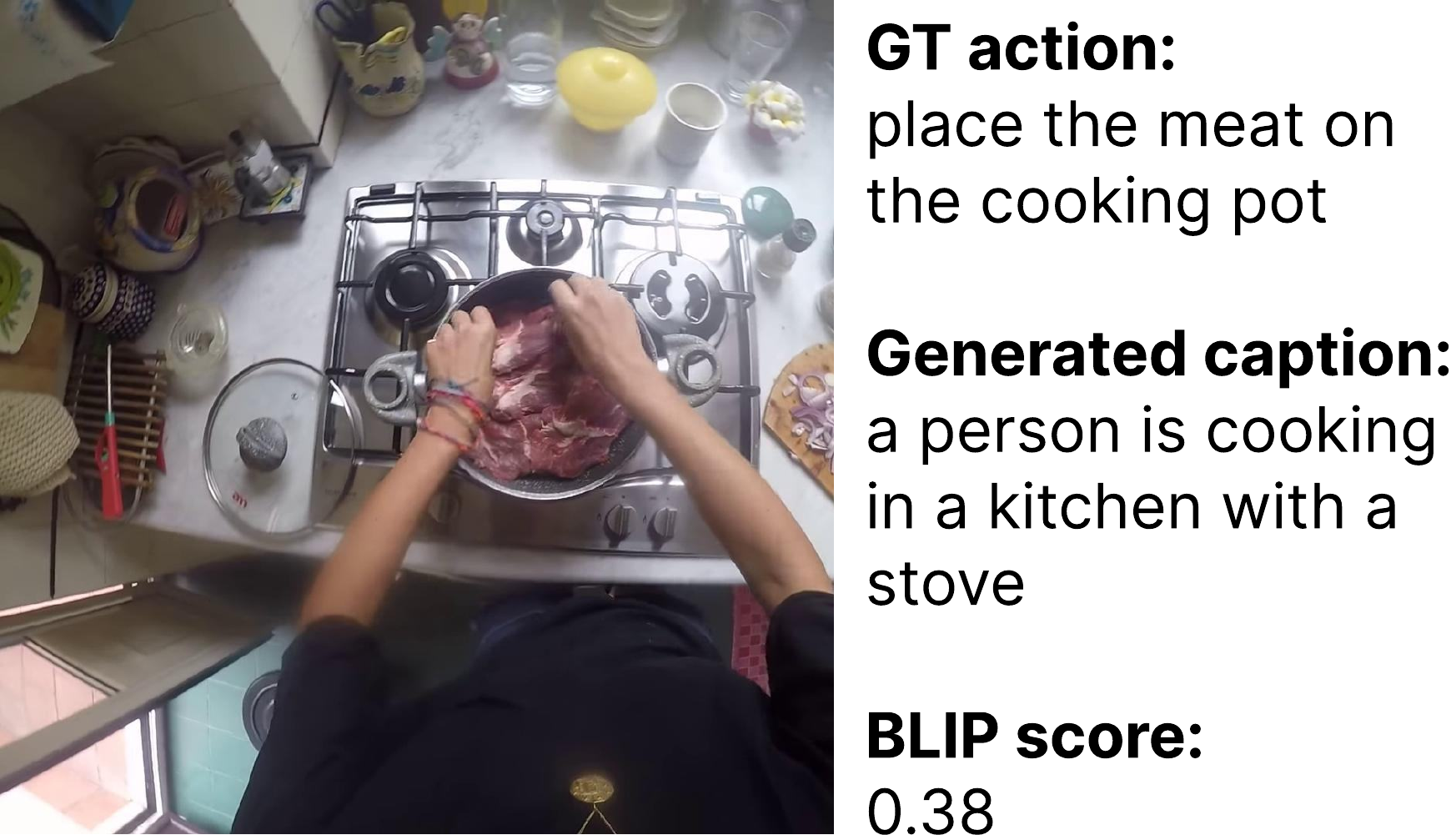}}
    \caption{\textbf{Example output of the BLIP-based image-text evaluation metric on the entry of the Ego4D dataset.} The generated caption is overly general which does not allow us to use the BLIP score as a metric efficiently.}
    \label{fig:blip-example}
\end{figure}

We also followed LEGO \cite{lai2024lego} to explore image-text evaluation metrics within our dataset, leveraging the \textbf{BLIP} \cite{li2022blip} model for image caption generation.
After generating captions, we employed a sentence transformer to encode both the generated and target image descriptions, using the cosine similarity score between these encodings as our evaluation metric.
These experiments were conducted on the EGTEA Gaze+ and Ego4D evaluation sets.

The resulting scores, however, were notably low. This is largely because the generated captions often capture only the broader context rather than the specific action. For instance, where the target action is “places the meat on the cooking pot,” the generated caption might state, “a person is cooking in a kitchen with a stove” (see \cref{fig:blip-example}). While the generated caption is not incorrect, it is overly general and misses the specific, detailed action—particularly for actions “open,” “close,” “put,” or “take,” where the generated text defaults to generalized statements about the overall scene or recipe. Consequently, this metric is not reliable for our purposes. For image-text metrics to be meaningful in this context, they would need to capture more granular details of the actions depicted.

\subsection{CLIP Distributions}

\begin{figure}
    \centering
    \includegraphics[width=\linewidth]{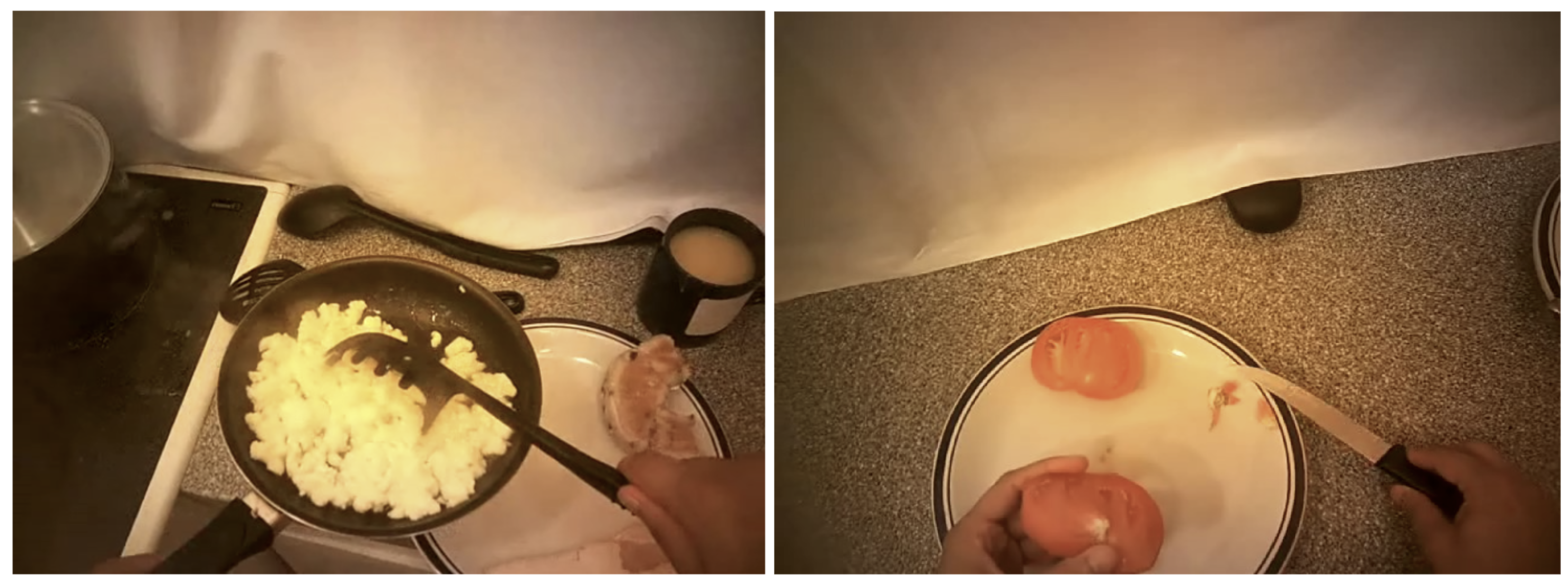}
    \caption{\textbf{Video frame selected for evaluation.} The frames from EGTEA~Gaze+ represent two actions: ``Transfer eggs from pan to plate" and ``Cut tomato".}
    \label{fig:distribution-selected}
\end{figure}

\begin{figure}
    \centering
    \footnotesize
    \begin{tabular}{c@{\hskip 3pt}cc@{\hskip 3pt}cc@{\hskip 3pt}c}
        & \multirow{5}{*}{\includegraphics[width=0.2\linewidth]{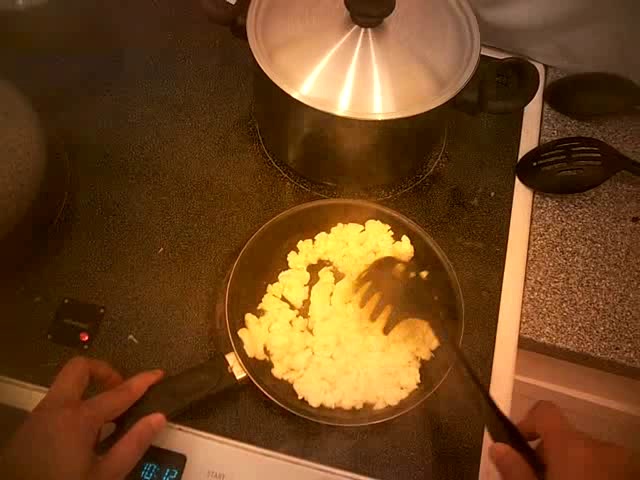}} & & \multirow{5}{*}{\includegraphics[width=0.2\linewidth]{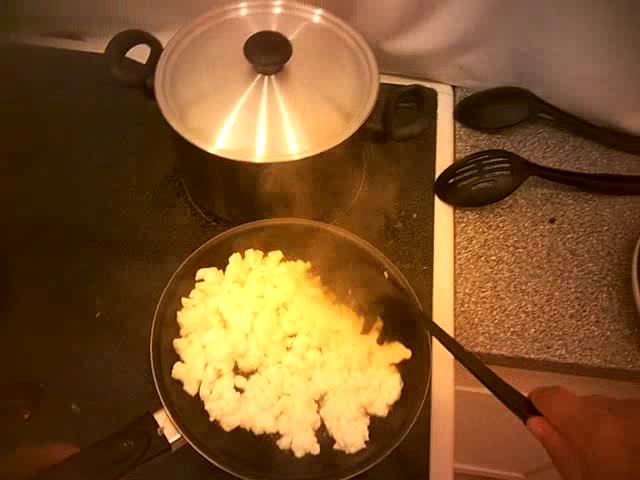}} &  & \multirow{5}{*}{\includegraphics[width=0.2\linewidth]{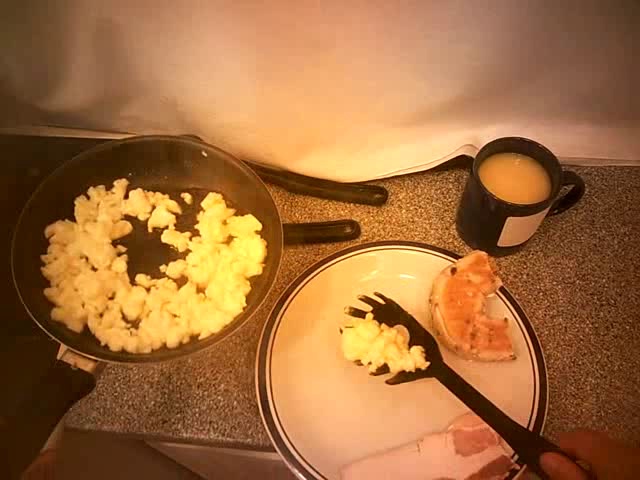}} \\
        17860 &  & 17980 &  & 18040 \\
        \\
        83.61 &  & 83.67 &  & 87.62 \\
        \\
        
        & \multirow{5}{*}{\includegraphics[width=0.2\linewidth]{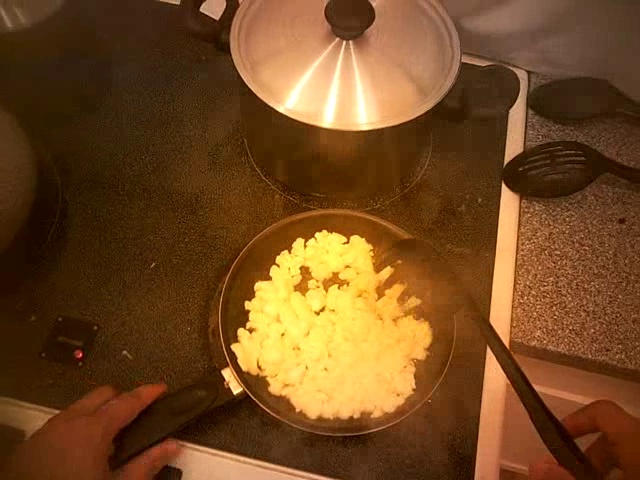}} & & \multirow{5}{*}{\includegraphics[width=0.2\linewidth]{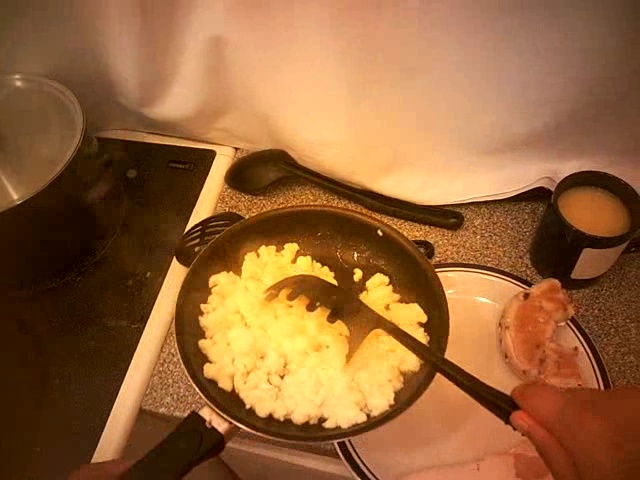}} &  & \multirow{5}{*}{\includegraphics[width=0.2\linewidth]{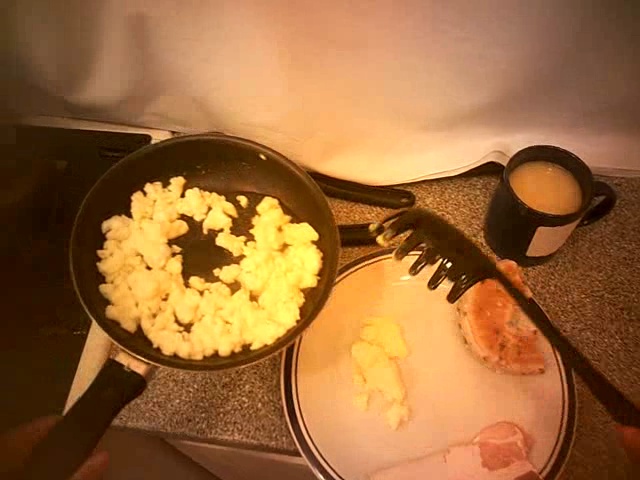}} \\
        17880 &  & 18000 &  & 18060 \\
        \\
        87.69 &  & 96.52 &  & 87.98 \\
        \\

        & \multirow{5}{*}{\includegraphics[width=0.2\linewidth]{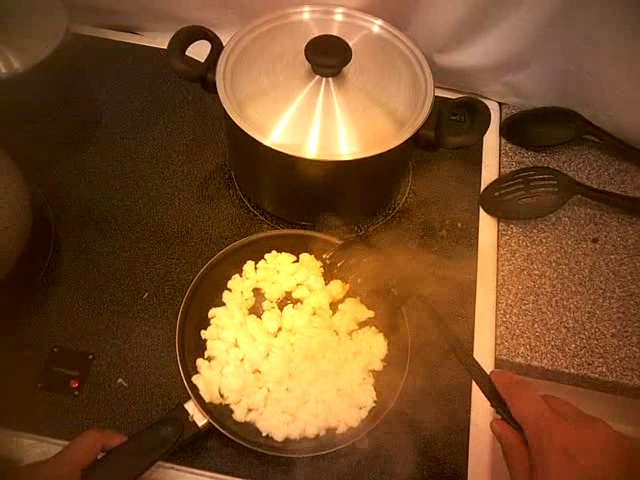}} & & \multirow{5}{*}{\includegraphics[width=0.2\linewidth]{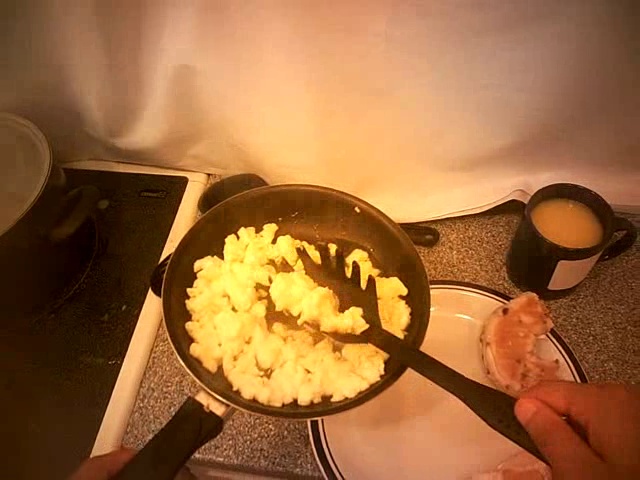}} &  & \multirow{5}{*}{\includegraphics[width=0.2\linewidth]{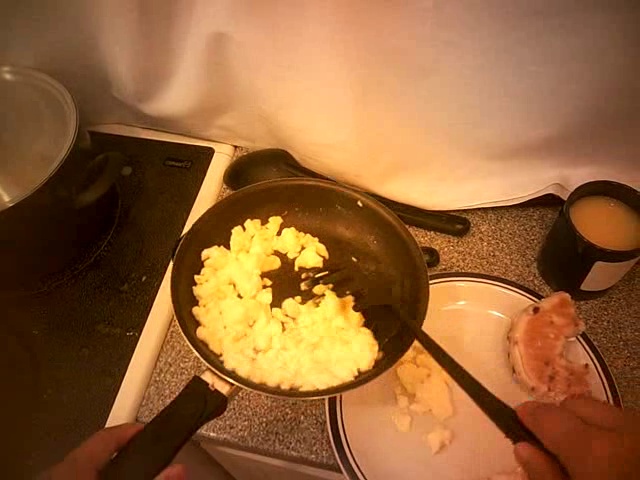}} \\
        17900 &  & 18020 &  & 18080 \\
        \\
        85.63 &  & 94.42 &  & 92.90 \\
        \\
        \\
        
        & \multirow{5}{*}{\includegraphics[width=0.2\linewidth]{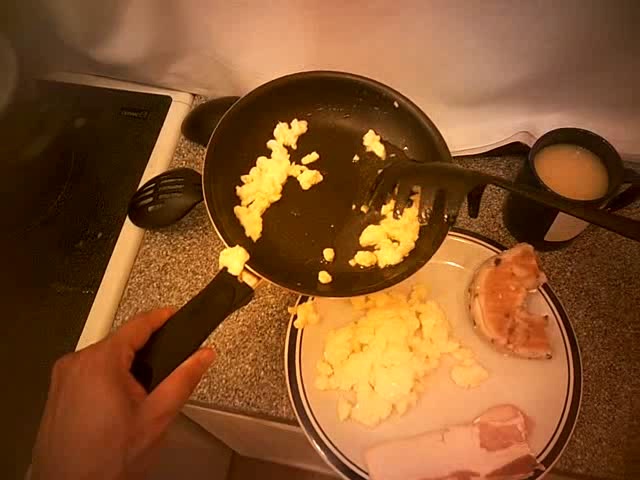}} & & \multirow{5}{*}{\includegraphics[width=0.2\linewidth]{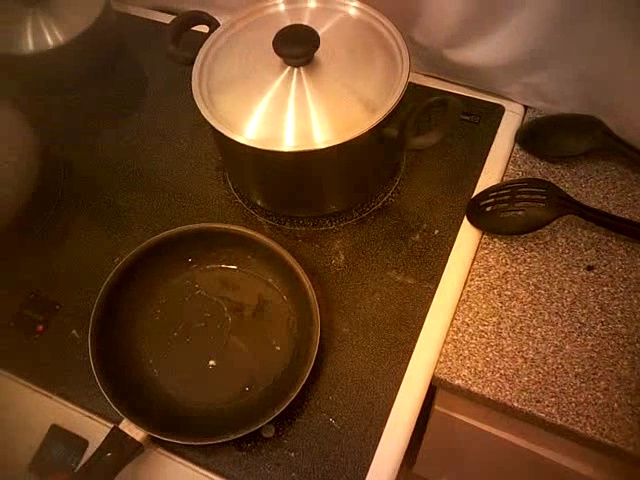}} &  & \multirow{5}{*}{\includegraphics[width=0.2\linewidth]{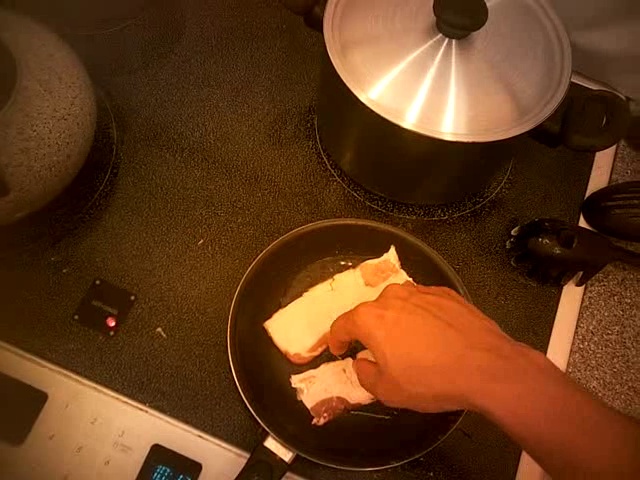}} \\
        18460 &  & 19220 &  & 19580 \\
        \\
        84.10 &  & 70.43 &  & 72.98 \\
        \\
        
        & \multirow{5}{*}{\includegraphics[width=0.2\linewidth]{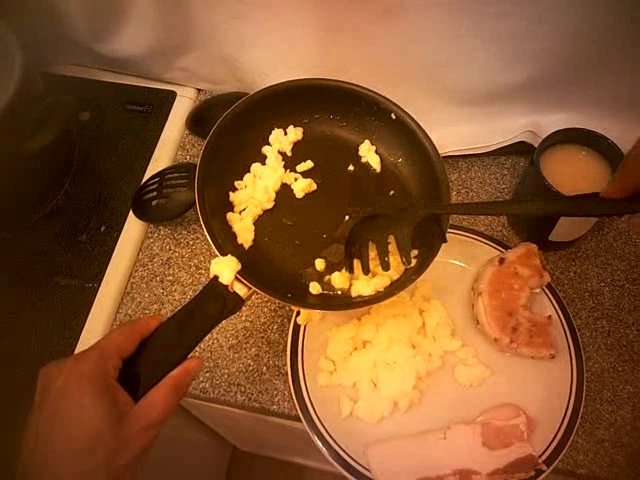}} & & \multirow{5}{*}{\includegraphics[width=0.2\linewidth]{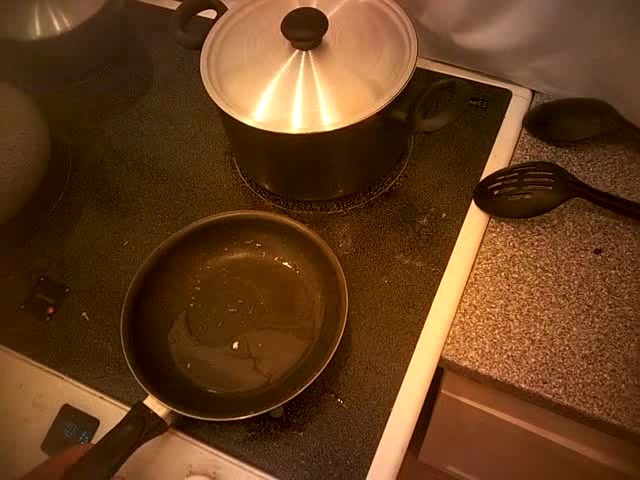}} &  & \multirow{5}{*}{\includegraphics[width=0.2\linewidth]{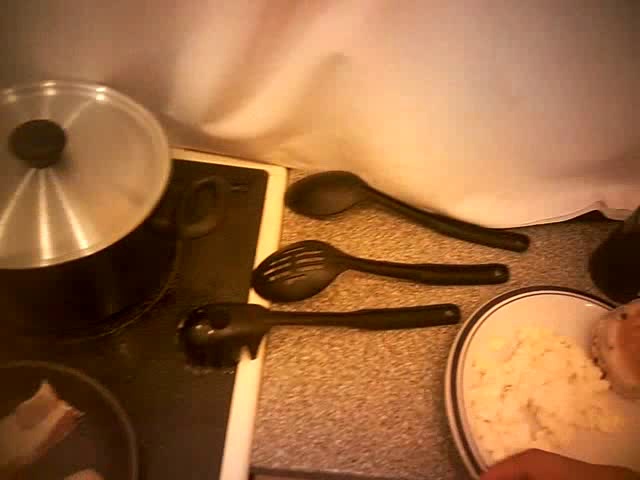}} \\
        18480 &  & 19240 &  & 19600 \\
        \\
        85.93 &  & 72.19 &  & 65.92 \\
        \\

        & \multirow{5}{*}{\includegraphics[width=0.2\linewidth]{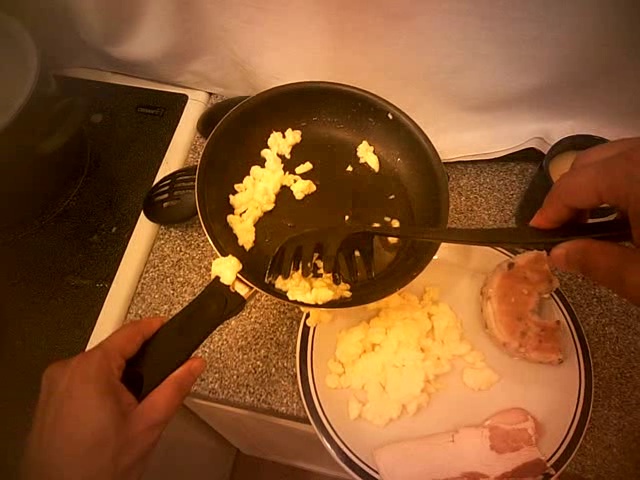}} & & \multirow{5}{*}{\includegraphics[width=0.2\linewidth]{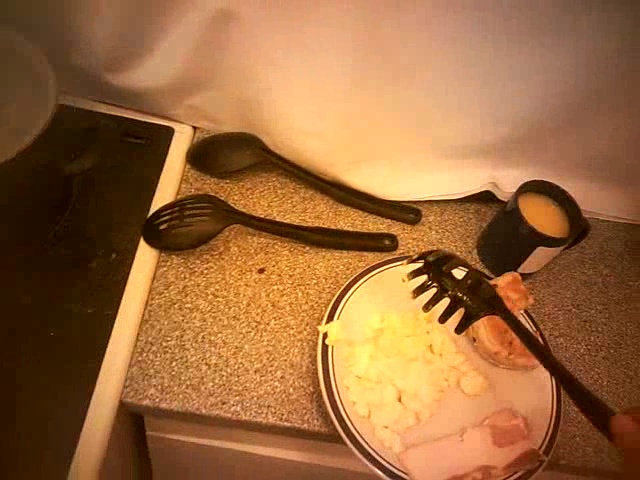}} &  & \multirow{5}{*}{\includegraphics[width=0.2\linewidth]{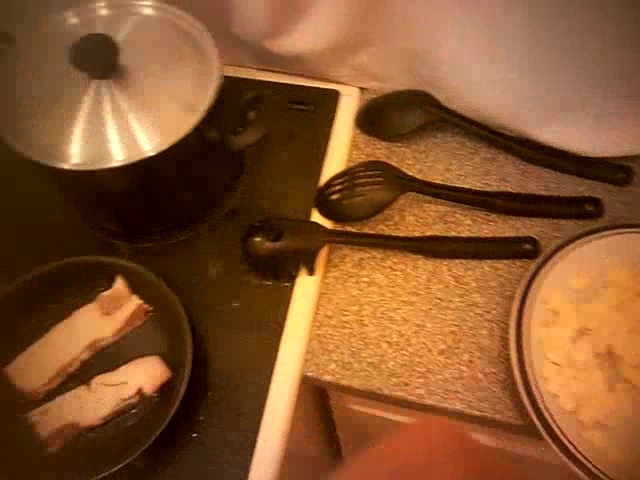}} \\
        18500 &  & 19260 &  & 19620 \\
        \\
        85.57 &  & 74.44 &  & 69.56 \\
        \\    
    \end{tabular}
    
    \caption{\textbf{CLIP similarity score distribution for action ``Transfer eggs from pan to plate".} We observe extremely high scores when the context corresponds to the given frame and the drop as soon as the context changes.}
    \label{fig:distribution-transfer-eggs}
\end{figure}

\begin{figure}
    \centering
    \footnotesize
    \begin{tabular}{c@{\hskip 3pt}cc@{\hskip 3pt}cc@{\hskip 3pt}c}
        & \multirow{5}{*}{\includegraphics[width=0.2\linewidth]{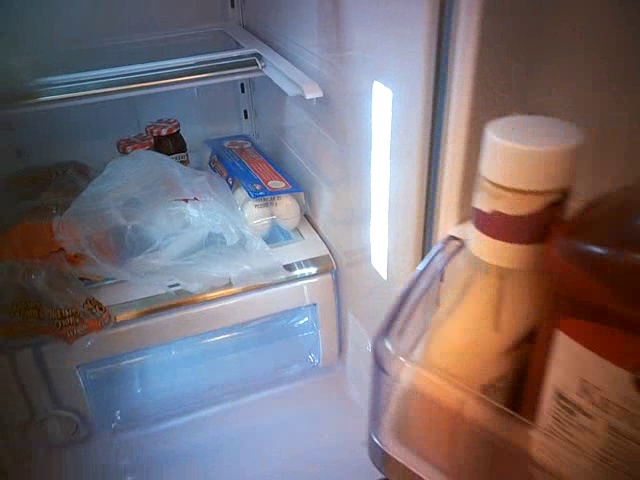}} & & \multirow{5}{*}{\includegraphics[width=0.2\linewidth]{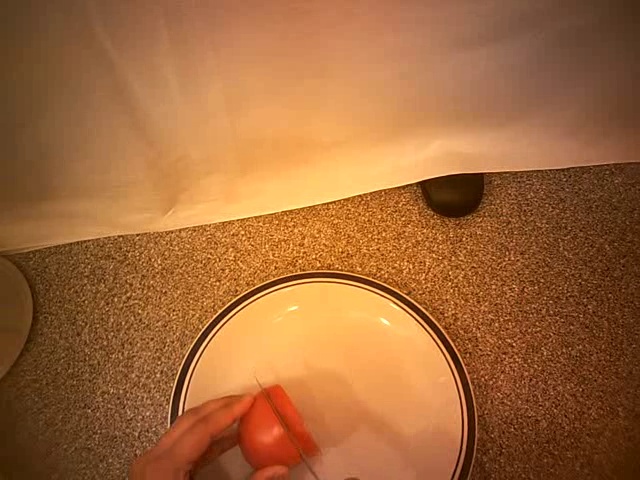}} &  & \multirow{5}{*}{\includegraphics[width=0.2\linewidth]{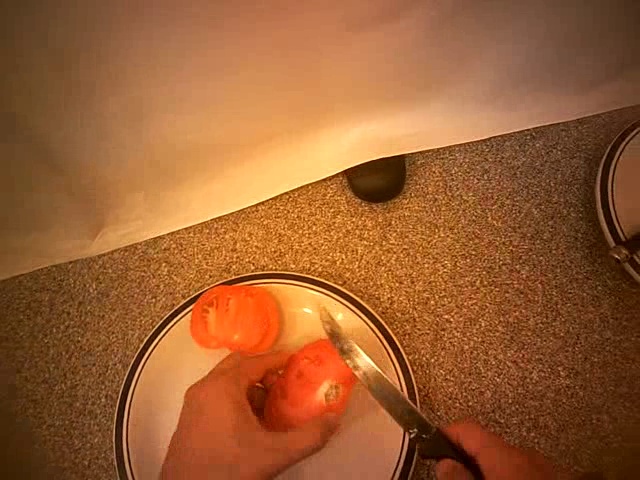}} \\
        3090 &  & 3330 &  & 3750 \\
        \\
        55.86 &  & 85.86 &  & 88.28 \\
        \\
        
        & \multirow{5}{*}{\includegraphics[width=0.2\linewidth]{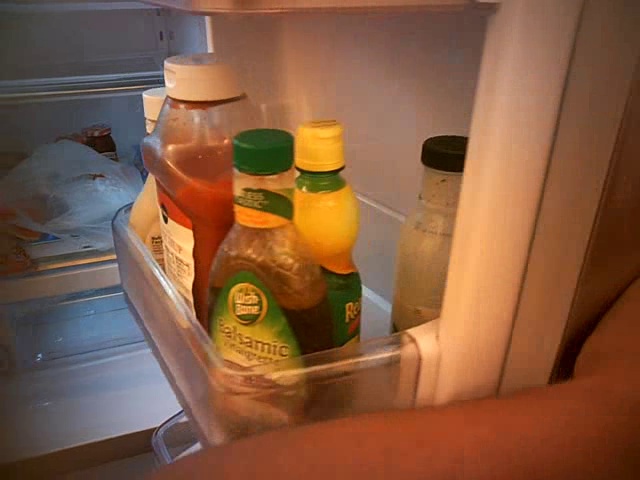}} & & \multirow{5}{*}{\includegraphics[width=0.2\linewidth]{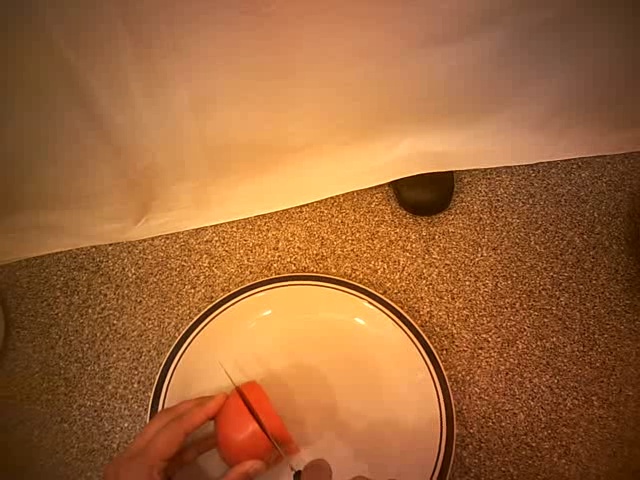}} &  & \multirow{5}{*}{\includegraphics[width=0.2\linewidth]{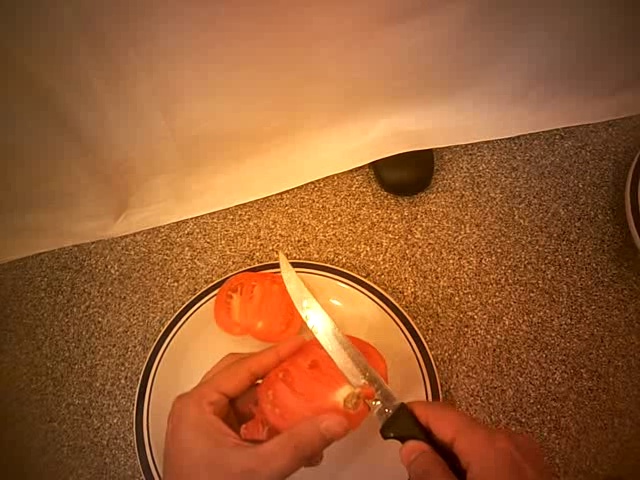}} \\
        3120 &  & 3360 &  & 3780 \\
        \\
        54.96 &  & 87.66 &  & 87.63 \\
        \\

        & \multirow{5}{*}{\includegraphics[width=0.2\linewidth]{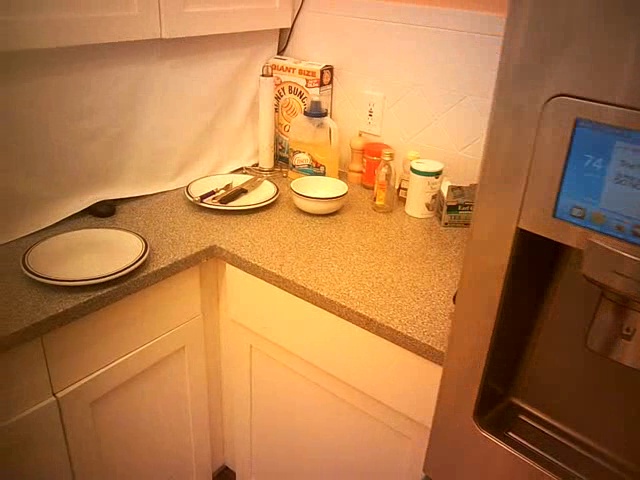}} & & \multirow{5}{*}{\includegraphics[width=0.2\linewidth]{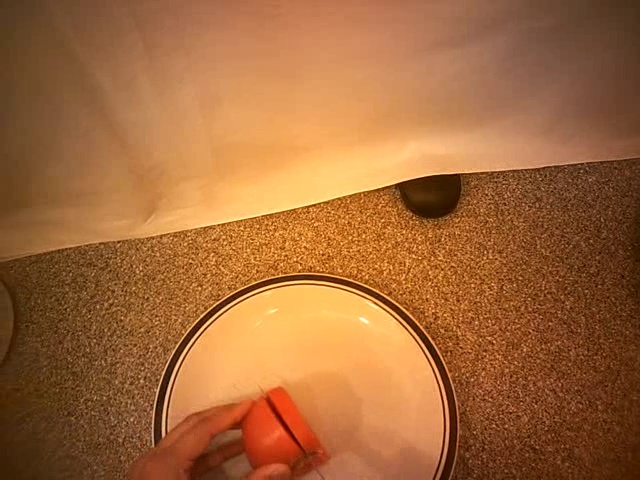}} &  & \multirow{5}{*}{\includegraphics[width=0.2\linewidth]{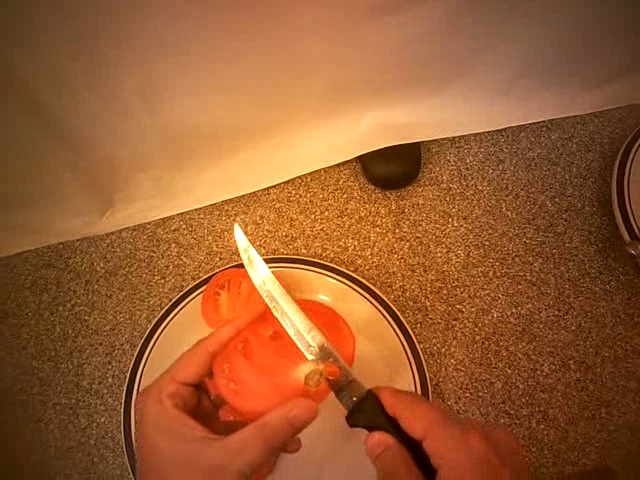}} \\
        3150 &  & 3390 &  & 3810 \\
        \\
        61.16 &  & 86.48 &  & 83.69 \\
        \\
        \\
        
        & \multirow{5}{*}{\includegraphics[width=0.2\linewidth]{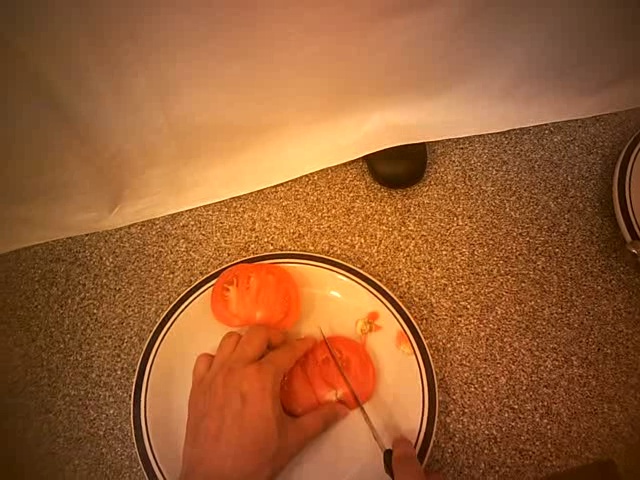}} & & \multirow{5}{*}{\includegraphics[width=0.2\linewidth]{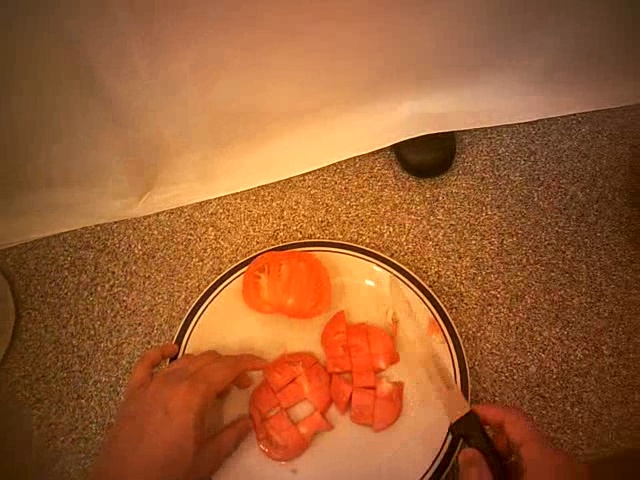}} &  & \multirow{5}{*}{\includegraphics[width=0.2\linewidth]{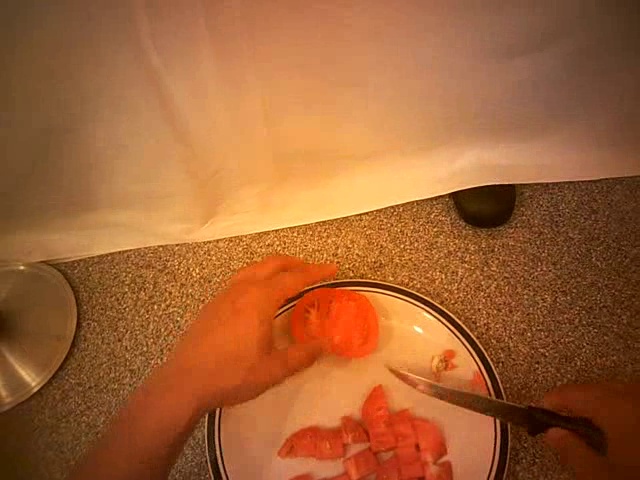}} \\
        4410 &  & 5190 &  & 5310 \\
        \\
        88.66 &  & 87.10 &  & 82.73 \\
        \\
        
        & \multirow{5}{*}{\includegraphics[width=0.2\linewidth]{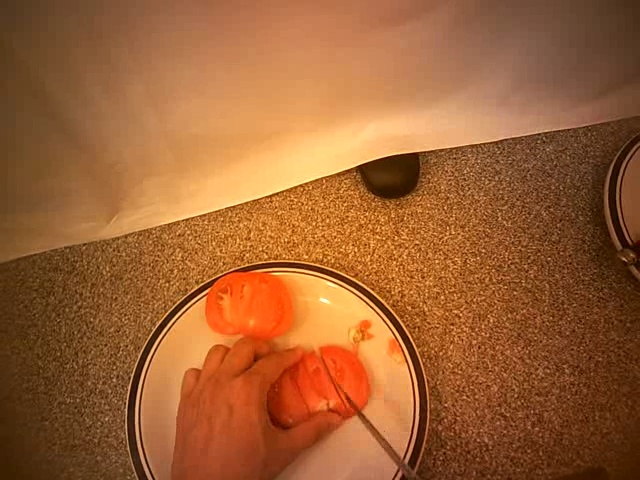}} & & \multirow{5}{*}{\includegraphics[width=0.2\linewidth]{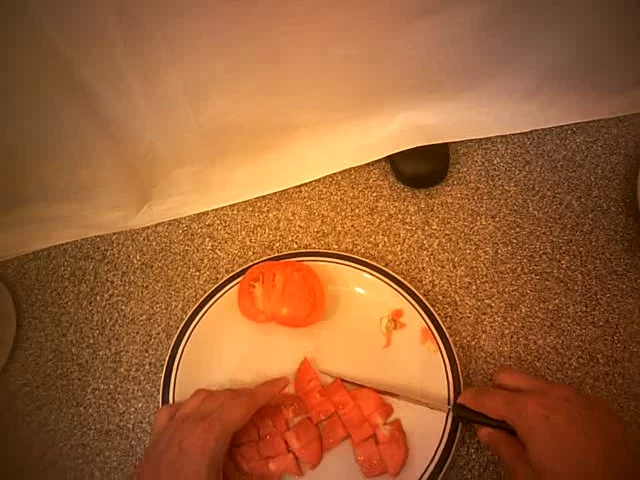}} &  & \multirow{5}{*}{\includegraphics[width=0.2\linewidth]{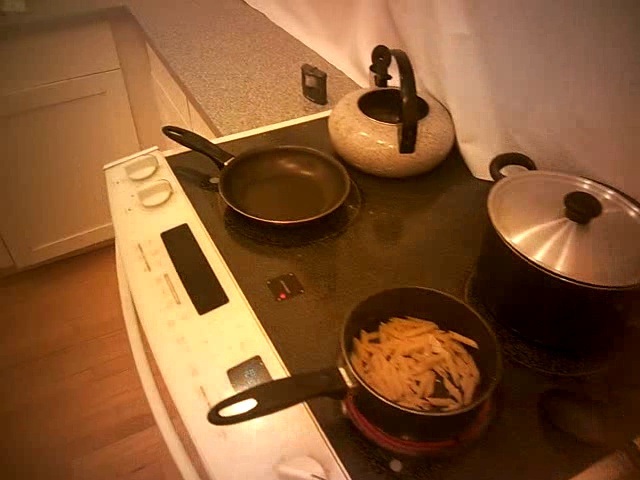}} \\
        4440 &  & 5220 &  & 5340 \\
        \\
        88.57 &  & 87.74 &  & 62.15 \\
        \\

        & \multirow{5}{*}{\includegraphics[width=0.2\linewidth]{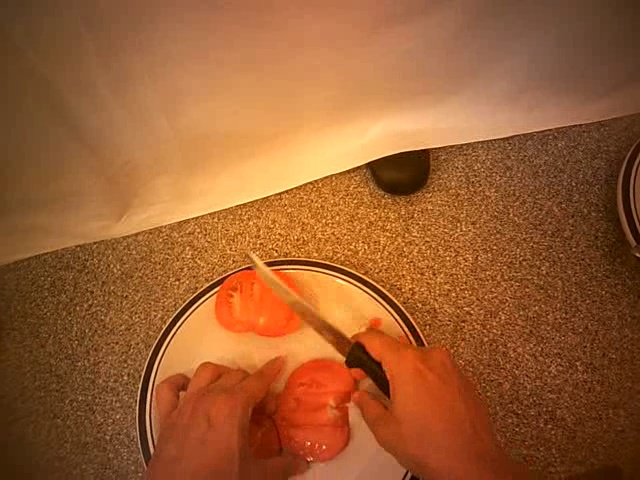}} & & \multirow{5}{*}{\includegraphics[width=0.2\linewidth]{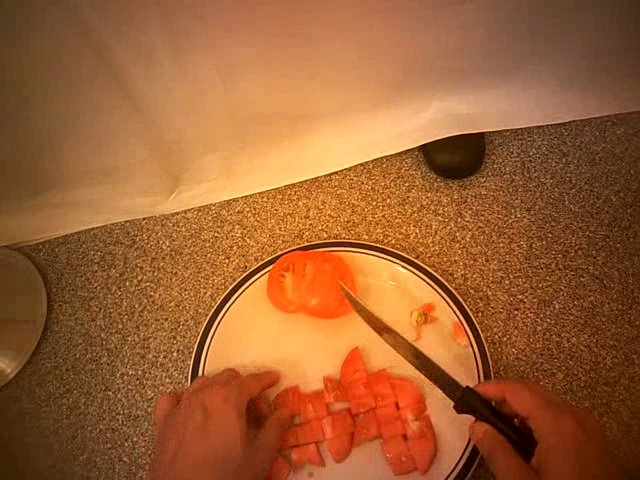}} &  & \multirow{5}{*}{\includegraphics[width=0.2\linewidth]{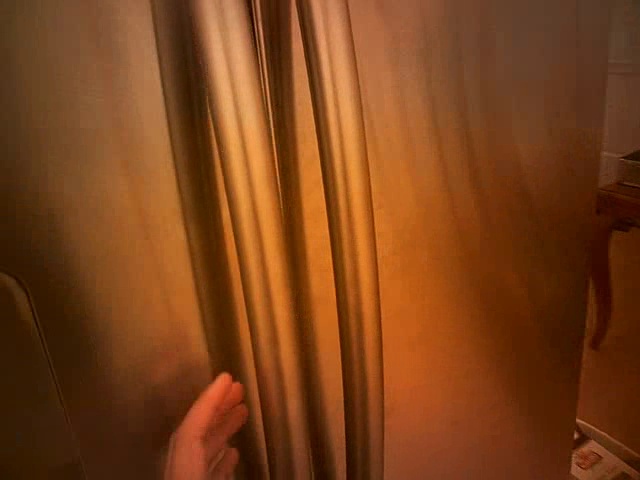}} \\
        4470 &  & 5250 &  & 5370 \\
        \\
        89.14 &  & 87.36 &  & 54.33 \\
        \\    
    \end{tabular}
    
    \caption{\textbf{CLIP similarity score distribution for action ``Cut tomato".} We observe the highest CLIP score for the frames that demonstrate the cutting tomato action in the same setting.}
    \label{fig:distribution-cut-tomato}
\end{figure}

\begin{figure}
    \centering
    \includegraphics[width=\linewidth]{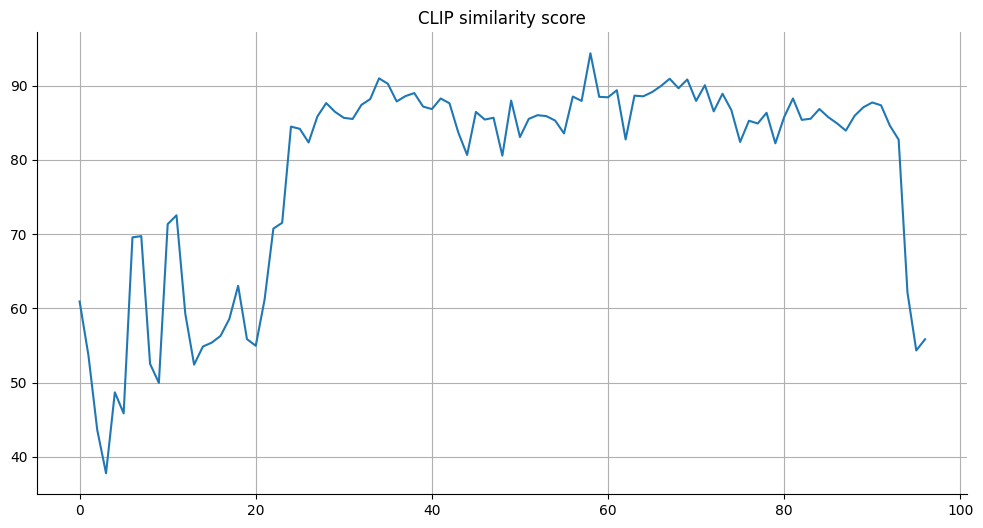}
    \caption{\textbf{CLIP similarity score distribution for action ``Cut tomato".} The score stabilizes at value 80 when the frames show the action and drops before and after the action when the semantic content is different.}
    \label{fig:distribution-score}
\end{figure}

To evaluate the effectiveness of the CLIP similarity score as a metric for frame similarity in video-based action recognition, we conducted an experiment analyzing how the CLIP similarity score changes over time. Specifically, our goal was to validate that the similarity metric could serve as a reliable evaluation tool for recognizing semantically similar frames across various actions within a video sequence.

For this experiment, we selected specific frames representing key actions to be part of the evaluation set. We computed the CLIP similarity score between each chosen evaluation frame and a series of video frames sampled at uniform intervals, generating a distribution of similarity scores across timestamps. This setup allowed us to observe how the CLIP score behaves in relation to visual and contextual shifts throughout the video sequence.

We provide two illustrative cases for the actions ``Transfer eggs from pan to plate" and ``Cutting tomato", with corresponding selected evaluation frames shown in Figure \ref{fig:distribution-selected}.

 The CLIP similarity score distribution for the action ``Transfer eggs from pan to plate" is shown in Figure \ref{fig:distribution-transfer-eggs}. Here, the timestamps (in seconds since the beginning of the video) and the CLIP similarity score between the given video frame and the selected video frame are presented in the first column, with each corresponding video frame in the second column. During frames showing elements like the stove, eggs in the pan, and nearby table the similarity score remains high, at around 80 since the same objects appear in the evaluation video frame. As the pan and camera shift to show the plate, the similarity score further increases, reaching approximately 90 when the frame nearly replicates the selected evaluation frame. Once the eggs are transferred to the plate, the visual context changes, therefore, resulting in a drop in the similarity score. This decrease persists through actions with an empty pan on the stove and the addition of bacon, where the score reaches its lowest point.

 A similar example distribution for the action ``Cut tomato" can be seen in Figure \ref{fig:distribution-cut-tomato}. Figure \ref{fig:distribution-score} illustrates the similarity score distribution relative to time. During the specific intervals depicting the tomato cutting action, the similarity score stabilizes at around 80 or higher. A significant decrease in the score occurs before and after the action, corresponding to frames where the scene’s semantic content is different from the cutting activity.

 This pattern of similarity score fluctuation is consistent across other entries in our dataset. Based on these findings, we identified an empirical threshold of 80 for the CLIP similarity score, beyond which frames can be considered semantically similar to the selected evaluation frames. This threshold, therefore, provides a practical criterion for automated evaluation of frame similarity in video-based action recognition tasks.

\section{Baselines Comparison}

In Table \ref{tab:method-comparison}, we provide method comparisons between \methodname and two closest baselines. \methodname differs from GenHowTo and LEGO by using a lightweight, mask-based inpainting approach that selectively modifies only relevant image regions, reducing computational overhead while maintaining visual-textual alignment through object classification and segmentation. Unlike the baselines, it avoids the need for extensive annotations or large-scale training data, yet supports both action and final-state goals, enabling efficient and accurate instructional image generation.

\begin{table}[]
    \caption{\textbf{Method comparison with the closest baselines.} Having a simpler and computationally cheaper architecture, \methodname can generate qualitative images.}
    \scriptsize
    \centering
    \vspace{2mm}
    \renewcommand{\arraystretch}{1.2}
    \begin{tabularx}{\linewidth}{
    >{\raggedright\arraybackslash}p{2.5cm}
    >{\raggedright\arraybackslash}X
    >{\raggedright\arraybackslash}X
    >{\raggedright\arraybackslash}X
    }
    \toprule
     & \textbf{GenHowTo} & \textbf{LEGO} & \textbf{\methodname} \\
    \hline
    Goal & $f_\text{action}$ \& $f_\text{final}$ & $f_\text{action}$ & $f_\text{action}$ \& $f_\text{final}$ \\ \hline
       
    Data & 200K instructional videos image triplets  & 95K Ego4D image triples and 70K from EK-100 & 11K Ego4D image triplets, 3K from EK-100, and 3K from EGTEA Gaze+ \\ \hline
       
    Annotations & produced by image captioning model & detailed action descriptions tuned by GPT & automatically extracted from the datasets\\ \hline
        
    Key method & diffusion models conditioned on images and text prompts & instruction tuning, incorporating  embeddings into a diffusion model & mask-based inpainting to modify only relevant parts \\ \hline

    Overhead & requires training on large-scale triplets & using finetuned VLLMs increases inference time & modifying relevant areas reduces computations \\ \hline

    Visual-textual alignment & uses image captioning which may not always capture object changes & using enhanced by GPT actions does not always align with visuals & achieved through object classification and segmentation\\

    \bottomrule
        
    \end{tabularx}
    \label{tab:method-comparison}
\end{table}

\section{Further Experiments}

\subsection{Using Explicitly Mentioned Objects}

As mentioned in ~\cref{ssec:llava}, we explore an alternative approach to object masking by parsing the narrations from Ego4D and the actions from EGTEA Gaze+ to extract all nouns, using these as inputs to the masking model.
This method contrasts with relying on relevant objects identified through LLaVA prompts. The analysis focuses on two key aspects:

\begin{enumerate}
    \item Comparing the use of only mentioned objects across the EGTEA Gaze+ and Ego4D datasets.
    \item Assessing the performance of using mentioned objects vs. relevant objects identified by LLaVA on the Ego4D dataset.
\end{enumerate}

As shown in \cref{tab:mentioned_ego_vs_egtea}, the average score increased when only mentioned objects were used in both the EGTEA and Ego4D datasets.
However, this improvement primarily results from a reduction in the number of non-empty masks.
When no mask is generated, the model replicates the initial image, leading to an artificially high score.
Notably, Ego4D exhibits a significantly higher percentage of non-empty masks (50\%) compared to EGTEA~Gaze+ (19\%).
This difference is due to the simpler and clearer object names in Ego4D narrations, which the segmentation model can interpret more effectively.
Conversely, EGTEA Gaze+ includes more complex object labels, such as ``deli container” and ``condiment container” which are more challenging for the model to recognize.

\cref{tab:mentioned_vs_relevant_ego} presents a comparison between using mentioned and relevant objects identified by LLaVA on the Ego4D dataset.
The results show minimal differences in scores when non-empty masks are considered.
However, a significant distinction lies in the proportion of non-empty masks: 86\% for LLaVA-derived objects versus 50\% for mentioned objects.
This indicates that LLaVA is more effective at generating meaningful masks, which leads to more substantial modifications in the generated images.

\begin{table}
    \caption[]{\textbf{Comparison between using only mentioned objects from LLaVA evaluated on Ego4D and EGTEA~Gaze+ evaluation sets with CLIP similarity used as a metric.} Ego4D exhibits a higher percentage of non-empty masks compared to EGTEA~Gaze+. SD stands for Standard deviation and R is the ratio with non-empty masks. \label{tab:mentioned_ego_vs_egtea}}
    \centering
    \footnotesize
    \vspace{2mm}
    \begin{tabular}{cccc}
    \toprule
    Subset & Metric & EGTEA Gaze+ & Ego4D\\
    \midrule
    \multirow{4}{*}{All}
	& CLIP & 83.93 & 83.59\\
	& SD & 7.28 & 13.89\\
	& Quantile $\geq$ 80  &73\% & 75\%\\
	& R &19\% & 50\%\\
    \hline
    \multirow{3}{*}{Non-empty masks}
	&CLIP&82.21 & 78.52\\
	& SD &6.82 & 16.21\\
    & Quantile $\geq$ 80 & 67\% & 58\%\\
    \bottomrule
    \end{tabular}
\end{table}

\begin{table}
    \caption[]{\textbf{Comparison between using only mentioned objects and using relevant objects from LLaVA evaluated on Ego4D evaluation set with CLIP similarity used as a metric.} LLaVA-derived objects produce a meaningful (non-empty) mask more often. SD stands for Standard deviation and R is the ratio with non-empty masks.
    \label{tab:mentioned_vs_relevant_ego}}
    \centering
    \footnotesize
    \vspace{2mm}
    \begin{tabular}{cccc}
    \toprule
     Subset & Metric & Mentioned obj. & LLaVA obj.\\
    \midrule
    \multirow{4}{*}{All}
		& CLIP & 83.59 & 79.68\\
		& SD & 13.89 & 12.57\\
		& Quantile $\geq$ 80  &75\% & 59\%\\
        & R  &50\% & 89\%\\
        \hline
        \multirow{3}{*}{Non-empty masks}
		& CLIP &78.52 & 78.27\\
        & SD  &16.21 & 12.59\\
        & Quantile $\geq$ 80 & 68\% & 56\%\\
    \bottomrule
    \end{tabular}
\end{table}

\begin{figure}
    \centering
    \footnotesize\includegraphics[width=0.45\linewidth]{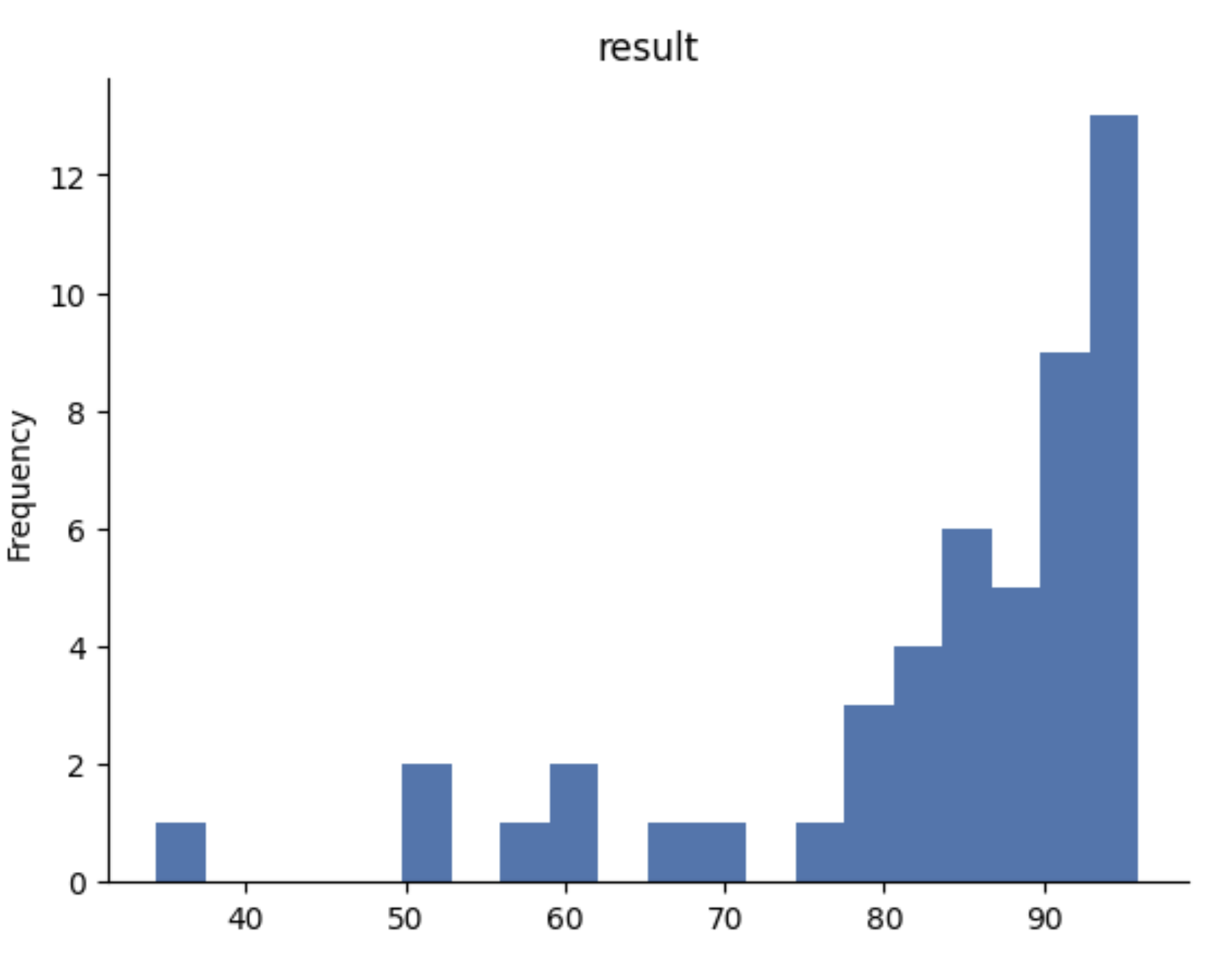}
    \includegraphics[width=0.45\linewidth]{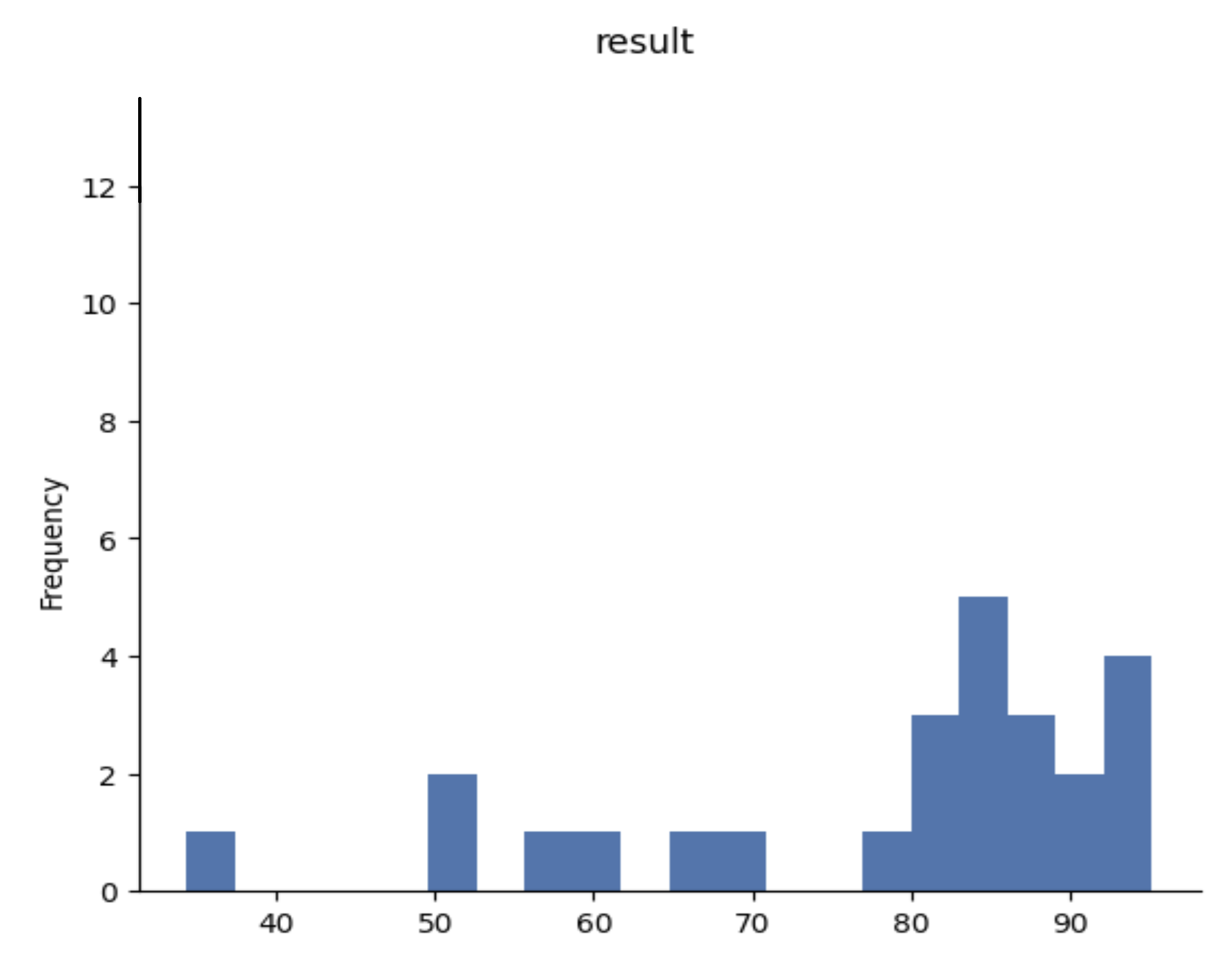}
    \caption{\textbf{CLIP score distribution among all available entries in the evaluation set vs. entries with non-empty masks.} After excluding entries with empty masks, most of the top-scoring results disappear, highlighting that empty masks significantly contribute to the overall score increase.}
    \label{fig:masked-all-vs-non-empty}
\end{figure}

Figure~\ref{fig:masked-all-vs-non-empty} highlights the score distribution for Ego4D when only mentioned objects are used.
The plot differentiates between the entire evaluation set and the subset where the mask is not empty.
The distribution reveals that the perceived improvement in the average score is mainly due to the presence of empty masks, which artificially inflate results.
Once entries with empty masks are excluded, most of the highest-scoring results are no longer present, underscoring the conclusion that empty masks are a major factor in the overall score increase.

\subsection{Empty Masks}
In some cases, due to vague image descriptions, poor image quality, or LLaVA’s limitations in identifying and classifying relevant objects, our approach generates an empty mask. 
This results in the pipeline taking no action, as no mask is provided. 
To avoid such cases, we simply treat the entire input image as positively masked. %
However, due to the specific process of frame filtering, we are guaranteed to see hands in the action frame, and the hands are always functional objects. This way, meaningful changes still happen in the \textit{action frames}.

\subsection{Comparison of Data Curation with LEGO}

\begin{table}
    \caption[]{\textbf{Comparison between our frame selection strategy and LEGO~\cite{lai2024lego}.} For action frames, our method better matches human judgment for representing the action. \label{tab:compare_our_lego_dataset}}
    \centering
    \footnotesize
    \vspace{2mm}
    \begin{tabular}{cccc}
    \hline
     Frame & Score & \methodname & LEGO \\
    \hline
    \multirow{2}{*}{Initial frame}
		 & CLIP & 83.53 & 86.84 \\
        & CLIP $\geq$ 80 & 71 & 81 \\
        \multirow{2}{*}{Action frame}
		 & CLIP & 90.87 & 73.61\\
		&CLIP $\geq$ 80  &92 &29\\
    \hline
    \end{tabular}
\end{table}

\begin{table}
    \caption[]{\textbf{Segmentation model confidence scores for our and LEGO~\cite{lai2024lego} selection strategies.} Our strategy achieves marginally higher confidence scores than LEGO.\label{tab:segmentation_scores}}
    \centering
    \footnotesize
    \vspace{2mm}
    \begin{tabular}{cccc}
    \hline
     Frame & Score & \methodname & LEGO \\
    \hline
    \multirow{2}{*}{Initial frame}
		& objects & 0.44 & 0.42 \\
        & hands & 0.35 & 0.33 \\
        \multirow{2}{*}{Action frame}
		& objects & 0.42 & 0.41\\
		& hands  & 0.39 & 0.30\\
    \hline
    \end{tabular}
\end{table}

LEGO \cite{lai2024lego} employs a similar strategy for selecting frames in scenarios without labeled data, such as in the Ego4D dataset.
Their approach involves choosing the initial frame 0.25 seconds before the action starts and selecting the action frame at 60\% of the action’s duration. However, LEGO’s method does not include final frame selection, so no direct comparison could be made for that aspect in our evaluation.

Table~\ref{tab:compare_our_lego_dataset} compares our frame selection strategy and strategy described in LEGO by computing CLIP similarity between frames selected by the respective strategy and frames picked by a human. As illustrated in the table, the initial frame selection scores using our approach are comparable to those of LEGO’s strategy, with only minor differences observed. In contrast, for action frames, our method shows a clear advantage, with selected frames better matching human judgment for representing the action. This higher alignment with the ground truth underscores the effectiveness of our approach compared to LEGO’s criteria.

Table~\ref{tab:segmentation_scores} presents the confidence scores from the Grounding-DINO~\cite{liu2024groundingdinomarryingdino} model for detecting relevant objects and hands in datasets curated with both our strategy and LEGO’s selection method. These confidence scores were assessed for initial and action frames.

For initial frames, our strategy achieves marginally higher confidence scores than LEGO’s for both object and hand detections, indicating that our approach slightly improves the clarity of relevant object and hand identification at the start of actions. The difference is more pronounced in action frames, particularly for hand detection. This higher confidence is advantageous, as hands play a critical role in executing actions and are a key focus in egocentric video datasets.

\begin{figure*}
    \centering \footnotesize 
    \begin{tabular}{c c c c}
        \multicolumn{2}{c}{$f_{\text{in}}$} & \multicolumn{2}{c}{$f_{\text{action}}$} \\
        \methodname & LEGO & \methodname & LEGO \\
        \includegraphics[width=0.2\linewidth]{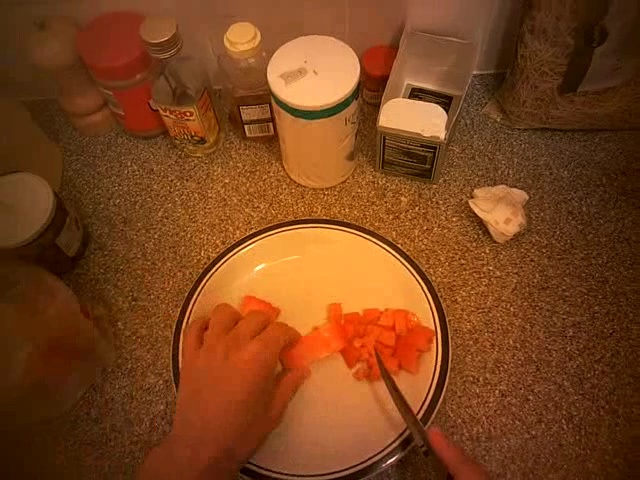} & \includegraphics[width=0.2\linewidth]{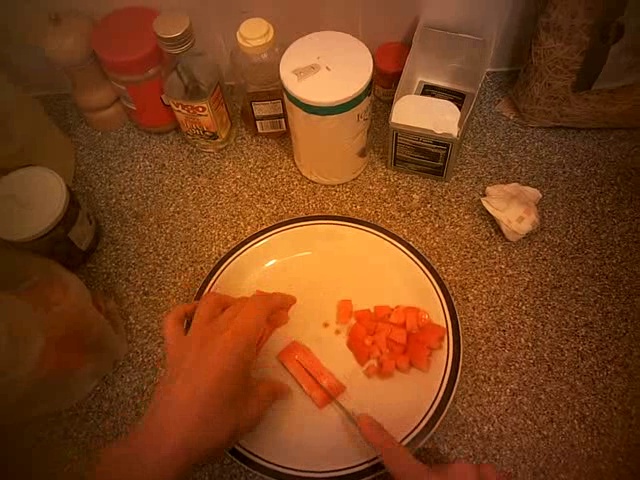} & \includegraphics[width=0.2\linewidth]{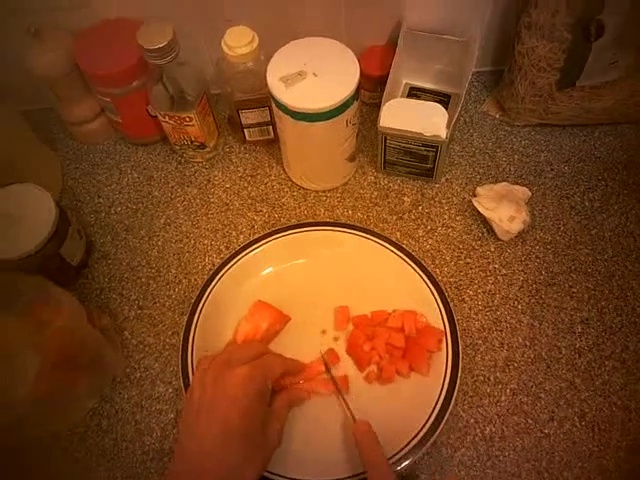} & \includegraphics[width=0.2\linewidth]{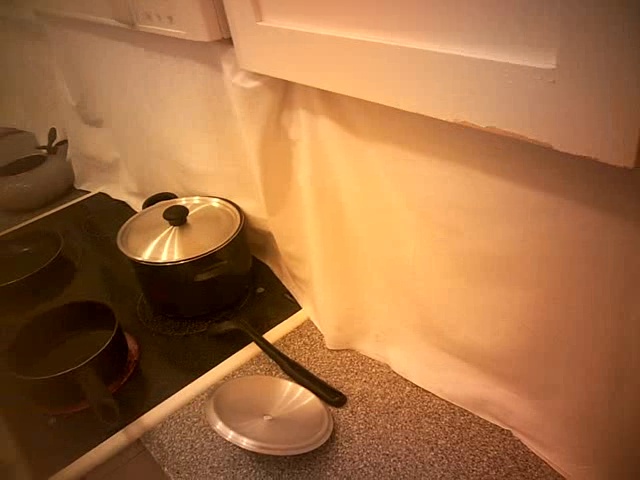} \\

        \includegraphics[width=0.2\linewidth]{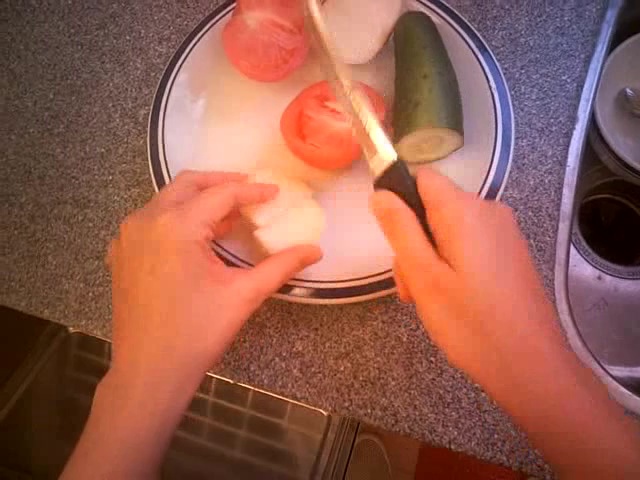} & \includegraphics[width=0.2\linewidth]{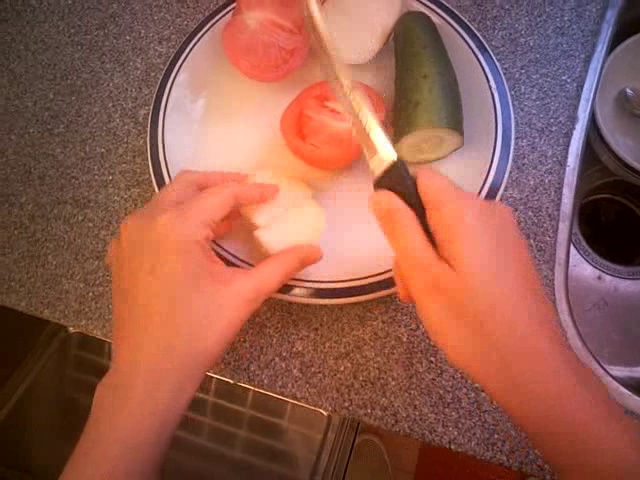} & \includegraphics[width=0.2\linewidth]{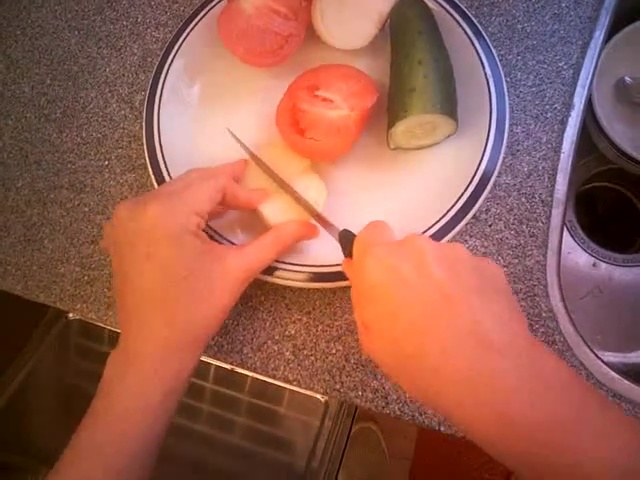} & \includegraphics[width=0.2\linewidth]{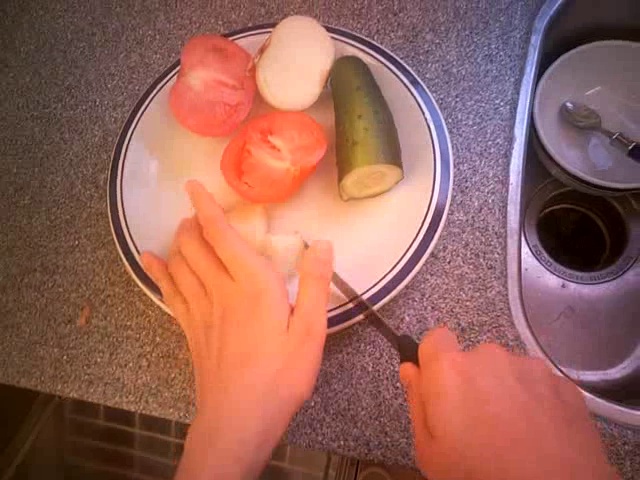} \\

        \includegraphics[width=0.2\linewidth]{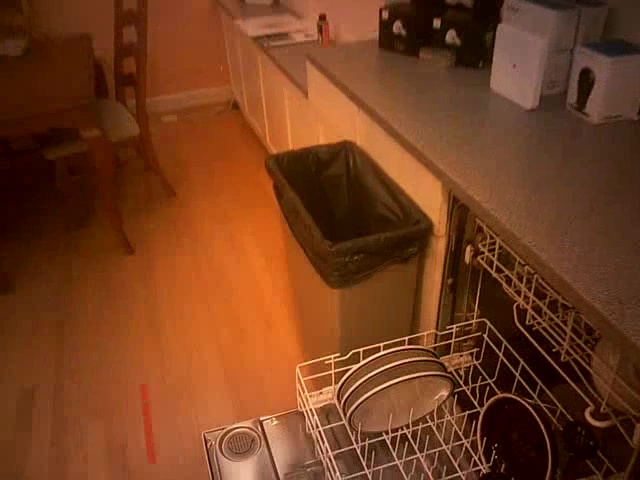} & \includegraphics[width=0.2\linewidth]{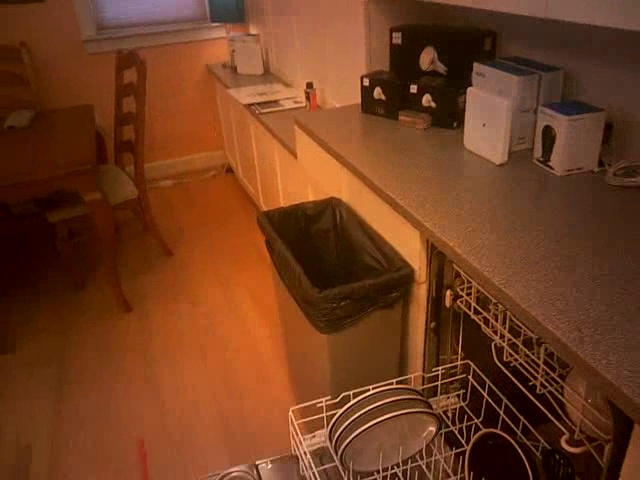} & \includegraphics[width=0.2\linewidth]{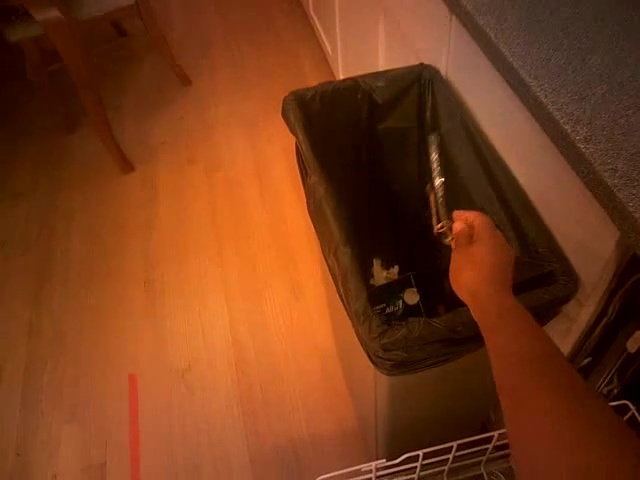} & \includegraphics[width=0.2\linewidth]{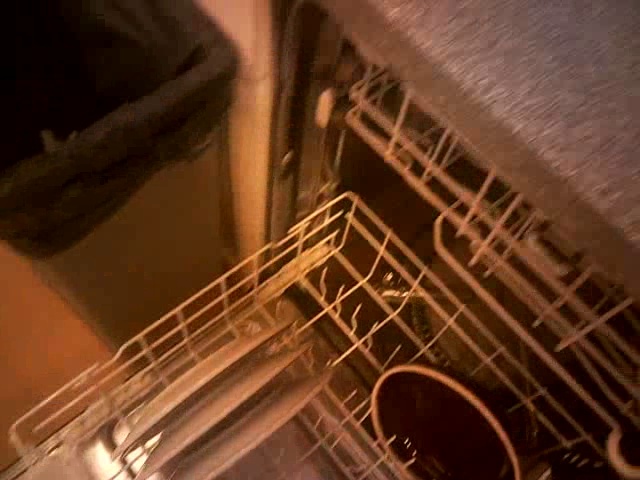} \\

         \includegraphics[width=0.2\linewidth]{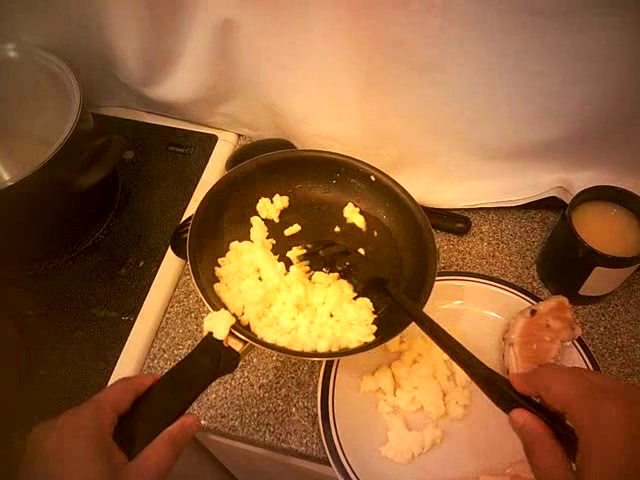} & \includegraphics[width=0.2\linewidth]{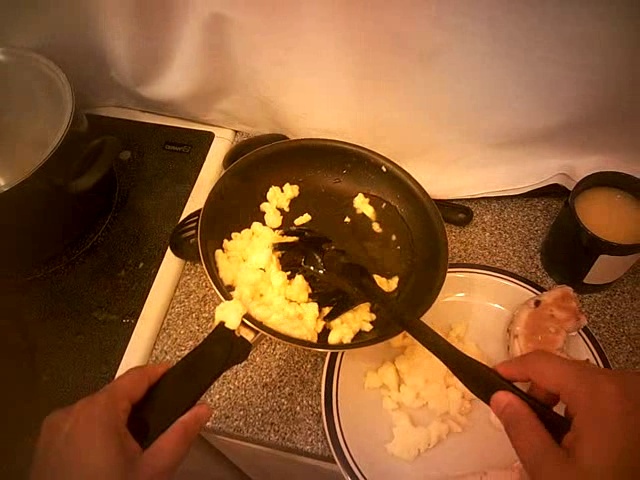} & \includegraphics[width=0.2\linewidth]{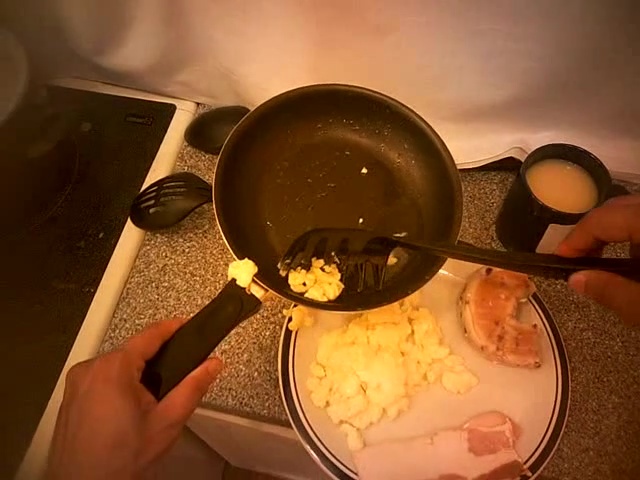} & \includegraphics[width=0.2\linewidth]{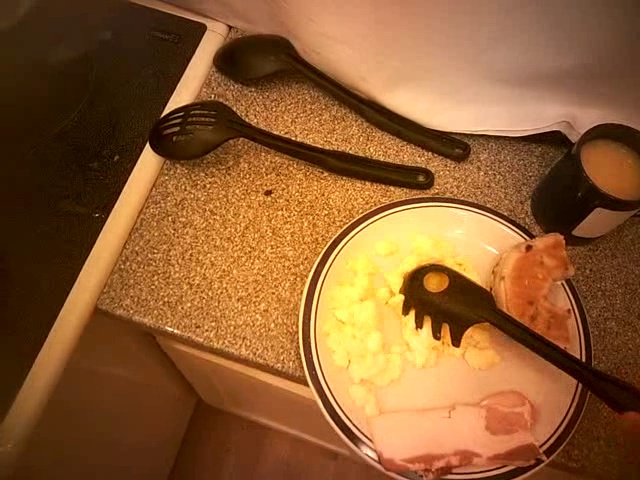} \\
        
    \end{tabular}
    \caption{\textbf{Qualitative comparison between initial and action frames selected from EGTEA~Gaze+ by \methodname and LEGO~\cite{lai2024lego} strategies.} Action frames selected according to the LEGO strategy often depict a state that does not represent the action process, but rather final states or transitions to the next activities.}
    \label{fig:selection-comparison-we-vs-lego}
\end{figure*}

Figure~\ref{fig:selection-comparison-we-vs-lego} illustrates the qualitative comparison between images selected from EGTEA~Gaze+ according to \methodname and LEGO strategies. Even though the initial frames selected by both methods are almost identical, the difference between selected action frames is significant. Selecting a frame after 60\% of the action (acc. to LEGO~\cite{lai2024lego}) is not always suitable to illustrate an action frame since in some cases this frame shows the final state or transition to the next activity.

\subsection{Ablation: Joint Training}

Table~\ref{tab:joint_training} evaluates the model performance when jointly trained on both action and final state frames. We fine-tuned the diffusion pipeline on the EGTE~Gaze+ training set for \textit{action} and \textit{final} frames both separately and jointly. Each model is evaluated on the selected set of 650 image triplets from EGTEA~Gaze+. We observe a slight drop in the metrics when the evaluation is complete with the joint model.

\begin{table}[]
    \caption{\textbf{Ablation of joint action and final frames training on EGTEA~Gaze+ dataset.} We observe a slight drop in values when using a jointly trained model instead of two separate models.}
    \vspace{2mm}
    \footnotesize
    \centering
    \begin{tabular}{ccccc}
        \hline
       \multirow{2}{*}{Method} & \multicolumn{2}{c}{Action frames} & \multicolumn{2}{c}{Final state frames}\\
       & CLIP $\uparrow$ & D-CLIP $\downarrow$ & CLIP $\uparrow$ & D-CLIP $\downarrow$ \\ %
       \hline
       Only actions & \textbf{67.31} & \textbf{15.19} & - & -\\
       Only final states & - & - & \textbf{66.90} & \textbf{13.54} \\
       Joint training & 66.76 &  15.75 & 66.38 & 14.28\\
       \hline
    \end{tabular}
    \label{tab:joint_training}
\end{table}

\begin{table}[]
\footnotesize
    \caption{\textbf{Ablation of object category.} We investigate how different object masks impact the quality of generated action images. We observe a drop in scores when we use functional objects which is the result of not having a fine-tuning procedure specific for them.}
    \centering
    \vspace{2mm}
    \begin{tabular}{cccc}
        \hline
       Method & CLIP $\uparrow$ & M-CLIP $\uparrow$ & D-CLIP $\downarrow$ \\ %
       \hline 
       Only core & 72.92 & \textbf{64.74} & 13.99 \\
       Core + location & \textbf{73.95} & 63.03 & \textbf{12.71} \\
       (Core + functional) + location  & 70.90 & 62.68 & 16.24 \\
       \methodname & 72.15 & 64.46 & 14.97\\
    \hline
    \end{tabular}
    \label{tab:masks}
\end{table}

\subsection{Ablation: Object Categories}
In \cref{tab:masks}, we ablate the type of object masks used in \methodname to analyze their influence. \textit{Only core} uses only the masks of core objects in inpainting, 
\textit{Core + Location} ignores the usage of functional objects, and \textit{(Core + functional) + location} uses all object types, but does not distinguish between core and functional objects, and uses their masks together in inpainting.

We use the same fine-tuned diffusion model, however, we exclude specific steps in different methods. Evaluation is complete on the pre-selected EGTEA~Gaze+ evaluation dataset. Since functional objects are only used in generating action frames, this ablation study was complete only for action frames.
Note that this pre-selected evaluation dataset has highly representative frames for both the action frame and the final state frame. The obtained results achieve better evaluation results. 

\begin{figure}
    \centering
    \footnotesize
    \includegraphics[width=0.3\linewidth]{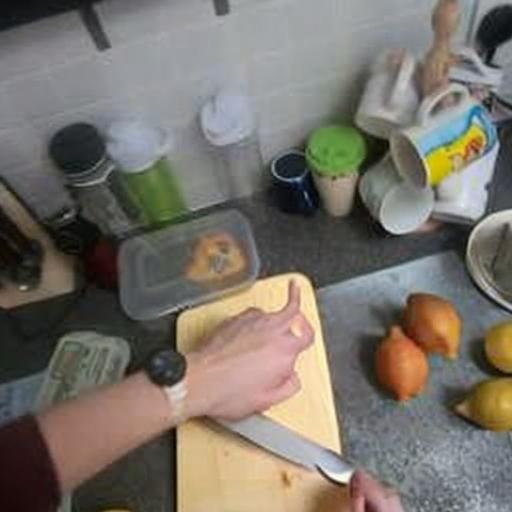}
    \includegraphics[width=0.3\linewidth]{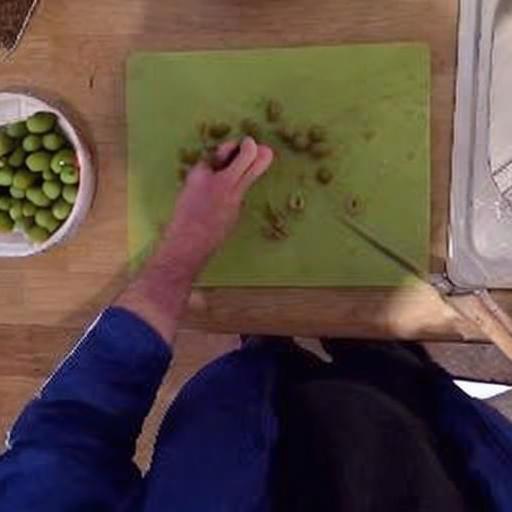}
    \includegraphics[width=0.3\linewidth]{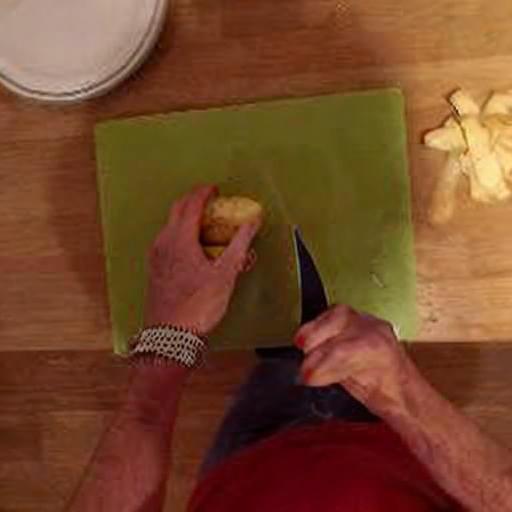}
    \caption{\textbf{Examples of lower consistency in processing functional objects.} Since we do not have a specific training procedure for functional objects, we observe an inconsistency in hand generation.}
    \label{fig:hands-inconsistency}
\end{figure}

We observe that not using functional objects at all is more beneficial than using them. However, even though the average scores signal this drop, this occurs due to lower consistency in processing functional objects in some of the cases (see \cref{fig:hands-inconsistency} for examples). This effect is emphasized in this pre-selected evaluation dataset as the reference action frames are of high quality. Since we do not have a specific training procedure for functional objects, we observe that some hand or object positions are not realistic in the generated frame. This is a limitation of our model that can be improved. 

\subsection{Generalization Failure}

\begin{figure}
    \footnotesize
    \centering
    \includegraphics[width=0.45\linewidth]{images/test-diff-actions/input.jpg}
    \includegraphics[width=0.45\linewidth]{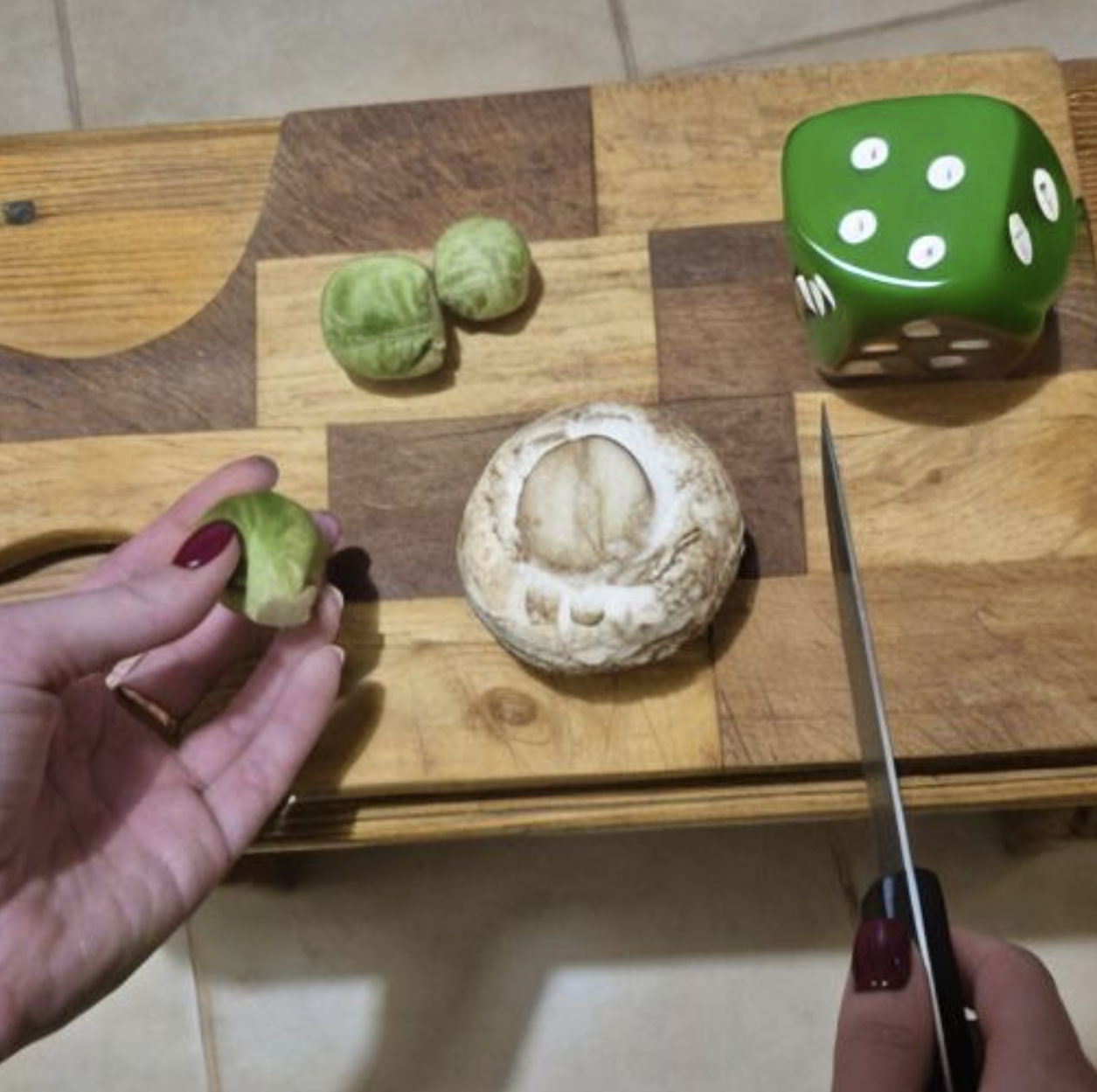}
    \caption{\textbf{Generation failure case for the ``dice tomato" action.} The model was trained on the dataset where action ``dice" did not exist, therefore, the generated image was based on the knowledge previously incorporated in the diffusion pipeline.}
    \label{fig:failure}
\end{figure}

When assessing the generalization ability of our model, we encounter cases where the input action we provide has never occurred in the training set. In this case, the generated result depends on the data this diffusion pipeline was pre-trained on. Figure~\ref{fig:failure} illustrates an example when the output for the action ``dice tomato" did not meet the expectations because the model was trained on the set (EGTEA~Gaze+ training set) that did not include the ``dice" action.

\section{Further Qualitative Results}

\begin{figure*}
    \centering
    \footnotesize
    \begin{tabular}{cc@{\hskip 10pt}c@{\hskip 2pt}c@{\hskip 2pt}c@{\hskip 2pt}c}
         &  \\
        & Input & IP2P & GenHowTo & LEGO & \methodname \\

        \multirow{6}{*}{\rotatebox{90}{Ego4D}}& \includegraphics[width=0.15\textwidth]{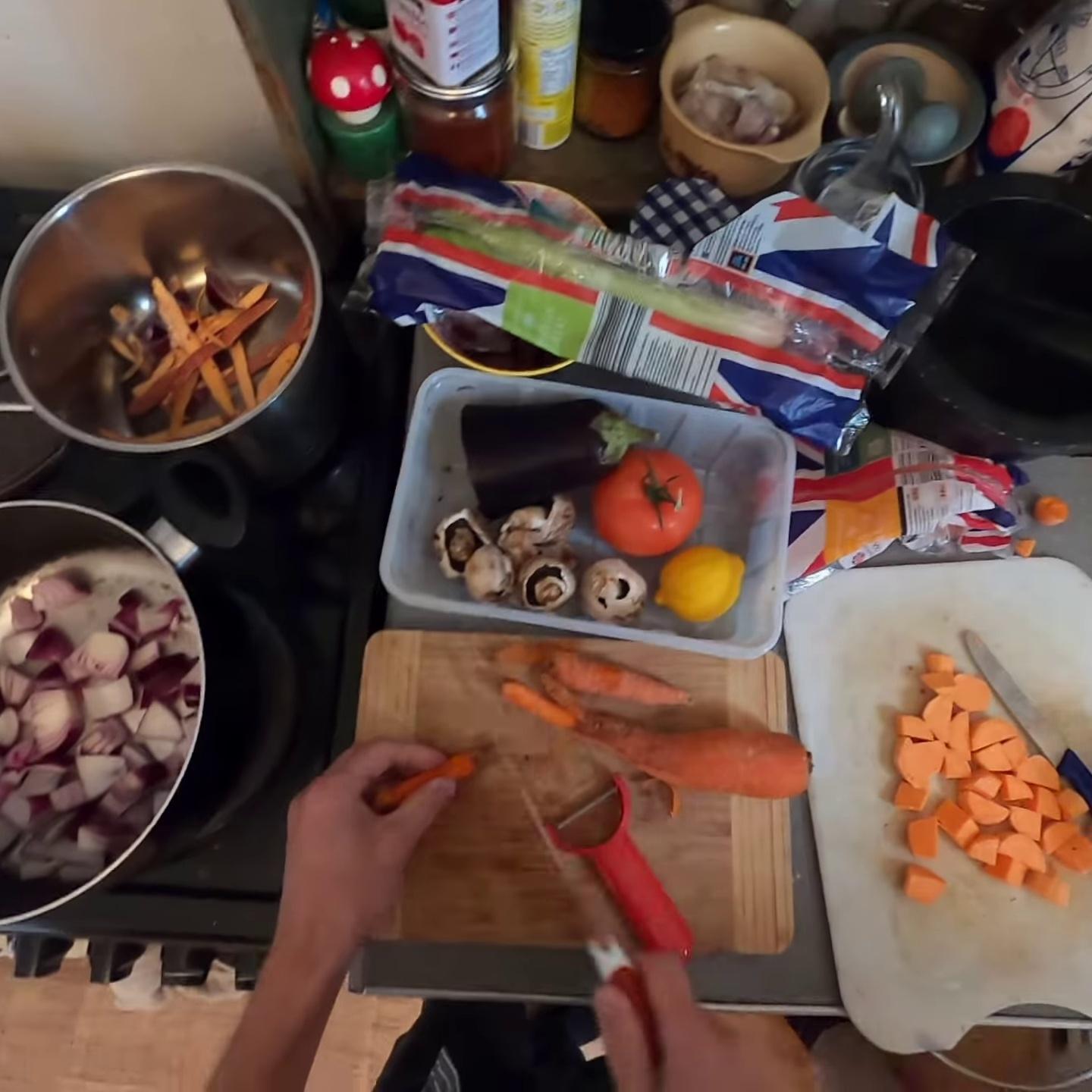}  & \includegraphics[width=0.15\textwidth]{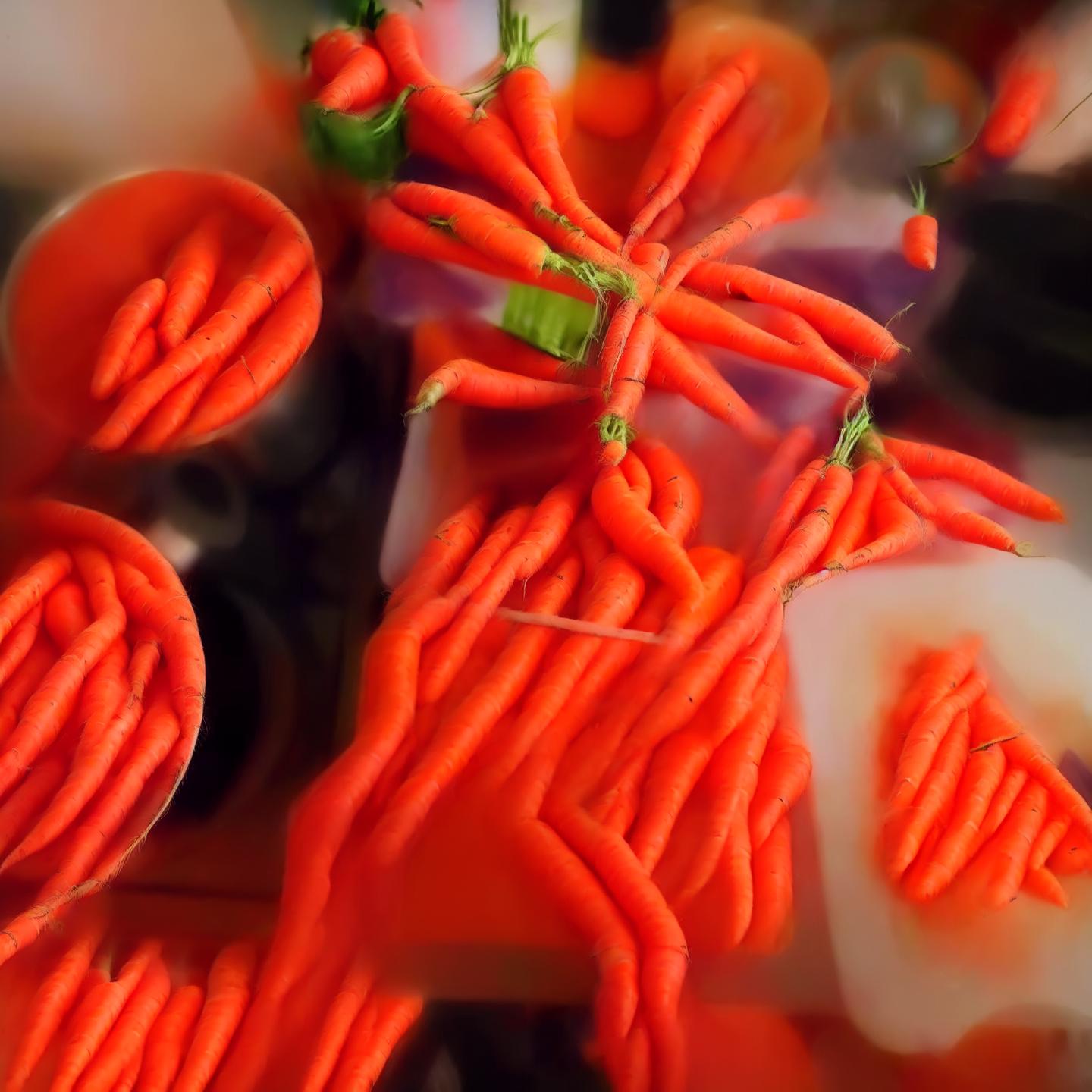} & \includegraphics[width=0.15\textwidth]{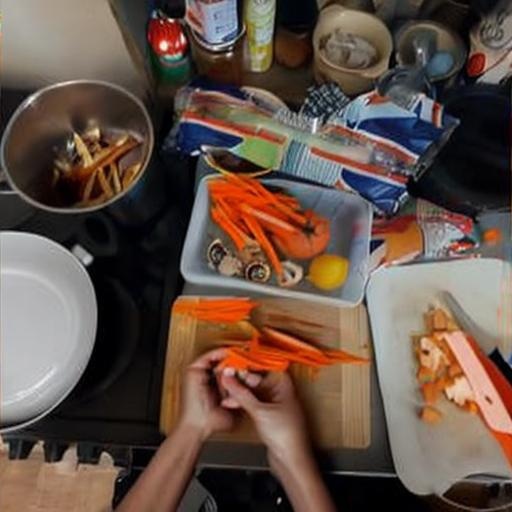}& \includegraphics[width=0.15\textwidth]{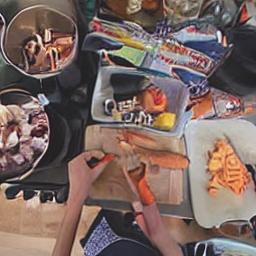} & \includegraphics[width=0.15\textwidth]{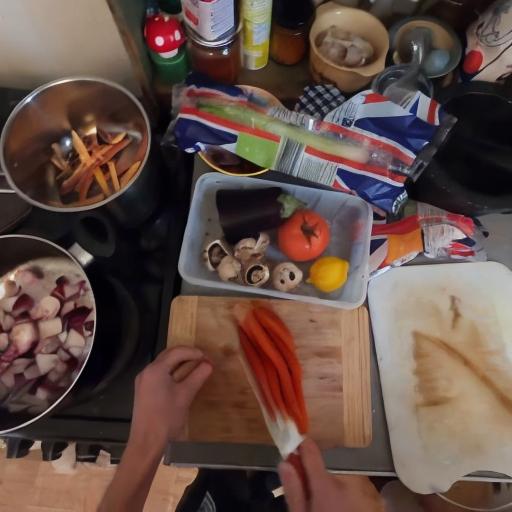} \\
        & Cut carrot \\
        & \includegraphics[width=0.15\textwidth]{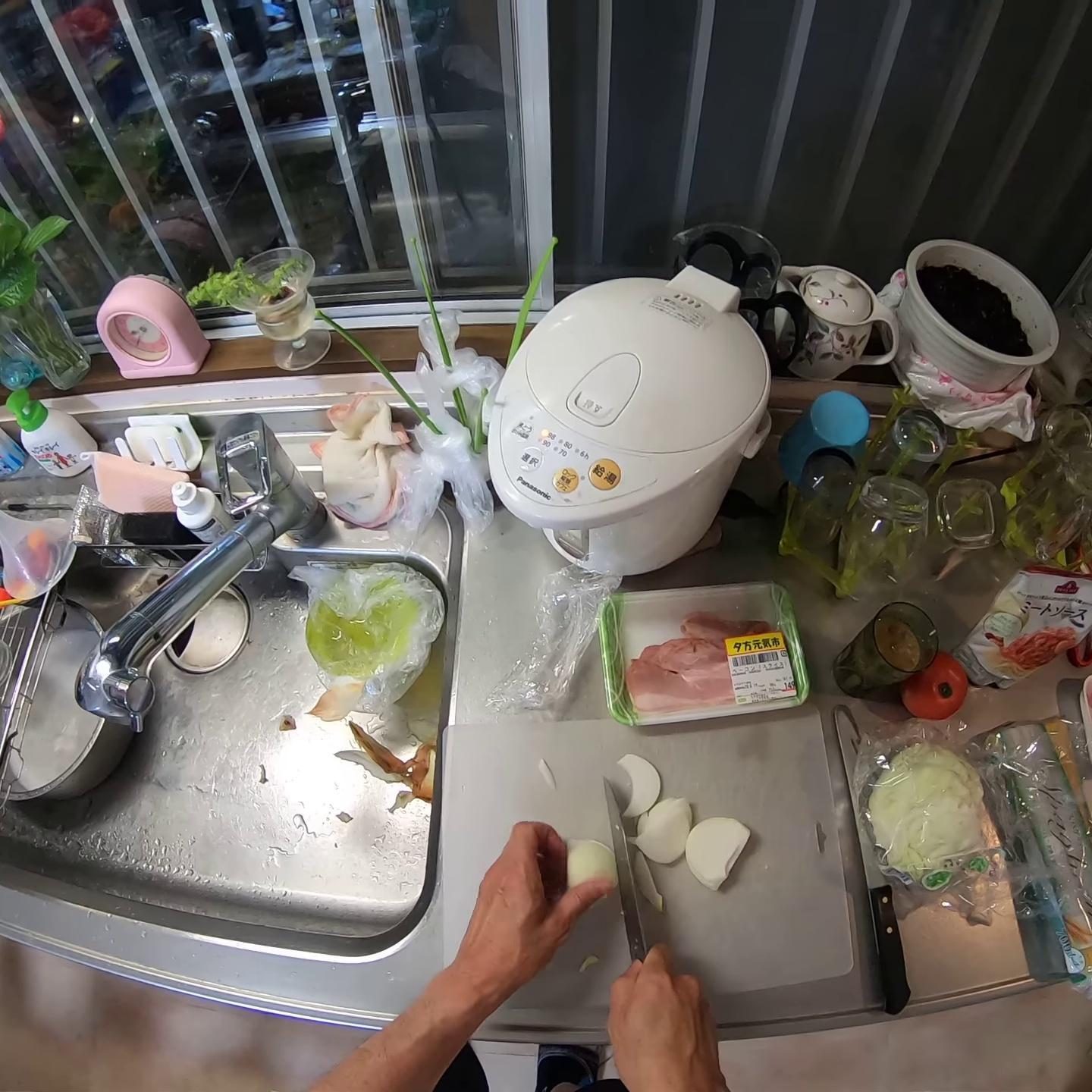} & \includegraphics[width=0.15\textwidth]{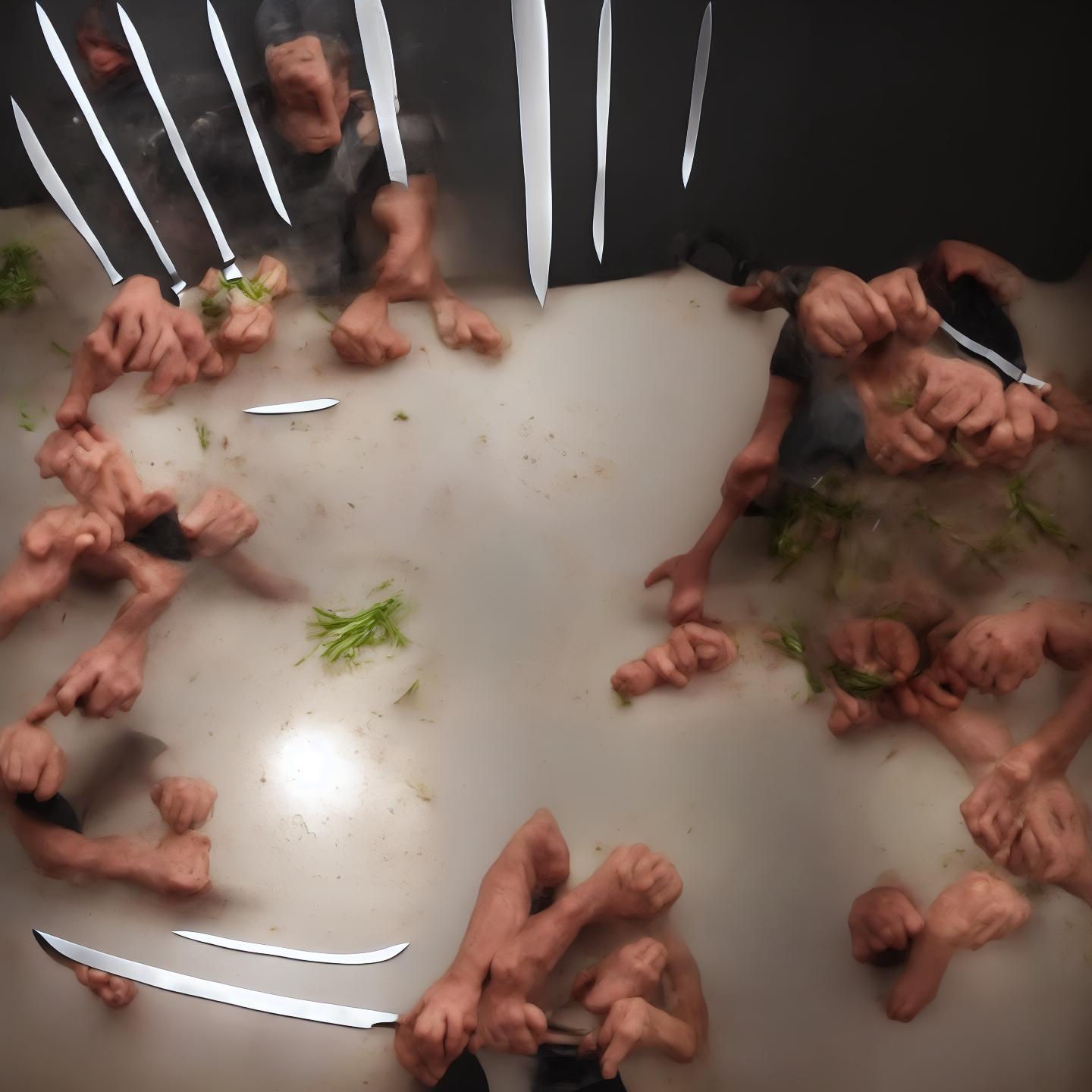} & \includegraphics[width=0.15\textwidth]{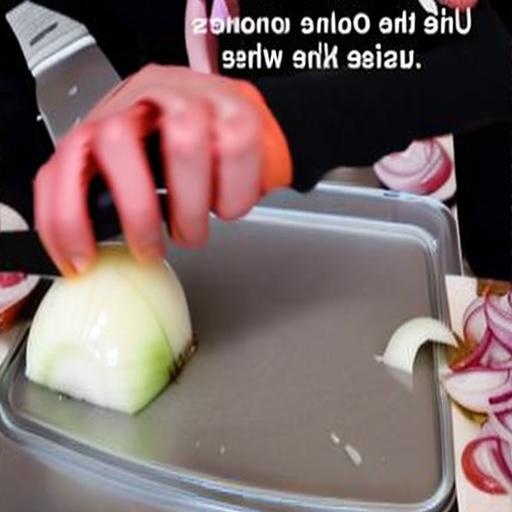}& \includegraphics[width=0.15\textwidth]{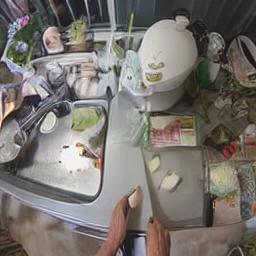} & \includegraphics[width=0.15\textwidth]{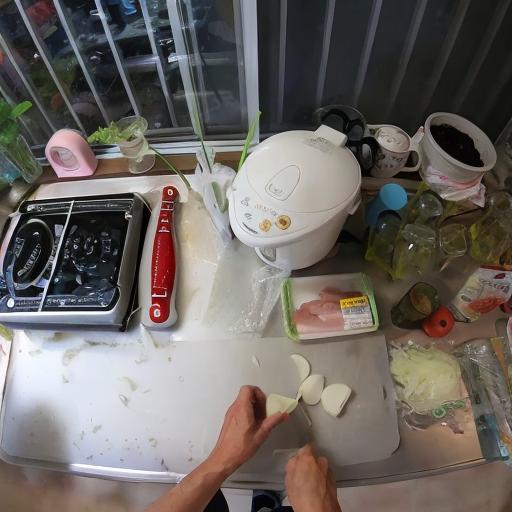} \\
        & Cut onion \\

        & \includegraphics[width=0.15\textwidth]{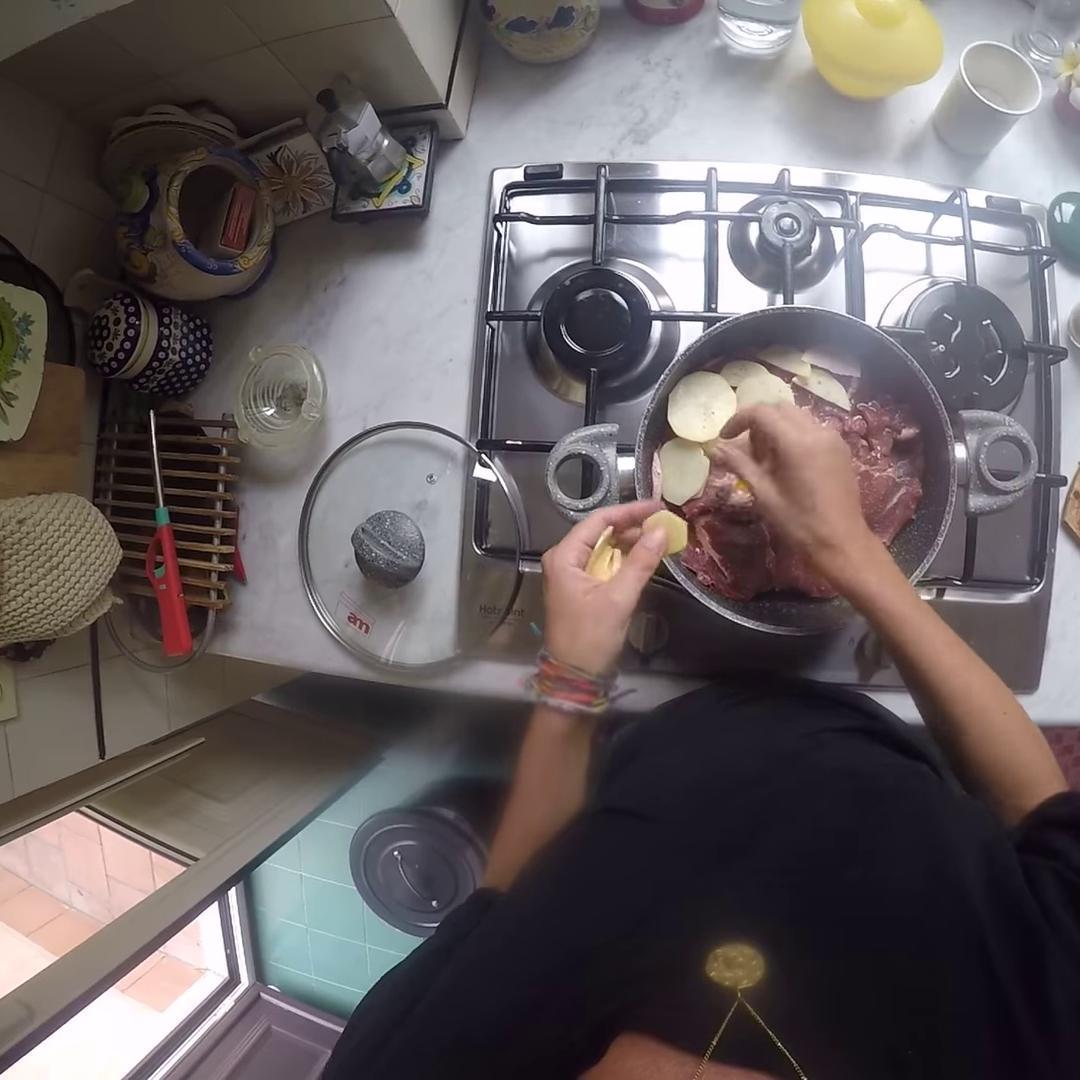} & \includegraphics[width=0.15\textwidth]{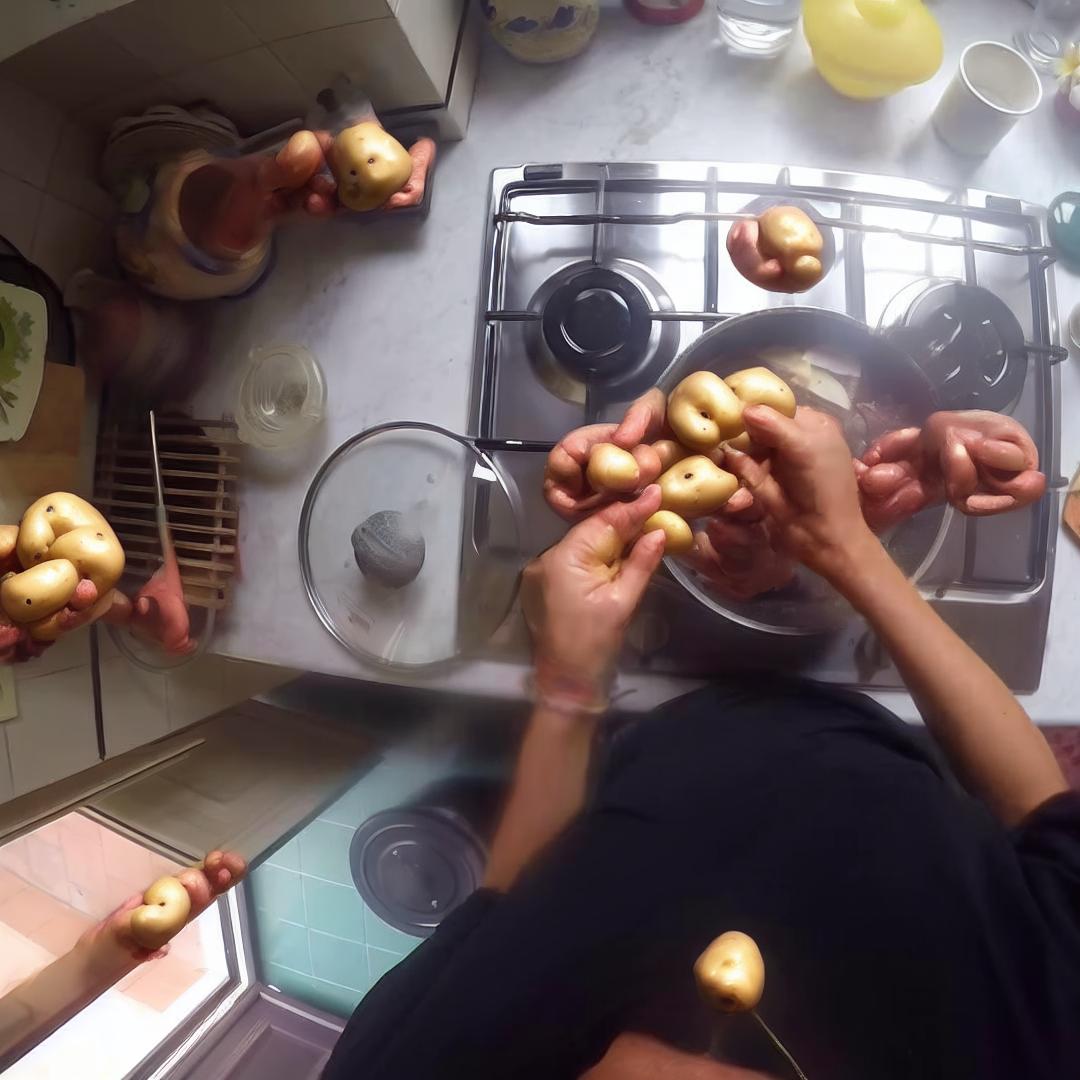} & \includegraphics[width=0.15\textwidth]{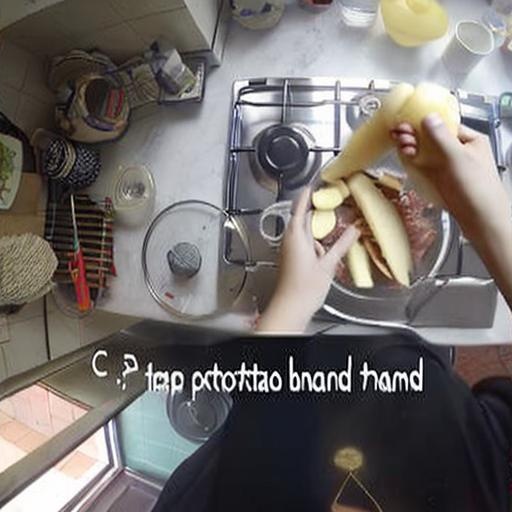}& \includegraphics[width=0.15\textwidth]{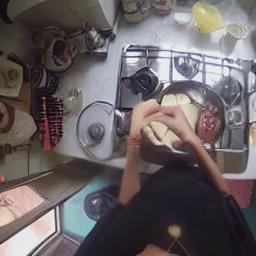} & \includegraphics[width=0.15\textwidth]{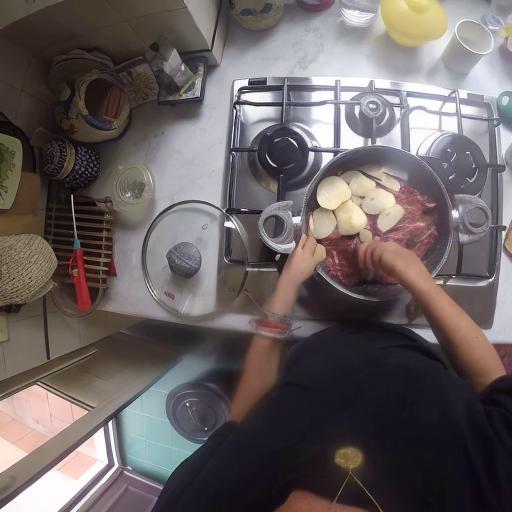} \\
        & Pull potato \\
        
        \multirow{6}{*}{\rotatebox{90}{EGTEA Gaze+}}
        & \includegraphics[width=0.15\textwidth]{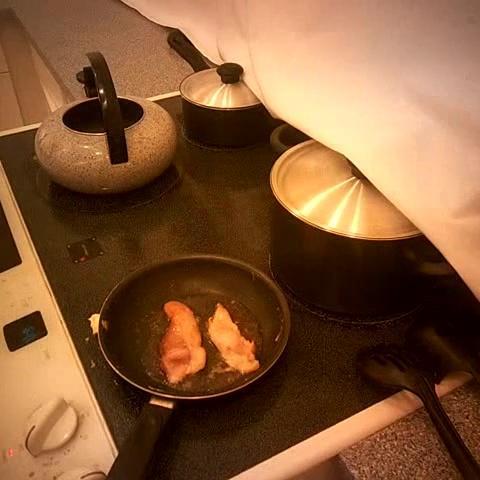} & \includegraphics[width=0.15\textwidth]{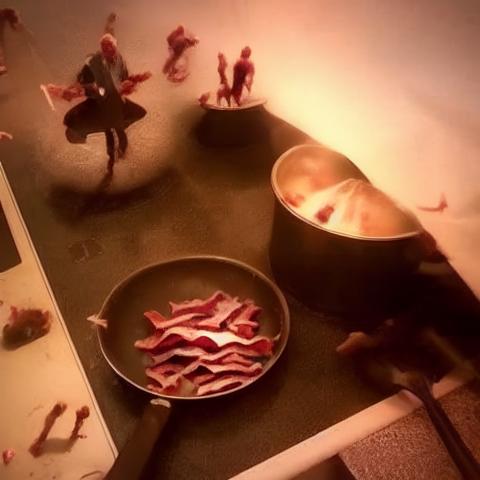} & \includegraphics[width=0.15\textwidth]{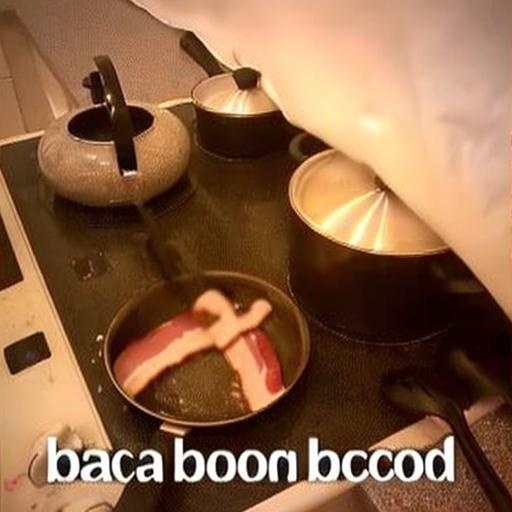}& \includegraphics[width=0.15\textwidth]{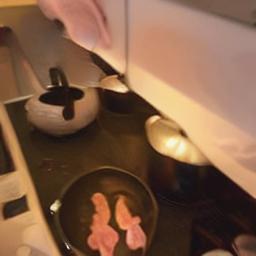} & \includegraphics[width=0.15\textwidth]{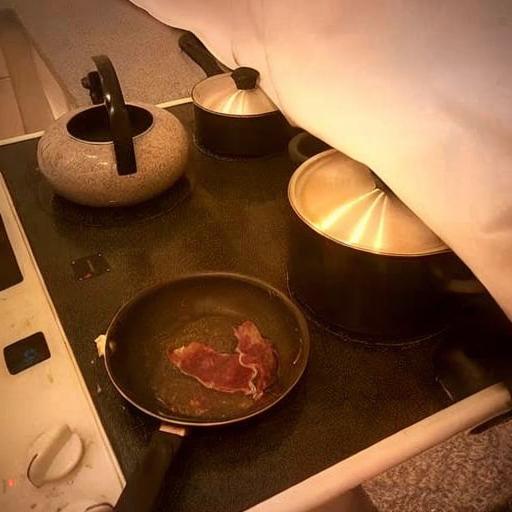} \\

        & Move bacon \\

        & \includegraphics[width=0.15\textwidth]{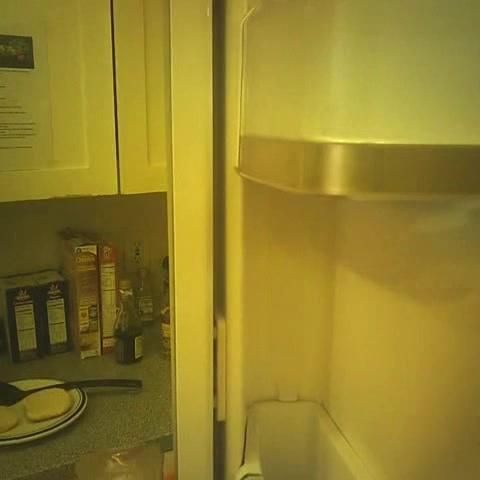} & \includegraphics[width=0.15\textwidth]{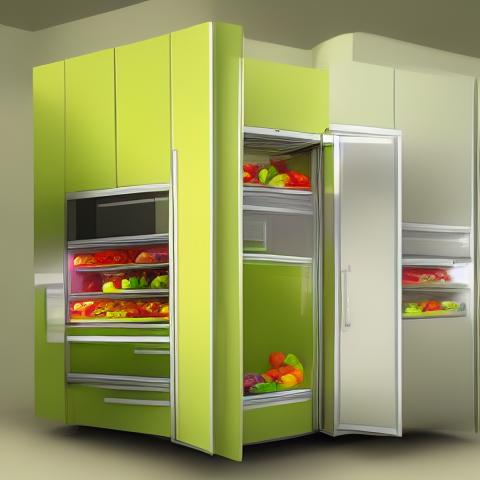} & \includegraphics[width=0.15\textwidth]{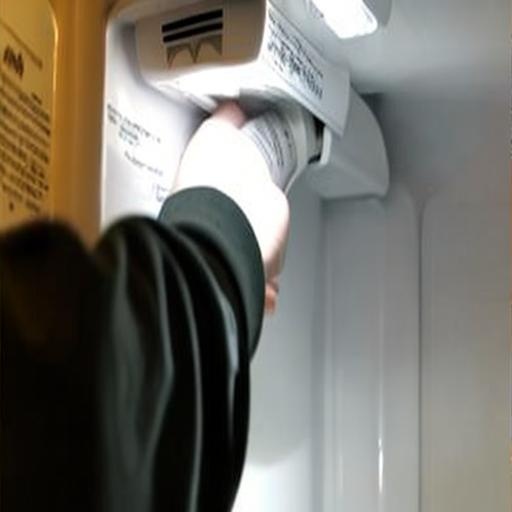}& \includegraphics[width=0.15\textwidth]{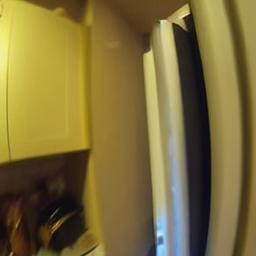} & \includegraphics[width=0.15\textwidth]{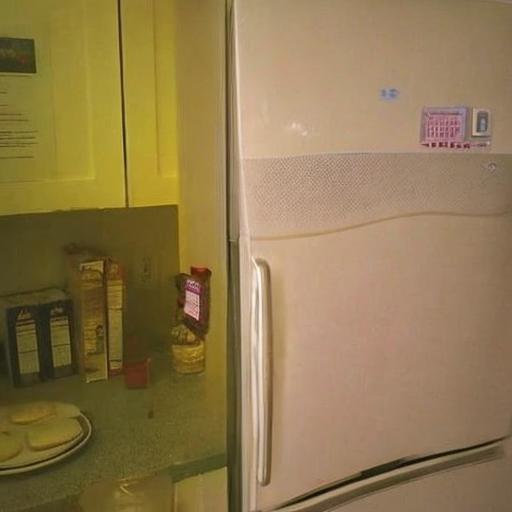} \\ 
        & Close fridge \\

        & \includegraphics[width=0.15\textwidth]{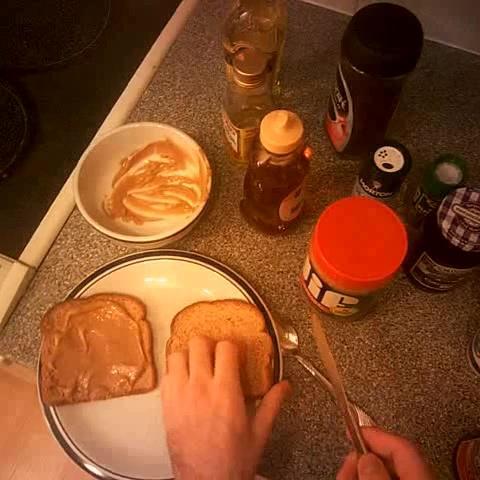} & \includegraphics[width=0.15\textwidth]{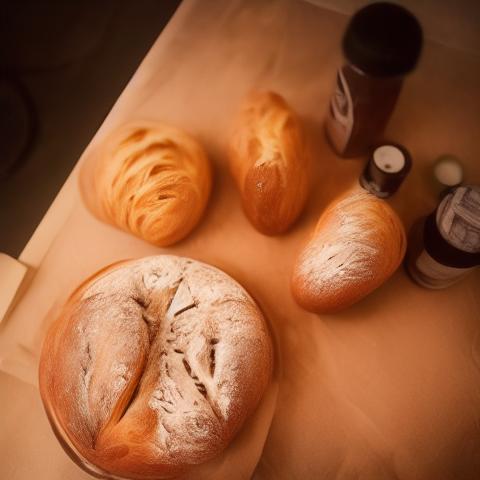} & \includegraphics[width=0.15\textwidth]{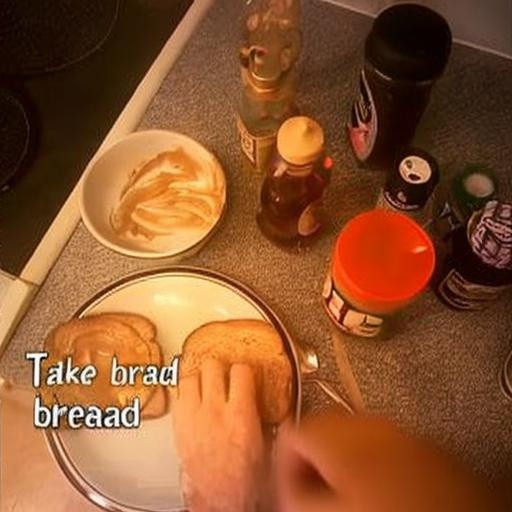}& \includegraphics[width=0.15\textwidth]{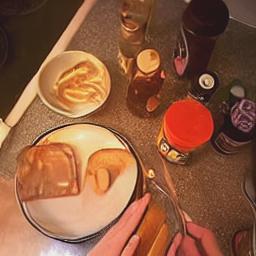} & \includegraphics[width=0.15\textwidth]{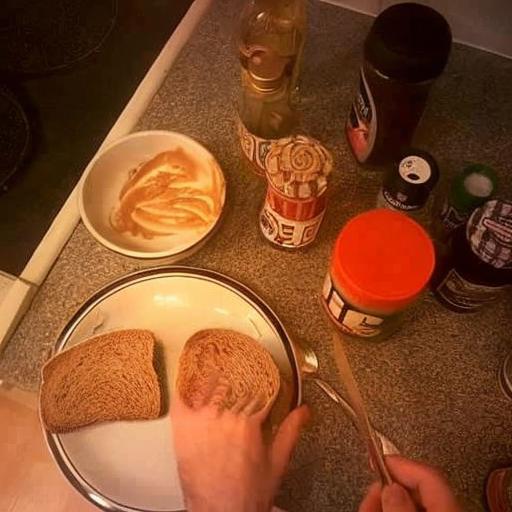} \\
        & Take bread \\

        \multirow{6}{*}{\rotatebox{90}{EK-100}}
        & \includegraphics[width=0.15\textwidth]{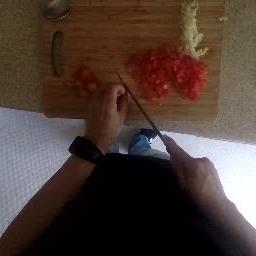} & \includegraphics[width=0.15\textwidth]{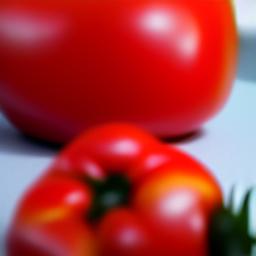} & \includegraphics[width=0.15\textwidth]{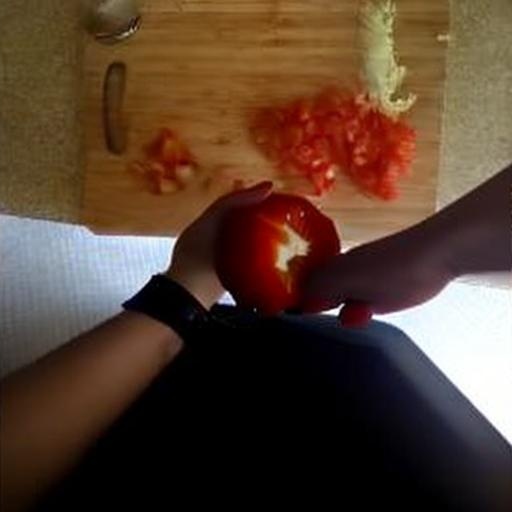}& \includegraphics[width=0.15\textwidth]{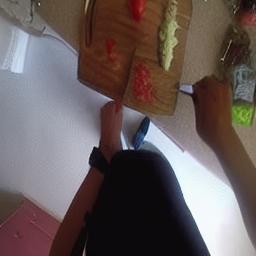} & \includegraphics[width=0.15\textwidth]{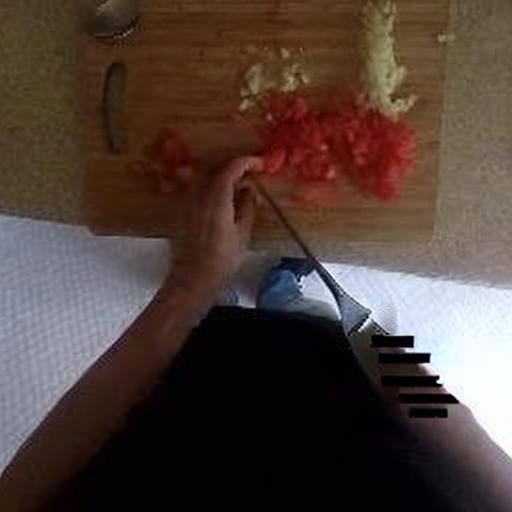} \\
        & Cut tomato    \\
        & \includegraphics[width=0.15\textwidth]{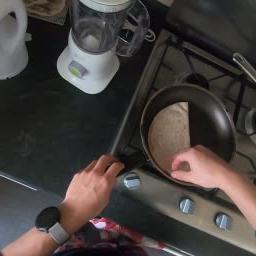} & \includegraphics[width=0.15\textwidth]{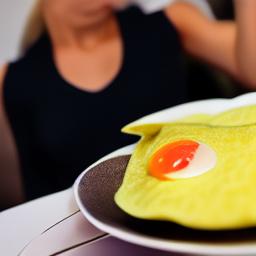} & \includegraphics[width=0.15\textwidth]{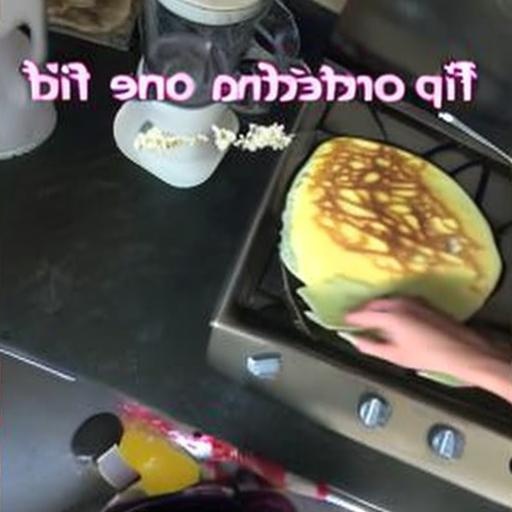}& \includegraphics[width=0.15\textwidth]{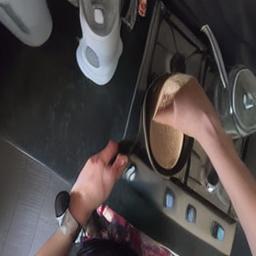} & \includegraphics[width=0.15\textwidth]{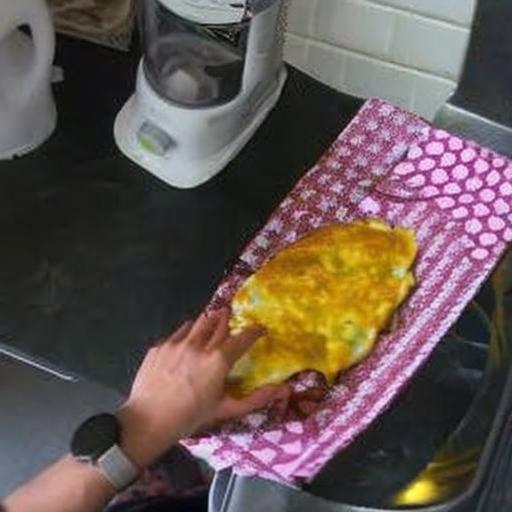} \\
        & Flip omelette \\

        & \includegraphics[width=0.15\textwidth]{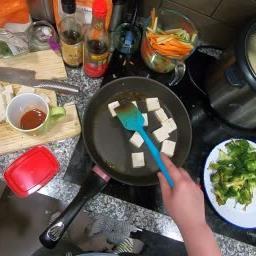} & \includegraphics[width=0.15\textwidth]{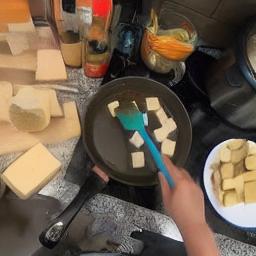} & \includegraphics[width=0.15\textwidth]{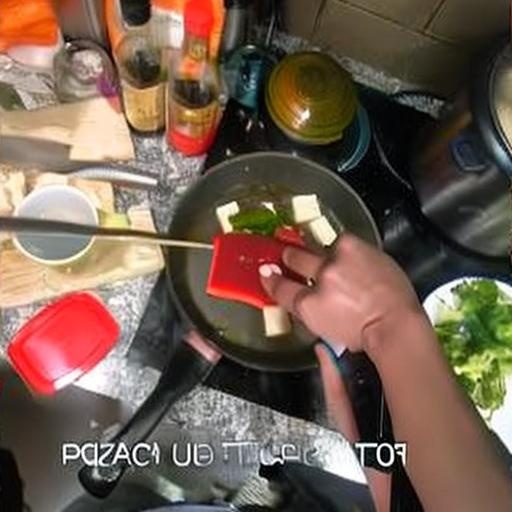}& \includegraphics[width=0.15\textwidth]{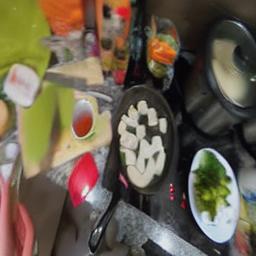} & \includegraphics[width=0.15\textwidth]{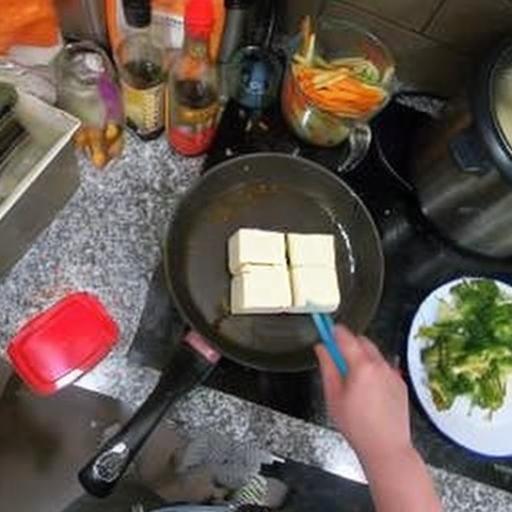} \\
        & Pick up tofu \\

    \end{tabular}
    \vspace{-0.7em}
    \caption{\textbf{Further qualitative comparison of action frames with related work.} \methodname has the best performance in aligning generated action images.}
    \label{fig:qualitative-comparison-further-action}
    \vspace{-3mm}
\end{figure*}

\begin{figure*}
    \centering
    \footnotesize
    \begin{tabular}{cc@{\hskip 10pt}c@{\hskip 2pt}c@{\hskip 2pt}c}
         & Input & IP2P & GHT & \methodname\\

        \multirow{6}{*}{\rotatebox{90}{Ego4D}}& \includegraphics[width=0.15\textwidth]{images/suppl/qualitative/ego-3-start-cut-carrot-6449.jpg} & \includegraphics[width=0.15\textwidth]{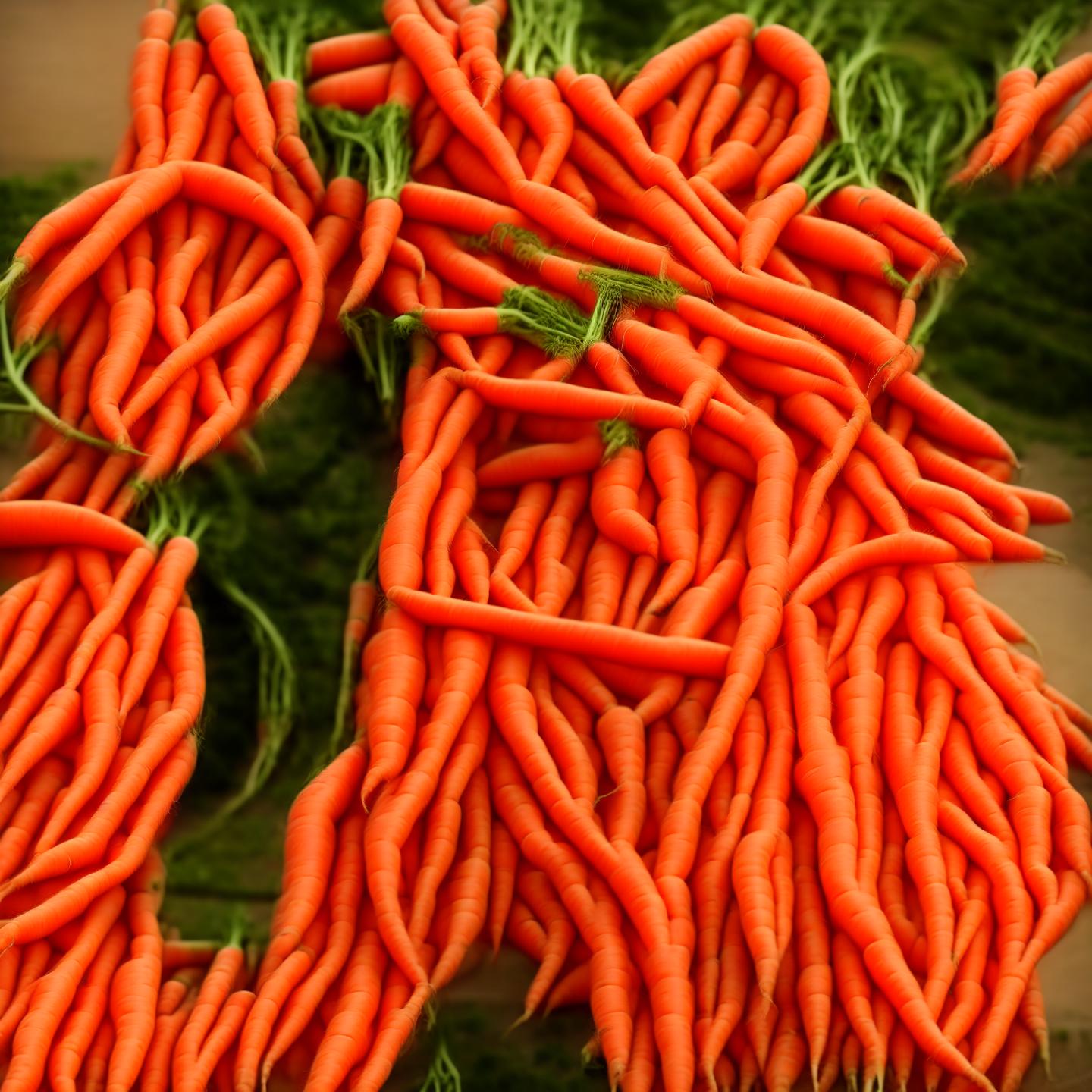} & \includegraphics[width=0.15\textwidth]{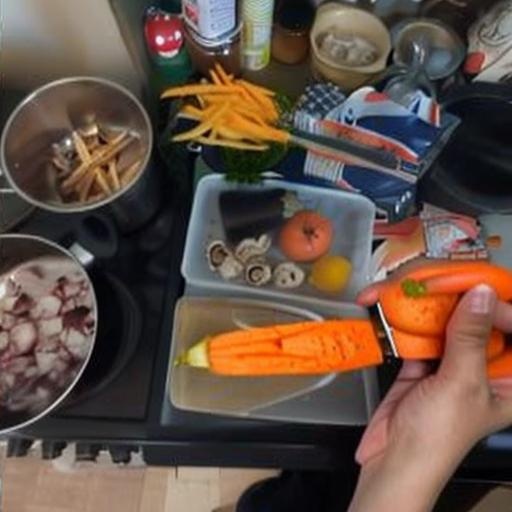} & \includegraphics[width=0.15\textwidth]{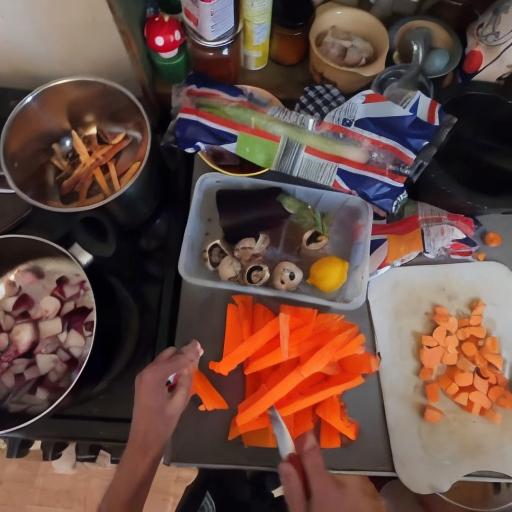} \\
        & Cut carrot \\
        & \includegraphics[width=0.15\textwidth]{images/suppl/qualitative/ego-4-304735ba-6bf5-4d39-bcb5-0dabddb11d68_0000009838_start.jpg} & \includegraphics[width=0.15\textwidth]{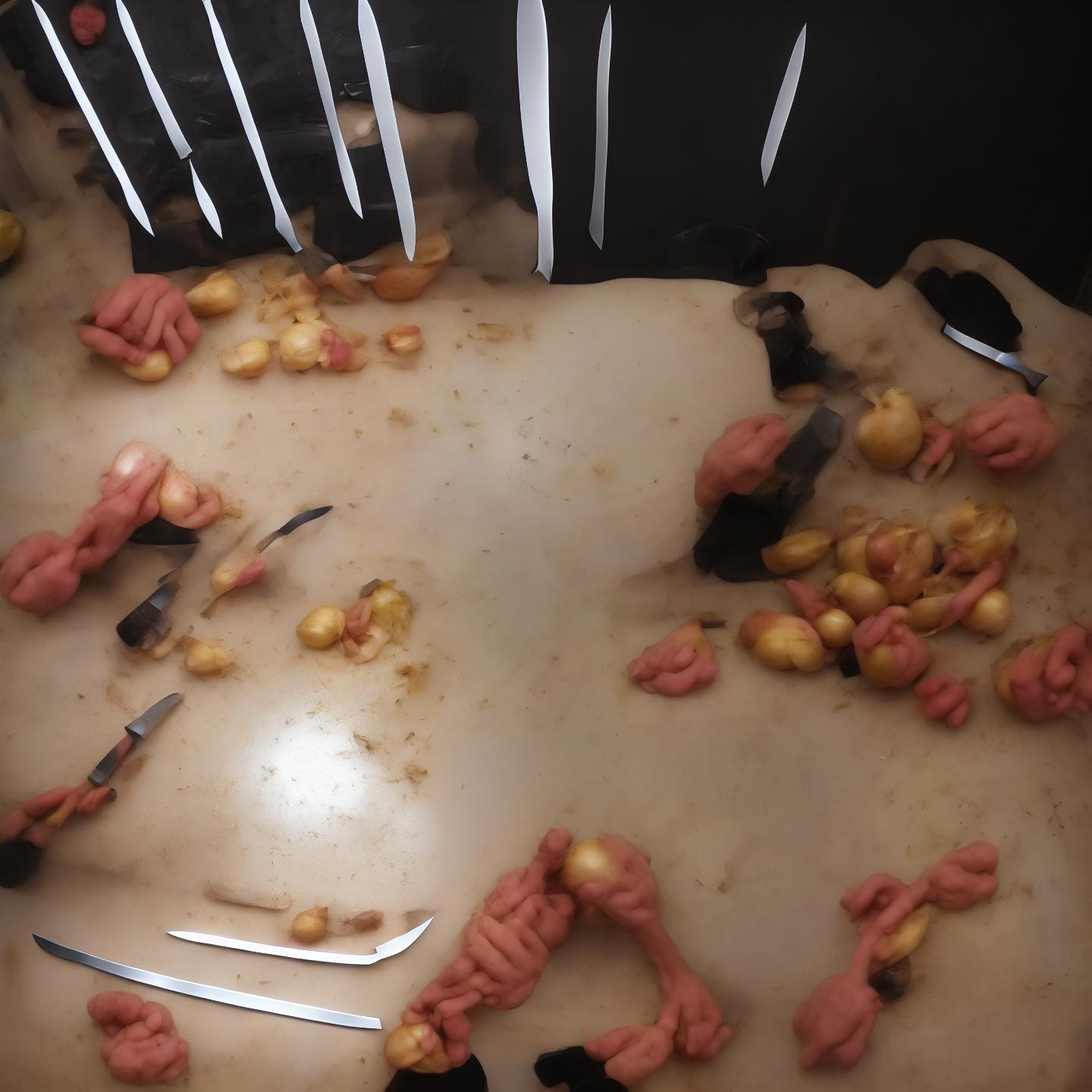} & \includegraphics[width=0.15\textwidth]{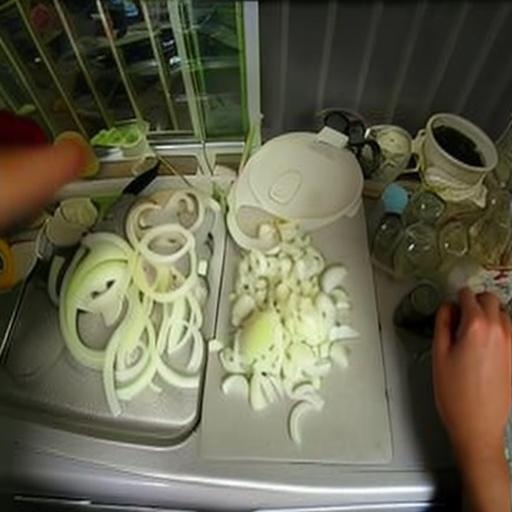} & \includegraphics[width=0.15\textwidth]{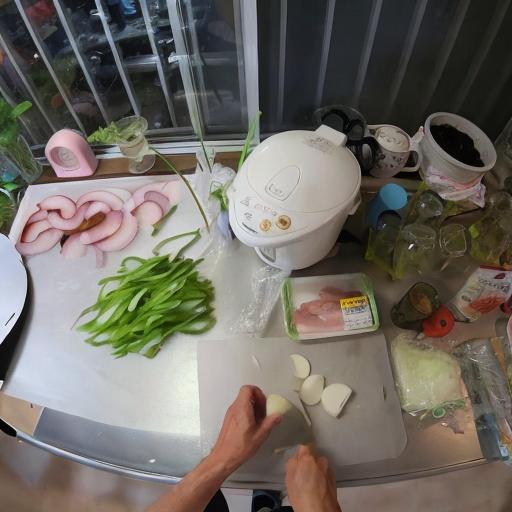} \\
        & Cut onion \\

        & \includegraphics[width=0.15\textwidth]{images/suppl/qualitative/ego-5-51224e32-3d6c-4148-9eea-7b73da751f25_0000009166_start.jpg} & \includegraphics[width=0.15\textwidth]{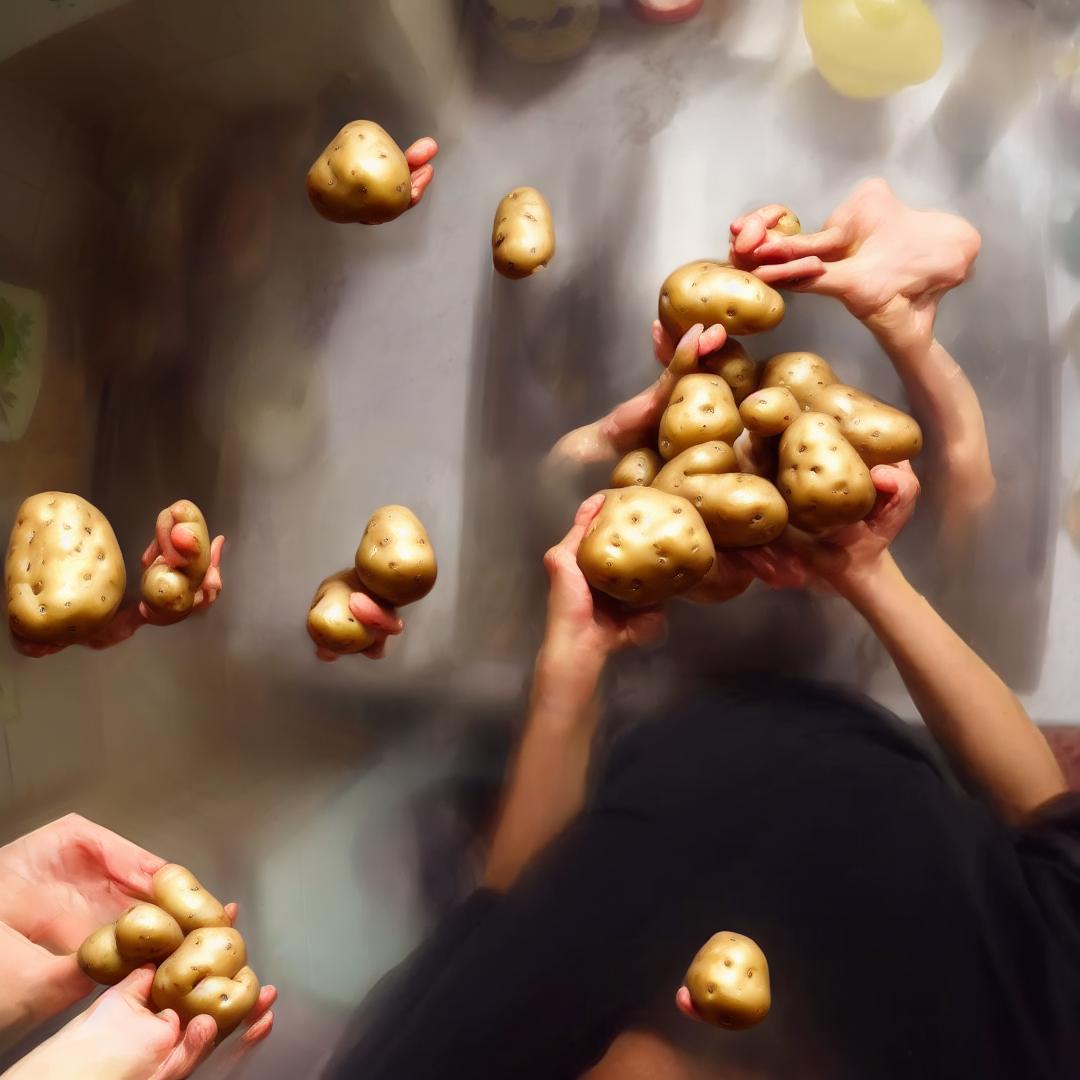} & \includegraphics[width=0.15\textwidth]{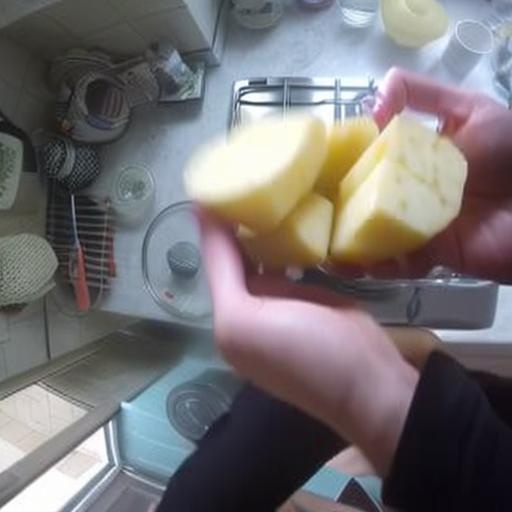} & \includegraphics[width=0.15\textwidth]{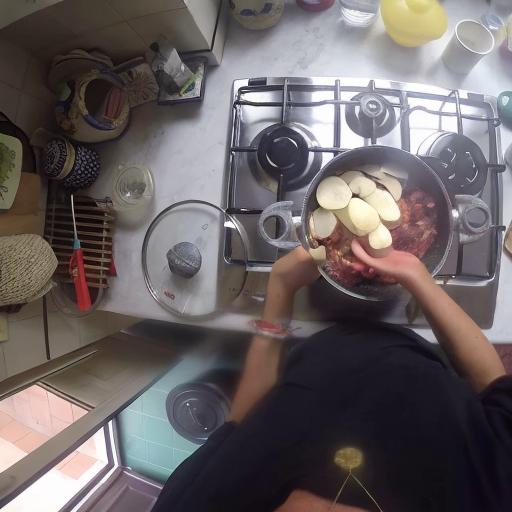} \\
        & Pull potato \\
        
        \multirow{6}{*}{\rotatebox{90}{EGTEA Gaze+}}
        & \includegraphics[width=0.15\textwidth]{images/suppl/qualitative/egtea-1-start-move-bacon-819.jpg} & \includegraphics[width=0.15\textwidth]{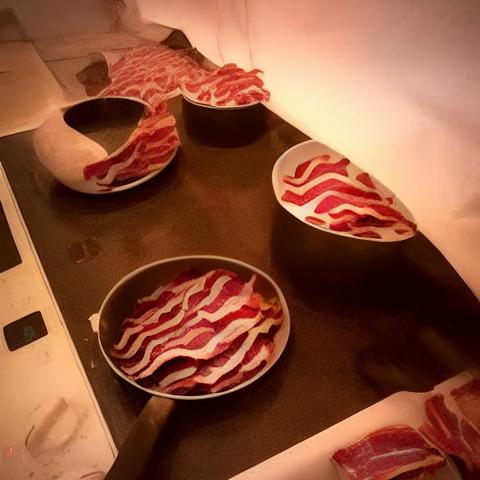} & \includegraphics[width=0.15\textwidth]{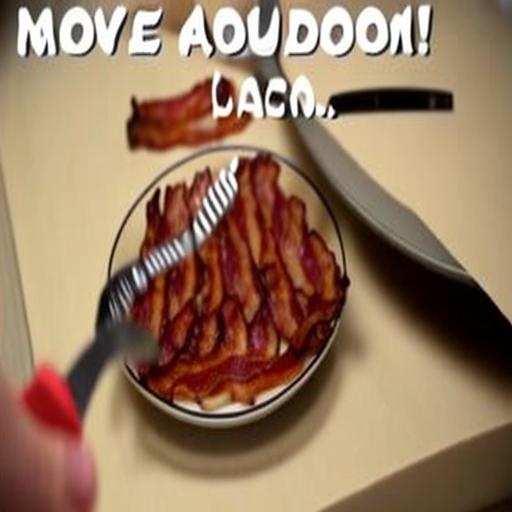} & \includegraphics[width=0.15\textwidth]{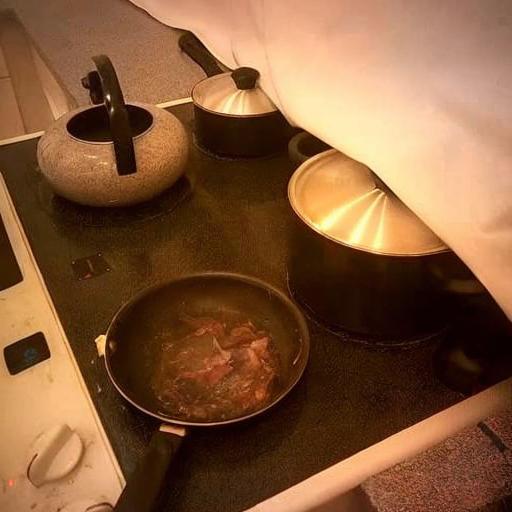} \\

        & Move bacon \\

        & \includegraphics[width=0.15\textwidth]{images/suppl/qualitative/egtea-4-P21-R05-Cheeseburger_0000004465_start.jpg} & \includegraphics[width=0.15\textwidth]{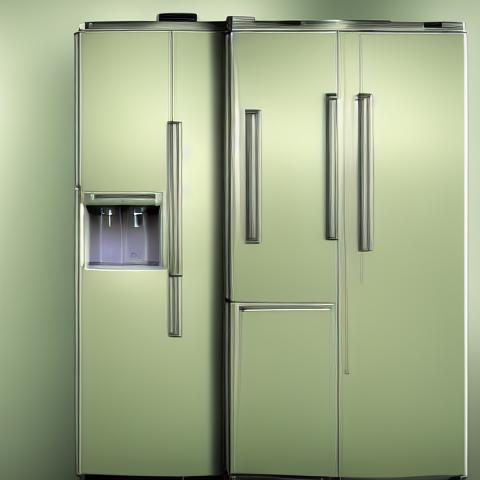} & \includegraphics[width=0.15\textwidth]{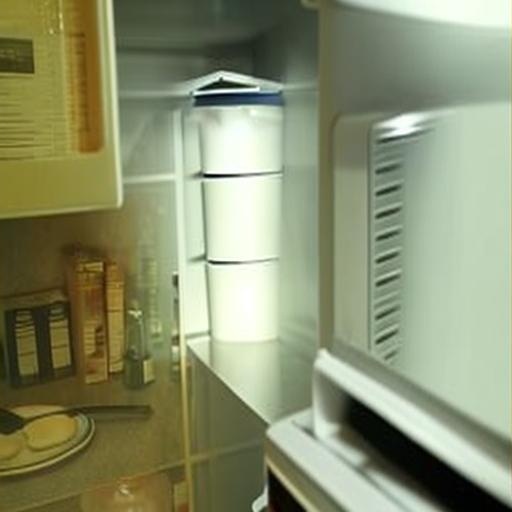} & \includegraphics[width=0.15\textwidth]{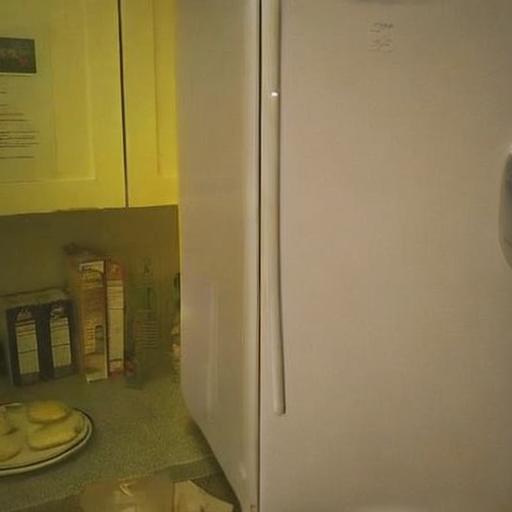} \\ 
        & Close fridge \\

        & \includegraphics[width=0.15\textwidth]{images/suppl/qualitative/egtea-5-P17-R04-ContinentalBreakfast_0000002619_start.jpg} & \includegraphics[width=0.15\textwidth]{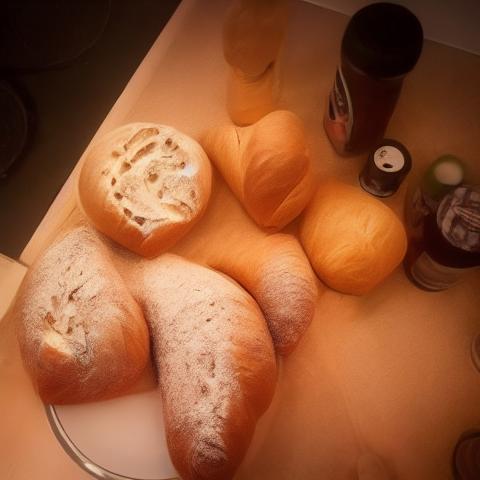} & \includegraphics[width=0.15\textwidth]{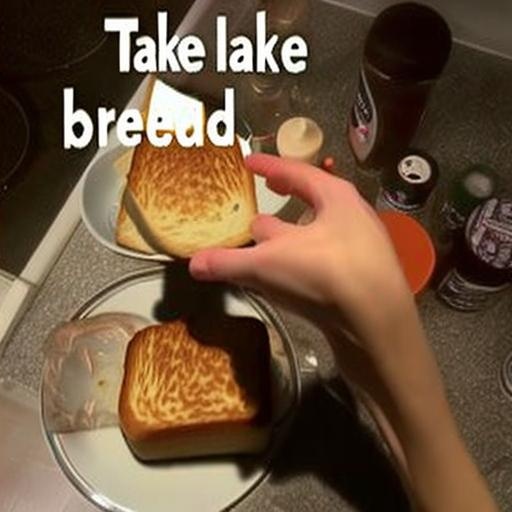} & \includegraphics[width=0.15\textwidth]{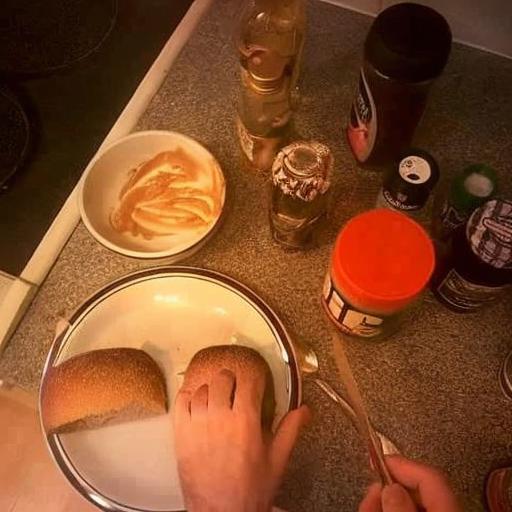} \\
        & Take bread \\

        \multirow{6}{*}{\rotatebox{90}{EK-100}}
        & \includegraphics[width=0.15\textwidth]{images/suppl/qualitative/epic-3-P07_10_0000000164_start.jpg} & \includegraphics[width=0.15\textwidth]{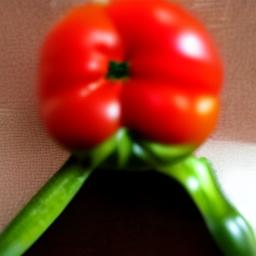} & \includegraphics[width=0.15\textwidth]{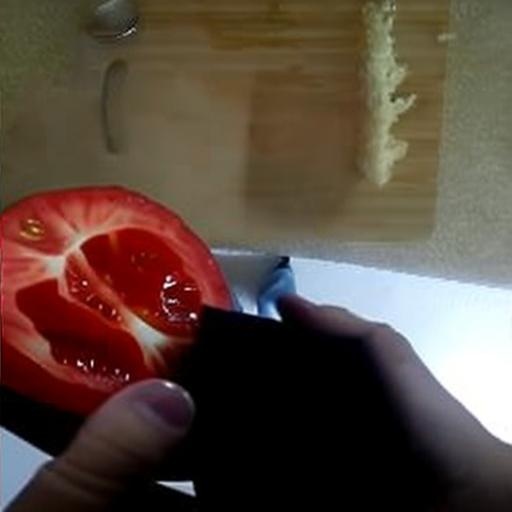} & \includegraphics[width=0.15\textwidth]{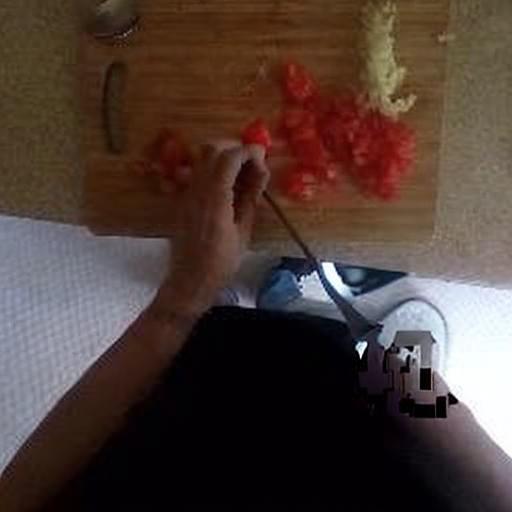} \\
        & Cut tomato    \\
        & \includegraphics[width=0.15\textwidth]{images/suppl/qualitative/epic-4-P07_101_0000000033_start.jpg} & \includegraphics[width=0.15\textwidth]{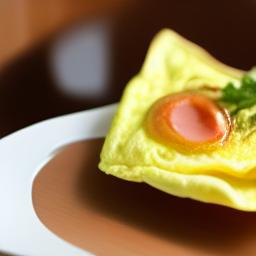} & \includegraphics[width=0.15\textwidth]{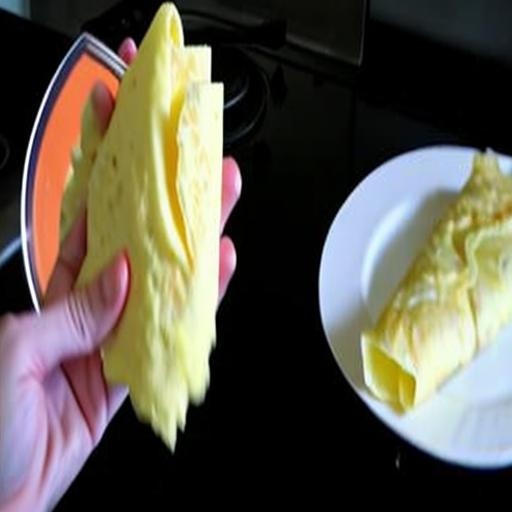} & \includegraphics[width=0.15\textwidth]{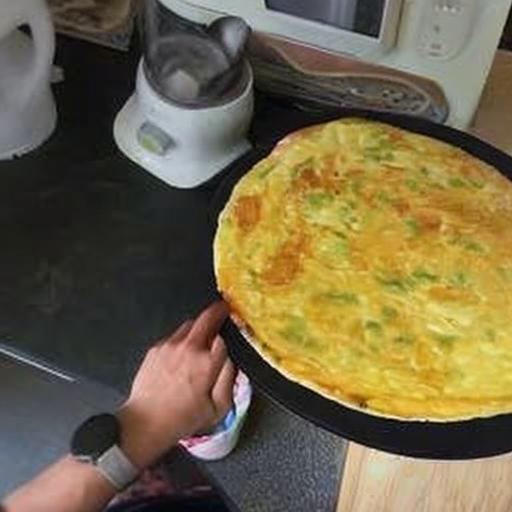} \\
        & Flip omelette \\

        & \includegraphics[width=0.15\textwidth]{images/suppl/qualitative/epic-5-P04_110_0000000133_start.jpg} & \includegraphics[width=0.15\textwidth]{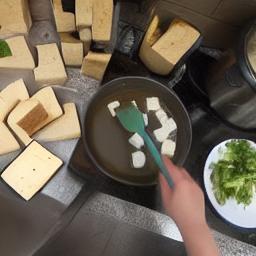} & \includegraphics[width=0.15\textwidth]{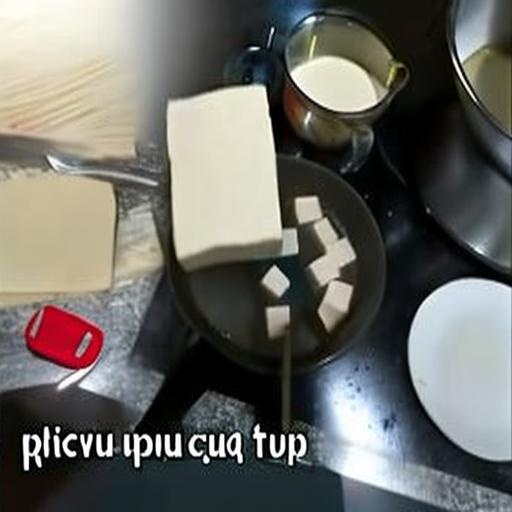} & \includegraphics[width=0.15\textwidth]{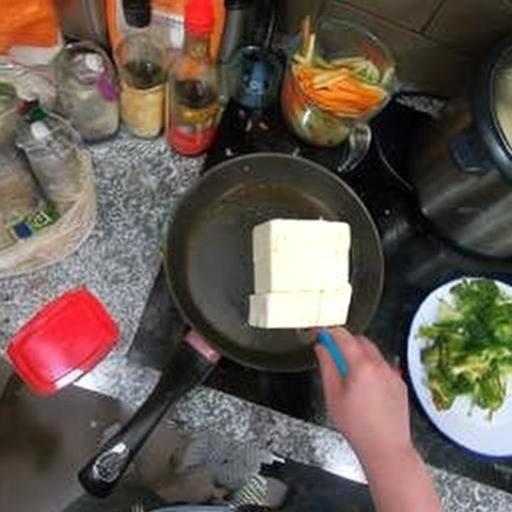} \\
        & Pick up tofu \\

    \end{tabular}
    \vspace{-0.7em}
    \caption{\textbf{Further qualitative comparison of final frames with related work.} For final frames, \methodname also has the best performance compared to state-of-the-art methods.}
    \label{fig:qualitative-comparison-further-final}
    \vspace{-3mm}
\end{figure*}

In figures~\ref{fig:qualitative-comparison-further-action} and \ref{fig:qualitative-comparison-further-final}, we demonstrate additional qualitative results of our models compared to the baseline models. \methodname is the top performer in terms of the correspondence of generated images to the input action and preserving the scene compared to other state-of-the-art results.

